\documentclass[10pt]{article} % For LaTeX2e
\usepackage[accepted]{tmlr}
\usepackage{natbib}
\usepackage{multirow}
\usepackage{amsmath,amsfonts,bm}

\def\Figref#1{Figure~\ref{#1}}

\def\eqref#1{equation~\ref{#1}}
\def\1{\bm{1}}

\DeclareMathAlphabet{\mathsfit}{\encodingdefault}{\sfdefault}{m}{sl}
\SetMathAlphabet{\mathsfit}{bold}{\encodingdefault}{\sfdefault}{bx}{n}

\usepackage{amsfonts}       % blackboard math symbols
\usepackage{nicefrac}       % compact symbols for 1/2, etc.
\usepackage{microtype}      % microtypography

\usepackage[subtle]{savetrees}
\usepackage{hyperref}
\usepackage{url}
\usepackage{booktabs}
\usepackage{graphicx}
\usepackage{xr}
\usepackage{xcolor}
\usepackage{algorithm}
\usepackage{algpseudocode}
\usepackage{amsthm}
\usepackage{lineno}
\usepackage[utf8]{inputenc}
\usepackage{float}

\usepackage[normalem]{ulem}
\usepackage[parfill]{parskip}
\usepackage{color,colortbl}
\usepackage{tcolorbox}
\usepackage{float}
\definecolor{mydarkblue}{rgb}{0,0.08,0.45}

\newcommand{\ie}{{i.e.,}}
\newcommand{\nslb}{{NLS}}

\tcbset{
  insightbox/.style={
  colback=cyan!15!white, 
  boxrule=0.2mm, 
  arc=1mm,
  top=1pt,
  bottom=1pt,
  }
}

\title{PDEInvBench:\\ A Comprehensive 
Dataset and Design Space Exploration of Neural Networks for PDE Inverse Problems}

\author{\name Divyam Goel \email divyam123@berkeley.edu \\
      \addr Department of Computer Science, UC Berkeley
      \AND
      \name Nithin Chalapathi \email nithinc@berkeley.edu \\
      \addr Department of Computer Science, UC Berkeley
      \AND
      \name Sanjeev Raja \email sanjeevr@berkeley.edu \\
      \addr Department of Computer Science, UC Berkeley
      \AND
      \name Aditi S.~Krishnapriyan \email aditik1@berkeley.edu\\
      \addr Departments of Computer Science and Chemical Engineering\\UC Berkeley; LBNL}
\let\origmaketitle\maketitle
\renewcommand{\maketitle}{%
  \begingroup
    \let\savedcr\\%
    \raggedright%
    \let\\\savedcr%
    \setlength{\parskip}{0.5pc}%
    \origmaketitle%
  \endgroup
}%

\def\month{05}  % Insert correct month for camera-ready version
\def\year{2026} % Insert correct year for camera-ready version
\def\openreview{\url{https://openreview.net/forum?id=MSjhqRnNyZ}} % Insert correct link to OpenReview for camera-ready version

\newcommand{\inversebench}{\textit{PDEInvBench}}

\begin{document}
\maketitle

\begin{abstract}
%\sr{think about some more interesting name for dataset like InverseBench or something?}
%\nc{Added a cmd for \inversebench ($\backslash$inversebench)}
%\ak{InverseBench is already taken :( }
%\ak{Should this still be focused on NO if other architectures are benchmarked too?}

Inverse problems in partial differential equations (PDEs) involve estimating the physical parameters of a system from observed spatiotemporal solution fields, a fundamental task in numerous scientific domains.
Neural networks are well-suited for PDE parameter estimation due to their capability to model function-to-function space transformations.
While existing benchmarks of machine learning methods for PDEs primarily focus on the forward problem --- mapping physical parameters to solution fields---to our knowledge, there are no similar comprehensive studies and benchmark datasets on PDE inverse problems, i.e., mapping solution fields to underlying physical parameters. We fill this gap by introducing \inversebench~, a comprehensive benchmark dataset consisting of numerical simulations for both time-dependent and time-independent PDEs across a wide range of physical behaviors and parameters. 
Our dataset includes evaluation splits that assess performance in both in-distribution and various out-of-distribution settings. 
Using our benchmark dataset, we comprehensively explore the design space of neural networks for PDE inverse problems along three key dimensions: (1) optimization procedures, analyzing the role of supervised, self-supervised, and test-time training objectives on performance, (2) problem representations, where we study the value of architectural choices with different inductive biases and various conditioning strategies, and (3) scaling, which we perform with respect to both model and data size. 
Our experiments reveal several practical insights: 1) neural networks perform best with a two-stage training procedure: initial supervision with PDE parameters followed by test-time fine-tuning using the PDE residual, 2) incorporating PDE derivatives as input features consistently improves accuracy, and 3) increasing the diversity of initial conditions in the training data yields greater performance gains than expanding the range of PDE parameters. 
We make our dataset and evaluation codebase freely available to facilitate reproducibility and further development of our work. The dataset is available at \url{https://huggingface.co/datasets/DabbyOWL/PDE\_Inverse\_Problem\_Benchmarking}, and the codebase is available at \url{https://github.com/ASK-Berkeley/PDEInvBench}.

\end{abstract}
\section{Introduction}

Inverse problems involve inferring unknown parameters or governing laws of a physical system using observed data, such as measurements of system behavior over space and time. They are ubiquitous in numerous domains including geophysics~\citep{tarantola_inverse_2005}, nanophotonics~\citep{molesky_inverse_2018}, biomedical imaging~\citep{vlaardingerbroek_magnetic_2013}, and fluid dynamics~\citep{karnakov_solving_2024}. In the case of systems modeled by partial differential equations (PDEs), the inverse modeling task is to map observed spatiotemporal solution fields to the underlying PDE parameters, which are typically physical constants such as viscosity, diffusivity, or the Reynolds number. 

Recent advances in machine learning (ML) have shown promise in learning effective representations for PDE-related tasks~\citep{rahman_u-no_2023, tran_factorized_2023, herde_poseidon_2024}. In particular, neural operators (NO) \citep{li_fourier_2021,kovachki_neural_2023} have emerged as powerful tools for modeling the mapping between function spaces. While significant progress has been made in establishing effective NO recipes for the \textit{forward} problem~\citep{gupta_towards_2022,takamoto_pdebench_2022, lu_comprehensive_2022}---predicting spatiotemporal solution fields from known PDE parameters---there has been comparatively less investigation into PDE inverse problems. 
The ill-posedness of PDE inverse problems, including issues of non-existence, non-uniqueness, and instability with respect to noisy observational measurements~\citep{heinz_regularization_1996,tarantola_inverse_2005,kitanidis_bayesian_2010, isakov_inverse_2017}, represents a challenging regime with distinct challenges compared to the forward problem.

In this work, we seek a principled understanding of the design space of neural network approaches --- both within and outside the neural operator framework ---  for solving PDE inverse problems. As the first step, we introduce \inversebench, which is, to our knowledge, the first comprehensive dataset dedicated to benchmarking the performance of ML approaches for solving PDE inverse problems. Our dataset consists of high-quality numerical simulations of 5 PDE systems at various spatial resolutions, including  Darcy flow (241x241, 421x421), reaction-diffusion (128x128, 512x512), unforced (64x64, 256x256) and forced Navier-Stokes (64x64, 2048x2048), and Korteweg-De Vries (256), at a wide range of physical parameter values and initial conditions. The PDEs cover a range of mathematical forms, including elliptic, hyperbolic and parabolic ~\citep{evans_partial_2022}. Similarly, parameter ranges are chosen to exhibit a variety of behaviors including Turing bifurcations, diffusion, steady-states, turbulence, and laminar flow. 
%\nc{todo Any states I missed?} 
We construct evaluation splits for each PDE system to investigate performance in both in-distribution and various challenging out-of-distribution settings.

Our second main contribution is the comprehensive benchmarking of ML-based methods for PDE inverse problems.
We organize our investigation around three fundamental design axes:

\begin{enumerate}
\itemsep-1pt
\item \textbf{Model optimization procedure:} What is the optimal strategy for training neural networks for PDE inverse problems? Is a supervised loss from paired parameter-solution data sufficient, or is it also beneficial to incorporate physics-based PDE residual terms? How can test-time adaptation be used to improve generalization to new parameter regimes without explicit data?
% While adding physics-informed constraints can improve physical consistency, it also creates a more complex loss landscape that may be harder to optimize effectively. Are there diminishing returns when increasing the complexity of the loss function? For test-time adaptation to new scenarios, how should self-supervised fine-tuning be constrained to prevent divergence from reasonable parameter ranges?
\item \textbf{Problem representation and inductive bias:} The choice of neural architecture encodes implicit assumptions about the underlying problem. Do spectral biases (as in the Fourier Neural Operator), local convolutional patterns (as in ResNets), or global attention mechanisms (as in Transformers) best capture the relationship between solution fields and physical parameters? Should features like the derivative terms of the PDE be explicitly provided as input, or left for the network to learn implicitly? 
%much temporal conditioning is necessary to reliably infer parameters for different PDE systems \ak{Does it make sense to more explicitly define temporal conditioning?}?
\item \textbf{Scaling properties:} Resource constraints often force tradeoffs. Is it more beneficial to increase model size or to generate more training data? When generating data, should one prioritize covering more physical parameter values or more initial conditions? 
\end{enumerate}

We evaluate models using standard ML metrics such as relative error and the slope of scaling curves. We also perform physically motivated evaluations, such as checking whether the energy spectra of numerical simulations evolved with the predicted PDE parameter match that of reference simulations.

From our extensive experiments, we distill key insights that can guide practitioners in deploying neural networks for inverse problems. 
\begin{itemize}
    \item A two-stage training approach is preferable when learning neural networks for PDE inverse problems: 1) training with a supervised data loss, followed by 2) test-time training using the PDE residual.
    \item Explicitly conditioning networks on spatial and temporal derivatives significantly improves performance across all models and datasets. 
    \item For time-dependent PDEs, the FNO architecture generally outperforms ResNet and Transformer-style architectures. However, given the vast space of architectural factors, more studies should be conducted to better understand how model inductive biases affect predictions for PDE inverse problems.
    \item For a fixed computational budget, generating multiple trajectories for each parameter value using different initial conditions yields greater performance gains than expanding parameter coverage.
\end{itemize}

We release the PDEInvBench dataset\footnote{
\scriptsize\url{https://huggingface.co/datasets/DabbyOWL/PDE\_Inverse\_Problem\_Benchmarking}} and evaluation codebase\footnote{\scriptsize \url{https://github.com/ASK-Berkeley/PDEInvBench}} as a modular, standardized environment designed to facilitate systematic exploration of the PDE inverse problem design space and serve as a foundation for future methodological advancements by the community.

\begin{figure}[htbp!]
    \centering
    \includegraphics[width=\linewidth]{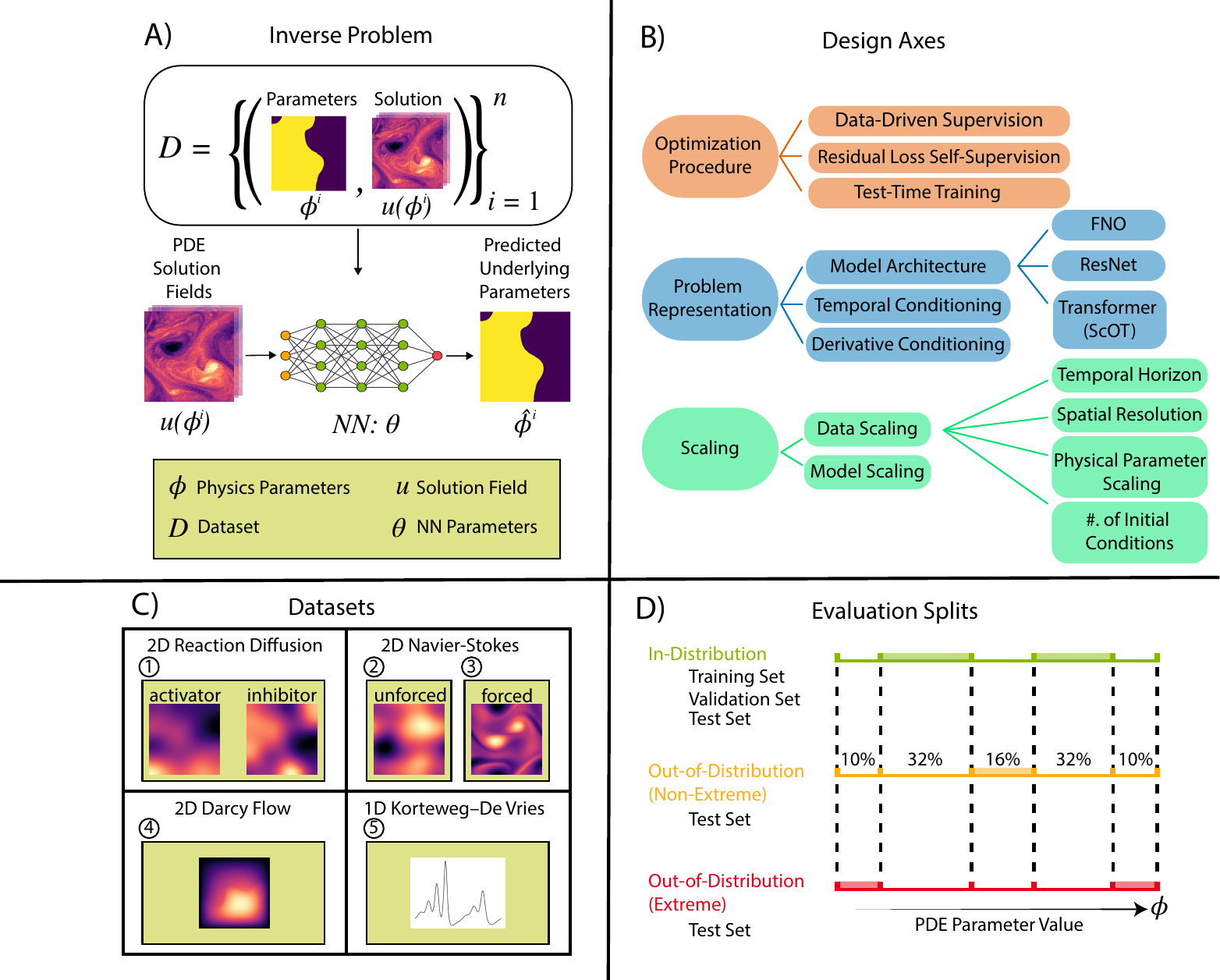}
    \caption{\textbf{Overview of design investigation of neural networks for PDE inverse problems.} \textbf{(A)} We consider the inverse problem setting in which a neural network learns to map from PDE solution fields $u(\phi)$ to predicted PDE parameters $\hat{\phi}$, using a dataset $D$ consisting of parameter-solution pairs as supervision. \textbf{(B)} Our investigation is split across three key design axes: optimization procedures, problem representation, and scaling properties. \textbf{(C)} We benchmark on diverse datasets spanning different PDE families, including 2D Reaction Diffusion, 2D Navier-Stokes, 2D Darcy Flow, and 1D Korteweg-De Vries. \textbf{(D)} We perform evaluations on both in-distribution parameter regimes and various out-of-distribution scenarios to assess generalization capabilities.}
    \label{fig:overview_fig}
\end{figure}

\section{Related works}

\paragraph{Neural operators.} 
Neural Operators (NOs) are a class of discretization-invariant, universal function approximators which learn mappings between infinite-dimensional function spaces, making them particularly well-suited for modeling PDE problems~\citep{kovachki_neural_2023}.
NOs typically involve a point-wise lifting operation, multiple layers of a learnable iterative kernel integration, followed by another point-wise projection to the output function space. 
For a mathematical description of the NO framework, see Appendix~\ref{sec:appendix-neural-operators-framework}.
Several types of NOs have emerged with different inductive biases. 
The Fourier Neural Operator (FNO)~\citep{li_fourier_2021}, including its many extensions~\citep{tran_factorized_2023,  kochkov_machine_2021} and generalizations~\citep{du_neural_2024}, leverages a spectral representation by parameterizing the integral kernel in the Fourier domain, enabling efficient modeling of global dependencies and smooth functions that naturally arise in physical systems. 
Transformers adapted for PDEs---like the scalable Operator Transformer (scOT)~\citep{herde_poseidon_2024} which modifies the Swin-Transformer~\citep{liu_swin_2022}---use self-attention to model both local and global dependencies in a manner compatible with the NO framework.
Multiple other attention-based architectures have been proposed~\citep{cao_choose_2021, hao_gnot_2023, alkin_universal_2024, wu_transolver_2024}, but we use scOT, which is implemented in the HuggingFace Transformers library~\citep{wolf_transformers_2020}.

\paragraph{Inverse modeling using neural networks and Neural Operators.}
%\sr{can we pick like 50-60\% of these and push the rest to an extended related work section in the appendix?}
%\nc{appendix label~\ref{sec:appendix-additional-related-works}}
Several studies have explored neural networks for inverse PDE parameter estimation. 
~\citet{li_fourier_2021} use a function space Markov chain Monte Carlo method to recover the distribution of initial conditions using FNO as a surrogate numerical solver.
~\citet{li_physics-informed_2024} extend this approach with a physics-informed losses during training and test-time refinement (similar to the test-time training (TTT) method we benchmark), enabling both direct inverse modeling and gradient-based parameter estimation using forward surrogates. 
~\citet{mackinlay_model_2021} investigate regularization terms when using gradient descent through a neural network forward surrogate to recover system parameters. 
~\citet{molinaro_neural_2023} propose Neural Inverse Operators (NIO), combining DeepONets with FNOs for recovering spatial coefficient fields from boundary measurements, and evaluate on electrical impedance tomography, inverse scattering, radiative transport, and seismic imaging datasets. While NIO proposes a specific architecture for a complementary set of PDE systems, our works focuses on benchmarking across architectures.
% ~\citet{molinaro_neural_2023} combine DeepONets~\citep{lu_learning_2021} with FNOs for coefficient reconstruction from boundary measurements, and evaluate on electrical impedance tomography, inverse scattering, radiative transport and seismic imaging datasets. 
~\citet{jiao_solving_2024} use DeepONets to identify diffusion coefficients on unknown manifolds.
Latent Neural Operators~\citep{wang_latent_2024} introduce a ``Physics-Cross Attention'' to map geometric inputs into a latent space for coupled forward-inverse prediction. 
~\citet{cho_physics-informed_2025} provide theoretical stability guarantees for inverse problems and introduce PI-DION, an architecture tailored for solution field reconstruction and parameter estimation from partial measurements. ~\citet{zheng2025inversebench} uses diffusion models to solve a broad class of inverse problems, but does not consider PDE-specific design choices, and does not consider multiple choices of PDE parameter.
\inversebench~advances this avenue of work by introducing a benchmark suite to accelerate model development and deriving relevant insights.

%\wip{Our work fills in the gap of extensive evaluations on a variety of fluids datasets spanning different classes with complete measurements.}

% \sr{I think we can cut this para for space}
% There have been a number of non-NO methods proposed including~\citet{zhao_learning_2022} who develop an approach through learning a generative prior by pre-training both a coordinate network and a Graph Neural Network forward model. 
% At test-time, they optimize a latent code to match sparse observations.
% ~\citet{pakravan_solving_2021} replace the decoder of an auto-encoder with numerical discretization schemes to recover PDE parameter \wip{fields}. \nc{what}
% ~\citet{goh_solving_2022} derive variational auto-encoders~\citep{kingma_auto-encoding_2013} from a divergence-based variational inference perspective to obtain uncertainty quantification for Bayesian inverse problems.
% Our work contributes to this space by deriving general insights that apply across modeling frameworks.

\paragraph{Existing datasets and benchmarks.}
Recent years have seen numerous PDE-specific benchmark datasets emerge, with most providing limited benchmarking of inverse problems, or consisting of numerical simulations performed with a limited range of PDE parameter values ~\citep{lu_comprehensive_2022, bhan_pde_2024, toshev_lagrangebench_2024, hao_pinnacle_2023}.
PDEBench~\citep{takamoto_pdebench_2022} evaluates several PDE systems but only generate a few PDE parameters per system, while varying the initial conditions used to evolve solutions.
PDEArena~\citep{gupta_towards_2022} examines Navier-Stokes across a range of bouyancy values, but focuses their analysis exclusively on the forward problem.
BubbleML~\citep{hassan_bubbleml_2023}, similar to PDEBench, only examines a limited number of parameters within the context of the forward problem.
The Well~\citep{ohana_well_2024} is a large-scale dataset covering a broad set of PDEs with a wide range of physical parameter values. While their baselines and benchmarks target the forward problem, several datasets in the The Well (e.g., \texttt{acoustic\_scattering}, \texttt{helmholtz\_staircase}) are well-suited for inverse problems, and integrating them into our inverse problem evaluation framework is an important future direction.
% The Well, a recent large dataset for the forward problem, covers a broad set of PDEs and focuses on generating a large number of initial conditions, but only covers a small range of physical parameter values.
~\citet{kohl_benchmarking_2023} focuses on fluid turbulence across a range of parameters, but benchmark autoregressive diffusion models only for the forward problem.
By contrast, our dataset ~\inversebench~is the only benchmark covering a wide range of PDE parameters (i.e., physical behaviors) and PDE systems in the inverse problem setting.
Other works either exclusively focus on the PDE forward problem, or do not thoroughly study the inverse problem on multi-parameter datasets and across multiple design axes. 

\paragraph{System identification.} The PDE inverse problems we consider in this work are related to a broad line of literature on system identification \cite{aastrom1971system, ljung1998system}. Recent advances in multi-environment adaptation for system identification ~\citep{blanke2024interpretable,pmlr-v162-kirchmeyer22a, koupa2024boosting} focus on meta-learning task-agnostic representations with test-time adaptation of task-specific components according to the underlying parameters. However, these approaches primarily focus on solving the forward problem and do not directly predict underlying parameters. In contrast, our focus is on directly predicting the underlying PDE parameters while leveraging knowledge of the governing PDE equations for test-time adaptation. This allows for straightforward test-time tuning without the need to distinguish between task-agnostic and task-specific representations.

\section{Preliminaries}
\label{sec:background}

\paragraph{Partial differential equations (PDEs).} We consider the PDEs of the form,
\begin{align}
    &\mathcal{F}_\phi(u(x, t)) = 0 &x \in \mathbb{X}, t \in [0, T], \phi \in \Phi 
    \label{eq:pde-definition} \\
    &\mathcal{B}(u(x, t)) = 0 & x \in \partial \mathbb{X} \notag\\
    &u(x, 0) = u_0 & x \in \mathbb{X}\notag
\end{align}

$\mathcal{F}$ is a differential operator characterizing a family of PDEs, with solutions $u$ defined on the spatial domain $\mathbb{X}$ and temporal range $[0, T]$. $x,t$ define spatial-temporal points, and $\phi$ denotes physical parameters of the PDE family (e.g., diffusion coefficient, viscosity, density) drawn from a distribution of possible parameters $\Phi$. 
In general, $\phi$ may be an arbitrarily complex function of space and time, but in the systems we evaluate $\phi$ is typically a constant scalar or a spatially varying scalar field $\phi: \mathbb{X} \rightarrow \mathbb{R}$.
$\mathcal{F}_\phi$ and $u$ are potentially highly non-linear in $x$ and $t$.
$\mathcal{B}$ and $u_0$ denote boundary and initial conditions. 
For brevity, we abbreviate the entire solution over the spatial domain at time $t$ as $u_t$, \ie~$u_t = u(\cdot, t)$.

\paragraph{Inverse problem and data loss.} 
The inverse problem denotes the task of learning a function $f_\theta$ with parameters $\theta$ that maps $k$ steps of observed PDE dynamics $u_{t - k}, ..., u_t$ from a PDE of the form of Equation \ref{eq:pde-definition} to the underlying PDE parameter $\phi$.
Formally, the optimization problem is posed as,
\begin{align}
\label{eq:basic_opt}
\theta^* = \underset{\theta}{\arg \min}\ \mathbb{E}_{\phi \sim \Phi}\| f_\theta(u_{t - k}, ..., u_t) - \phi\|_2,
\end{align}
where $u$ satisfies $\mathcal{F}_\phi(u) = \mathbf{0}$.
In practice, the expectation is approximated by an average over a finite dataset $\mathcal{D}_\text{train} = \{ \phi^i, u^i \}_{i=1}^N$ of $N$ PDE parameters paired with solution trajectories from a numerical simulation, where each $u^i$ satisfies $\mathcal{F}_\phi(u^i) = \mathbf{0}$. 
We refer to supervised learning using paired PDE parameter data as \emph{data-driven supervision}.
The data-driven inverse problem objective is then:
\begin{align}
    \label{eq:data_loss}
    \mathcal{L}_\text{data} = \frac{\| f_\theta(u_{t - k}^i, ..., u_t^i) - \phi^i\|_2}{||\phi^i||_2}.
\end{align}
Note that we use the relative error instead of the standard $L_2$ loss as physical parameters may span several orders of magnitude (e.g., viscosity $\nu \in [10^{-5}, 10^{-2}]$ in 2D turbulent flow).
% TODO: revisit this, sounds clunky
Our problem setting is challenging due to the presence of a range of PDE parameters $\Phi$, which induces a wide variety of physical behaviors in the solution $u$.
For example, given two parameter settings $\phi_1, \phi_2 \in \Phi$, the solution to $\mathcal{F}_{\phi_1}(u(x, t)) = \mathbf{0}$ may be diffusive, while the solution to $\mathcal{F}_{\phi_2}(u(x, t)) = \mathbf{0}$ may converge to a nontrivial steady-state.
This is distinct from prior works on neural PDE solvers which typically consider a single or a few values of $\phi$ and only vary the initial or boundary conditions. Additionally, PDE inverse problems may be ill-posed; distinct parameters $\phi_1 \neq \phi_2$ can produce observations that are close or indistinguishable, \ie $|u^{\phi_1} - u^{\phi_2}|$ may be small even when $|\phi_1 - \phi_2|$ is large. 
The degree of ill-posedness depends on the number of observation frames, spatial resolution, and model input conditioning information. 
These factors are investigated in further depth in Section~\ref{sec:rep_results}
% \nc{cite some of the benchmarks to lend more weight to the statement} \dg{add cits to reinforce problem statement importance}

\paragraph{PDE residual loss.} 
We can incorporate physics-based constraints via the PDE residual $\|\mathcal{F}_\phi(u(x, t))\|_2$, where the derivatives in $\mathcal{F}_\phi$ can be computed using autodifferentiation or a finite difference scheme. We elect to use the latter; henceforth, when $\mathcal{F}$ appears in a loss function, this refers to a finite difference approximation of the true differential operator on a finite-resolution mesh. 
We can use the PDE residual as a self-supervised learning signal \cite{li_physics-informed_2024}. 
Given an inverse model $f_\theta$ that predicts parameters $\hat{\phi} = f_\theta(u_{t - k}^i, ..., u_t^i)$, the residual loss is,
\begin{equation}
  \mathcal{L}_{\text{res}} = \|\mathcal{F}_{\hat{\phi}}(u_{t - k}^i, ..., u_t^i)\|^2_2 = \|\mathcal{F}_{f_\theta(u_{t - k}^i, ..., u_t^i)}(u_{t - k}^i, ..., u_t^i)\|^2_2
  \label{eq:residual-loss-definition} \\
\end{equation}

\paragraph{Test-time training.} 
Test-time training (TTT) leverages the self-supervised nature of the residual loss for adaptation to specific parameters that may fall outside the training distribution. 
Given observed dynamics $u_{t - k}, ..., u_t$ at test time, we can adapt the inverse model $f_\theta$ via gradient updates on the following loss:
\begin{align}
    \label{eq:ttt_loss}
   \mathcal{L}_{\text{Tailor}} = \mathcal{L}_{\text{res}} + \alpha \mathcal{L}_{\text{anchor}}
\end{align}
where $\alpha \in [0, 1]$ is a weighting coefficient and,    
\begin{align}
   \mathcal{L}_{\text{anchor}}= \frac{\|f_{\theta}(u_{t - k}, ..., u_t) - f_{\theta_\text{frozen}}(u_{t - k}, ..., u_t)\|_2}{||f_{\theta_\text{frozen}}(u_{t - k}, ..., u_t)||_2} ,
\end{align}
where $\theta_\text{frozen}$ are model weights obtained after the initial training process. 
$\mathcal{L}_{\text{anchor}}$ prevents excessive deviations from the original model which we find helps stabilize tailoring.
\section{Datasets and evaluation metrics}
\label{sec:Datasets}
%\sr{beef up this section to make it seem like a central part/standalone contribution of the paper, not just a setup for the investigations}

\inversebench~contains five different PDE systems, each of which are simulated with a predefined range of PDE parameters chosen to span various physical behaviors including turbulence, steady-states, laminar flows, Turing patterns, and diffusion. 
A summary of the datasets are given in Table \ref{tab:pde_datasets}. 
For each sampled parameter value within the range (the number of distinct values and the sampling scheme is given in the ``Number of Parameter Values" column), a fixed set of Gaussian random fields, serving as initial conditions, are evolved using a numerical solver according to the governing equations.
After a fixed initial burn-in phase, solutions are recorded until reaching convergence (e.g., Turing bifurcations~\citep{borckmans_turing_1995}) or for a preset time horizon.
Details about the numerical solver parameters used, including the type of solver, time-stepping scheme, and burn-in periods, can be found in Appendix~\ref{sec:appendix-pdes-solver} alongside a complete description of the governing equations in Appendix~\ref{sec:appendix-dataset-eqn}. We also provide analysis of numerical convergence and energy spectra of the datasets in Appendix~\ref{sec:numerical_convergence_1} and Appendix~\ref{sec:appendix-tf-convergence}. Here, we provide a high-level description of all the datasets.

\paragraph{2D Reaction Diffusion (RD) [parabolic].} 
2D RD models two chemical species coupled non-linearly through the Fitzhugh-Nagumo equations~\citep{klaasen_stationary_1984,takamoto_pdebench_2022}. 
We focus on \emph{activator-inhibitor} systems where the activator promotes the production of both species while the inhibitor suppresses both.
There are 3 physical parameters $k$, $D_u$, and $D_v$. 
$D_u$ and $D_v$ are the diffusion coefficients for the activator and inhibitor, respectively.
Meanwhile, $k$ represents the balance between both species or the threshold for excitement. 
Our dataset contains simulations with a range of values for $k$, $D_u$, and $D_v$, spanning a wide range of induced physical behaviors, from completely dissipative to Turing bifurcations~\citep{borckmans_turing_1995}.
For simplicity, when training inverse models, we only learn over \emph{one} parameter at a time (i.e., either $k$ or $D_u$) while treating the others as known. 
When $k$ is unknown, we refer to the dataset as 2D RD-$k$ and similarly, when $D_u$ is unknown as 2D RD-$D_u$.
% Solutions are generated using an RK45 explicit solver~\citep{dormand_family_1980, takamoto_pdebench_2022}.

\paragraph{Unforced 2D Navier-Stokes (NS) [parabolic].}
The unforced NS equations govern fluid flow with no external energy injection~\citep{du_neural_2024}. The parameter range covers the laminar flow regime.
Both the data and the solver use the vorticity form of the PDE, with a single parameter, viscosity $\nu$, serving as a proxy for internal friction.
% Solutions are generated using a pseudo-spectral solver~\citep{du_neural_2024} with a Crank-Nicolson time-stepping scheme~\citep{canuto_spectral_2007, li_fourier_2021}.
\paragraph{Forced 2D Navier-Stokes (TF) [parabolic].}
2D TF models fluids across a range of Reynolds numbers in the turbulent regime. 
We use a Kolmogorov forcing function term on the second wavenumber (i.e., $-2\cos(2y)$)~\citep{kochkov_machine_2021}. 
As with unforced NS, the physical parameter of interest for the inverse problem is the viscosity $\nu$.
% All solutions in this dataset exhibit turbulence and are generated using a pseudo-spectral solver with a Crank-Nicolson time-stepping scheme~\citep{dresdner_learning_2023}.

\paragraph{Korteweg-De Vries (KdV) [hyperbolic].}
KdV models waves on a shallow-water surface and includes a dispersive ($\partial_{xxx}u$) and an advection term ($u\partial_xu$)~\citep{brandstetter_lie_2022}.
The physical parameter $\delta$ corresponds to the strength of the dispersive term~\citep{zabusky_interaction_1965}.
% Solutions are generated using a pseudo-spectral solver with an implicit Runge-Kutta method (Radau IIA)~\citep{hairer_stiff_1999,virtanen_scipy_2020, brandstetter_lie_2022}.

\paragraph{Darcy Flow (DF) [elliptic].}
% To complement our generated datasets, we also include Darcy Flow from~\cite{li_fourier_2021}. 
Darcy Flow models the movement of a fluid through a porous medium~\citep{li_fourier_2021}. 
All solutions converge to a non-trivial steady-state solution based on the diffusion coefficient field, making the system time-independent unlike previously described systems. 
The diffusion coefficient field, which is the parameter we learn over for the inverse problem, is a spatially varying scalar parameter rather than a constant scalar.

\begin{table}[htbp!]
\centering
\caption{Summary of PDE systems simulated in~\inversebench. The dataset includes 4 time-dependent PDEs and 1 time-independent PDE, with physical parameter regimes for each chosen to span a wide range of physical behaviors, including turbulence, steady-states, laminar flows, Turing patterns, and diffusion.} 
\resizebox{\textwidth}{!}{%
\begin{tabular}{lccccccc}
\toprule
\textbf{PDE System} & \textbf{Number of} & \textbf{Time-} & \textbf{Spatial} & \textbf{Parameter} & \textbf{Temporal Resolution (s)/} & \textbf{Number of} & \textbf{Number of ICs/Trajectories} \\
 & \textbf{Partial Derivatives} & \textbf{dependent} & \textbf{Resolutions} & \textbf{Range} & \textbf{Time Horizon (s)} & \textbf{Parameter Values} & \textbf{(per parameter combination)} \\
\midrule
 2D Reaction & 5 & \checkmark & 128$\times$128 & $k \in [0.005, 0.1]$, & [0.049 / 5] & $k$ - 2 linear-spaced, & 5 \\
Diffusion (2D-RD) & & & & $D_u \in [0.01, 0.5]$, & & $D_u$ - 28 linear-spaced, & \\
 & & & & $D_v \in [0.01, 0.5]$ & & $D_v$ - 27 linear-spaced & \\
\midrule
2D Navier-Stokes & 3 & \checkmark & 64$\times$64 & $\nu \in [10^{-4}, 10^{-2}]$ & [0.0468 / 3] & 101 log-spaced & 192 \\
Unforced (2D-NS) & & & & (Reynolds: 80-8000) & & & \\
 & & & & & & & \\
\midrule
2D Navier-Stokes & 3 & \checkmark & 64$\times$64 & $\nu \in [10^{-5}, 10^{-2}]$ & [0.23 / 14.75] & 120 log-spaced & 108 \\
Forced (2D-TF) & & & & & & & \\
 & & & & & & & \\
\midrule
Korteweg-De & 4 & \checkmark & 256 & $\delta \in [0.8, 5]$ & [0.73 / 102] & 100 linear-spaced & 100 \\
Vries (1D-KdV) & & & & & & & \\
\midrule
Darcy Flow (2D-DF) & 4 & $\times$ & 241$\times$241 & Piecewise constant & - & 2048 (i.i.d Thresholded Gaussian) & - \\
 & & & & diffusion coefficient & & & \\
 & & & & $a \in L^{\infty}((0,1)^2; \mathbb{R}^+)$ & & & \\
\bottomrule
\end{tabular}%
}
\label{tab:pde_datasets}
\end{table}

\paragraph{Evaluation settings.}
Each dataset consists of a broad range of parameter values for the training distribution. For all time-dependent PDEs, we create three different evaluation splits which test generalization in different settings.  Splits are categorized in relation to the training parameter range. The
\emph{Out-of-Distribution (Extreme)} split, referred to as OOD Extreme, tests generalization to extreme parameter values.
In the extreme set, parameters are drawn from the smallest and largest $10\%$ of the parameter range. 
Similarly, the \emph{Out-of-Distribution (Non-Extreme)} split, referred to as OOD Non-Extreme, tests generalization to out-of-distribution parameter values in the middle $16\%$ of parameter values which are omitted from the training parameter range. 
Finally, the \emph{In-Distribution} (ID) split consists of held-out parameters drawn from the training range. ~\Figref{fig:overview_fig}(D) visualizes how splits are partitioned for the single parameter case. For systems with more than one PDE parameter (e.g., 2D Reaction Diffusion), we generalize the percentages to a hypercube with each axis representing a single parameter (see Appendix~\ref{sec:appendix-eval-splits}) for a visual depiction. 
    
For Darcy Flow (the only time-independent PDE in our dataset), the physical parameter is a spatial \textit{field} which takes on one of two values, 3 or 12, in each position. We make only ID and OOD Extreme splits for Darcy flow. We define the splits by computing the fraction of grid points in each coefficient field that take on the maximum value (12), a statistic which is approximately normally distributed across the dataset. We partition coefficient fields according to this distribution using $\pm 1.5$ standard deviations, with the central mass defining the ID test set and the tails defining the OOD Extreme split (see Appendix~\ref{sec:appendix-eval-splits} for additional details).

\paragraph{Evaluation metrics.} 
We present results using two primary metrics: \emph{relative error} ($\downarrow$) and \emph{the negative slope of the line of best fit (\nslb)} ($\uparrow$).
%\nc{There has to be a better way to refer to this metric instead of NSLBF.}
Relative error is defined as $||\phi - \hat{\phi}||_2/||\phi||_2$ where $\hat{\phi}$ and $\phi$ denote the predicted and true physical parameter, respectively. This metric is used to assess the quality of an inverse model.
For experiments where we compare scaling approaches (e.g., relative error performance with respect to scaling the number of initial conditions versus scaling the number of generated physical parameters in~\Figref{fig:results-scaling}D-F), we use the \nslb.
From a plot of relative error versus a particular scaling axis, we compute a line of best fit, and we extract the \nslb~as the negative slope of that line. \nslb~measures the effect of scaling on the relative error by capturing the \emph{decrease} in the $y$-axis (i.e., relative error) per $x$-axis unit (i.e., method of scaling).
More negative values of the \nslb~indicate better scaling (i.e., faster reduction of the relative error due to scaling). This metric is related to well-established techniques used in the literature on neural scaling laws \cite{hestness2017deep, bahri2024explaining} and machine learning force fields. We select the best performing models on the validation splits using these metrics (see Appendix~\ref{sec:appendix-metrics}). In Appendix~\ref{sec:self-consistency-metric}, we present an additional evaluation to test the physical consistency of the predicted PDE parameters. We roll out numerical simulations with predicted PDE parameters and compare the resulting energy spectra with that of reference simulations run with the true PDE parameters. 
\section{Selected investigations for PDE inverse problems}
\label{sec:methods}
We now describe our methodology for exploring key design axes for solving PDE inverse problems using neural networks. We investigate 1) optimization procedure, 2) problem representation and inductive biases, and 3) scaling properties, with several experiments within each axis. For all experiments, we train models with 3 different random seeds and report error bars of the corresponding standard deviation. In this section, we select the most significant investigations, but include additional experiments and details in Appendices~\ref{sec:appendix-additional-investigations} and~\ref{sec:appendix-additional-results}.

\subsection{Optimization procedure} We investigate how different optimization approaches affect performance. Our exploration focuses on two key aspects: loss formulations and test-time training strategies.

\paragraph{Loss functions.}
We consider two primary loss formulations: a supervised, purely data-driven setting with the data loss defined in Equation \ref{eq:data_loss}, and a purely self-supervised, residual-driven setting with the PDE residual loss defined in Equation \ref{eq:residual-loss-definition}. In Appendix~\ref{sec:appendix-additional-results}, we also consider a generalized ``physics-informed" loss which combines the data-driven and PDE-residual setting.

\paragraph{Test-time training.} As defined in Section \ref{sec:background}, we adapt a pretrained inverse model at inference time using a self-supervised objective (Equation~\ref{eq:ttt_loss}). 
We take 50 gradient steps with a batch size of 32. We provide further details on test-time training in Appendix~\ref{sec:appendix-opt-procedure} 

\subsection{Problem representation and inductive biases} 
The choice of input features and inductive biases can significantly impact downstream performance. We investigate this through a comparative analysis of architectures and input representations.

\paragraph{Architecture comparison.} We compare four architectures in a parameter-controlled ($\sim 5$ million learnable parameters) manner (hyperparameters in Appendix~\ref{sec:appendix-investigations-problem-rep}).
The \emph{Fourier Neural Operator (FNO)} ~\citep{li_fourier_2021} serves as our baseline architecture and uses a spectral representation, similar to classical pseudo-spectral numerical methods.
\emph{ResNet}~\citep{he_deep_2016} tests the usefulness of locality in the spatial domain. % \ak{seems like an incomplete sentence}. 
\emph{DeepONet} ~\citep{lu2021deeponet} tests the usefulness of separating the encoding of input fields from the representation of output coordinates via branch-trunk decomposition.
Finally, we use the \emph{scalable Operator Transformer (scOT)} ~\citep{herde_poseidon_2024}, based on the Swin Transformers~\citep{liu_swin_2022}, to test the effectiveness of modeling local and global spatial dependencies through hierarchical attention mechanisms.
% Both FNO and scOT are by construction neural operators, and we include both to examine different classes of models.
Both FNO and scOT are neural operator architectures whose encoders are discretization invariant, and we include both to examine different classes of inductive biases developed for PDE applications.
DeepONet is also a neural operator architecture, but differs in that it parameterizes the operator through a learned basis expansion rather than enforcing discretization invariance through convolutional or attention-based encoders.
While not strictly a neural operator, ResNet operates over arbitrarily sized inputs due to the nature of convolutions which makes it a useful point of comparison. 
Note that the full inverse-problem pipeline (encoder followed by convolutional downsampling and an MLP regression head) is not discretization invariant for any architecture, as reducing a spatial field to a scalar parameter inherently requires resolution-dependent pooling and aggregation operations.

\paragraph{Input representations.} We examine the impact of explicitly providing derivative information as conditioning to the inverse model. 
In addition to the past $k$ observed frames $u_{t - k}, ..., u_t$, the partial derivatives appearing in the differential operator $\mathcal{F}$ are concatenated as additional channels to the input. Additionally, in Appendix \ref{sec:appendix-additional-results-repr}, we study the effect of varying the number of temporal conditioning frames $k$ provided to the inverse model. 

\paragraph{Partial observability.} We include a brief study of partial observability in the inverse problem setting. We instantiate partial observability in two primary ways: 1) by introducing noise into the input solution fields (Appendix \ref{sec:noisy_inputs_results}), and 2) removing spatial grid lines to induce a non-uniform discretization (Appendix\ref{sec:non_uniform_grid_results}). 

\subsection{Scaling properties} We conduct data scaling experiments along the number of initial conditions, total number of parameters and training time horizon, and model scaling experiments along the channel width (Appendix~\ref{sec:appendix_scaling}). 

\paragraph{Data scaling.} 
For each data scaling experiment, we evaluate the performance of FNO, ResNet, and scOT to isolate the effects of data quantity. We vary dataset size along three dimensions. In our \emph{Initial Condition Scaling} experiments, we train on a variable number of unique initial conditions per PDE parameter value, namely 20\%, 35\%, 50\%, 75\%, and 100\% of the total number of initial conditions. 
Similarly, in our \emph{Physical Parameter Scaling} experiments, we vary the number of unique PDE parameters sampled from the predefined range, varying over 20\%, 35\%, 50\%, 75\%, and 100\% of the full number of parameters given in Table \ref{tab:pde_datasets}, while using 100\% of available initial conditions.
Lastly, in our \emph{Temporal Horizon Scaling} experiments (Appendix~\ref{sec:appendix-additional-results-scaling}) we train on solution frames within 10\%, 25\%, 50\%, 60\%, and 75 \% of the total time horizon of the simulated trajectories and evaluate on the held-out final 25\%. For temporal horizon scaling, we use 100\% of the PDE parameter values and initial conditions. Additionally, we investigate how the performance of different architectures behaves as data is scaled along \emph{Initial Condition Scaling} and \emph{Physical Parameter Scaling} (Appendix~\ref{sec:appendix-additional-results-scaling}). We note that \emph{initial condition scaling} and \emph{temporal horizon scaling} experiments are not applicable in the Darcy flow case, as it is a time-independent system (Appendix~\ref{sec:darcy_flow_experiments}).

\paragraph{Model Scaling.} For our model scaling experiments, we logarithmically scale the model size across the total number of parameters: 0.5 million, 5 million, and 50 million parameters. We primarily scale the total number of hidden channel dimensions rather than the model depth to avoid the training difficulties associated with deep neural operators~\citep{tran_factorized_2023, koshizuka_expressivity_2024, qin_specboost_2024}.

\section{Results}
\label{sec:Results}
We present selected findings of our evaluation centered on the three fundamental design axes (Section~\ref{sec:methods}).
Our baseline model is an FNO with an encoder-downsampler structure (details in Appendix~\ref{sec:appendix-general-details}). Complete results can be found in Appendix~\ref{sec:appendix-additional-results}.

%\paragraph{Base Model Configuration.} Our baseline model is an FNO with an encoder-downsampler-MLP structure, where $f_\theta(\hat{u}) = \mathcal{M}_\theta \circ \mathcal{D}_\theta \circ \mathcal{E}_\theta(\hat{u}) = \hat{\phi}$. This structure processes spatiotemporal solution fields through the encoder, aggregates spatial information via the downsampler, and outputs the predicted PDE parameters through the MLP.

\subsection{Optimization}
\label{sec:opt_results}
\begin{figure}[htbp!]
    \centering
    \includegraphics[width=\linewidth]{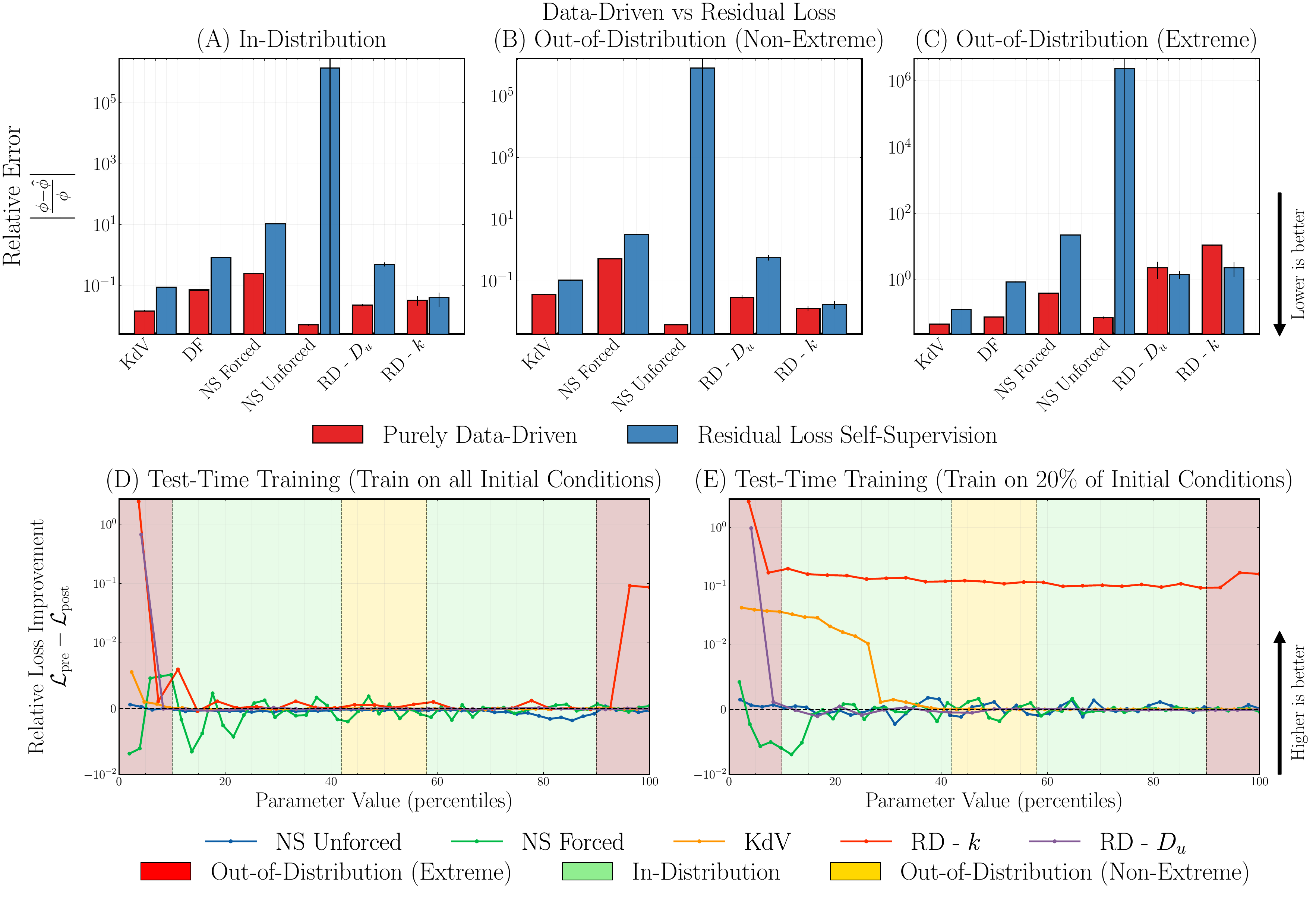}
    \caption{ \textbf{Optimization approaches for NOs in PDE inverse problems}: (\textbf{A}--\textbf{C}) The performance of FNO on purely data-driven supervision versus self-supervision using only the PDE residual. For virtually all systems and evaluation settings, purely data-driven supervision consistently outperforms self-supervision using the PDE residual. As expected, performance degrades on the OOD Non-Extreme and OOD Extreme splits. (\textbf{D}--\textbf{E}) Change in relative error before and after test-time training (TTT) FNO weights, when the initial model is trained on trajectories from all (\textbf{D}) or 20\% (\textbf{E}) of initial conditions in the dataset.
    Larger values indicate greater improvements from TTT. For the KdV and 2D RD systems, TTT yields noticeable performance improvements in the extreme out-of-distribution parameter regime, and the improvement is more pronounced when performing TTT on unseen initial conditions during test time.}
    \label{fig:datavresidual-and-ttt}
\end{figure}

\paragraph{Data supervision vs. self-supervision.}
Figure \ref{fig:datavresidual-and-ttt} A--C compares purely data-driven training (Equation \ref{eq:data_loss}) with purely self-supervised training using the PDE residual (Equation \ref{eq:residual-loss-definition}). On all three test sets, direct supervision consistently outperforms self-supervised training, except in cases where both methods produce relative errors close to $\sim 100\%$ (OOD Extreme split of 2D RD). In Appendix Section \ref{sec:appendix-additional-results-optimization}, we perform additional experiments combining the data loss with the PDE loss, finding that purely data-driven training generally always outperforms any setting with a nonzero coefficient on the PDE loss term. We also perform experiments comparing classical methods such as Newton-CG, L-BFGS-B, and SLSQP against FSNO to better contextualize the results (Appendix Section \ref{sec:appendix-additional-results-optimization}).

%\nc{Previously we also said: "while supervised training provides better overall parameter fitting, it may sometimes lead to less robust extrapolation in dramatically different parameter regimes". @Divyam How was this conclusion drawn? Seems like the 2D RD cases are just bad in the extreme OOD}
%In extreme OOD scenarios, the pattern becomes more nuanced. 
%When the self-supervised model's performance is particularly poor (relative error approaching or exceeding 1.0), direct supervision occasionally performs worse than self-supervision. 
%This suggests that while supervised training provides better overall parameter fitting, it may sometimes lead to less robust extrapolation in dramatically different parameter regimes. 

\paragraph{Test-time training (TTT).}
Figure \ref{fig:datavresidual-and-ttt} D--E shows the results of performing TTT (optimizing Equation \ref{eq:ttt_loss}) after purely data-driven, supervised training. We consider two settings: 1) training on trajectories resulting from all initial conditions in the dataset, and subsequently evaluating on held-out trajectories from these same initial conditions (\Figref{fig:datavresidual-and-ttt}D), and 2) training on 20\% of initial conditions and evaluating on trajectories from unseen initial conditions (\Figref{fig:datavresidual-and-ttt}E). In both figures, we show the improvement in PDE parameter relative loss resulting from TTT across the normalized parameter space (displayed in percentiles). We find that the effectiveness of TTT is generally higher when evaluating on trajectories from unseen ICs and at the extremes of the parameter range, but the degree of improvement is system-dependent. For example, we observe negligible improvements for 2D-NS and 2D-TF, while achieving 10-100\% improvements for 2D-RD. The fact that TTT yields greater improvements in OOD initial condition or parameter settings is consistent with our intuition that incorporating information about the mathematical structure of the problem (via the PDE residual) would aid in OOD generalization. In general, since we observe that TTT never appreciably \textit{hurts} performance, we recommend that it be used for PDE inverse problems. 

% EXPLANATION STUFF WHICH NEEDS TO BE REVISTED BC IT'S NOT GIVING A CLEAR CONCLUSION: We hypothesize that the lack of improvement from TTT in the 2D-NS and 2D-TF systems stems from the fact that despite the spatiotemporal resolution being sufficient for numerical convergence of the ground truth simulations (see Appendix~\ref{sec:appendix-dataset-eqn}), the resolution may be insufficient to render the PDE residual a useful proxy for parameter prediction error (see Appendix -\ref{sec:appendix-ttt-resolution-exp} for a detailed analysis). 

\subsection{Problem representation}
\label{sec:rep_results}

\begin{figure}[htbp!]
    \centering
    \includegraphics[width=\linewidth]{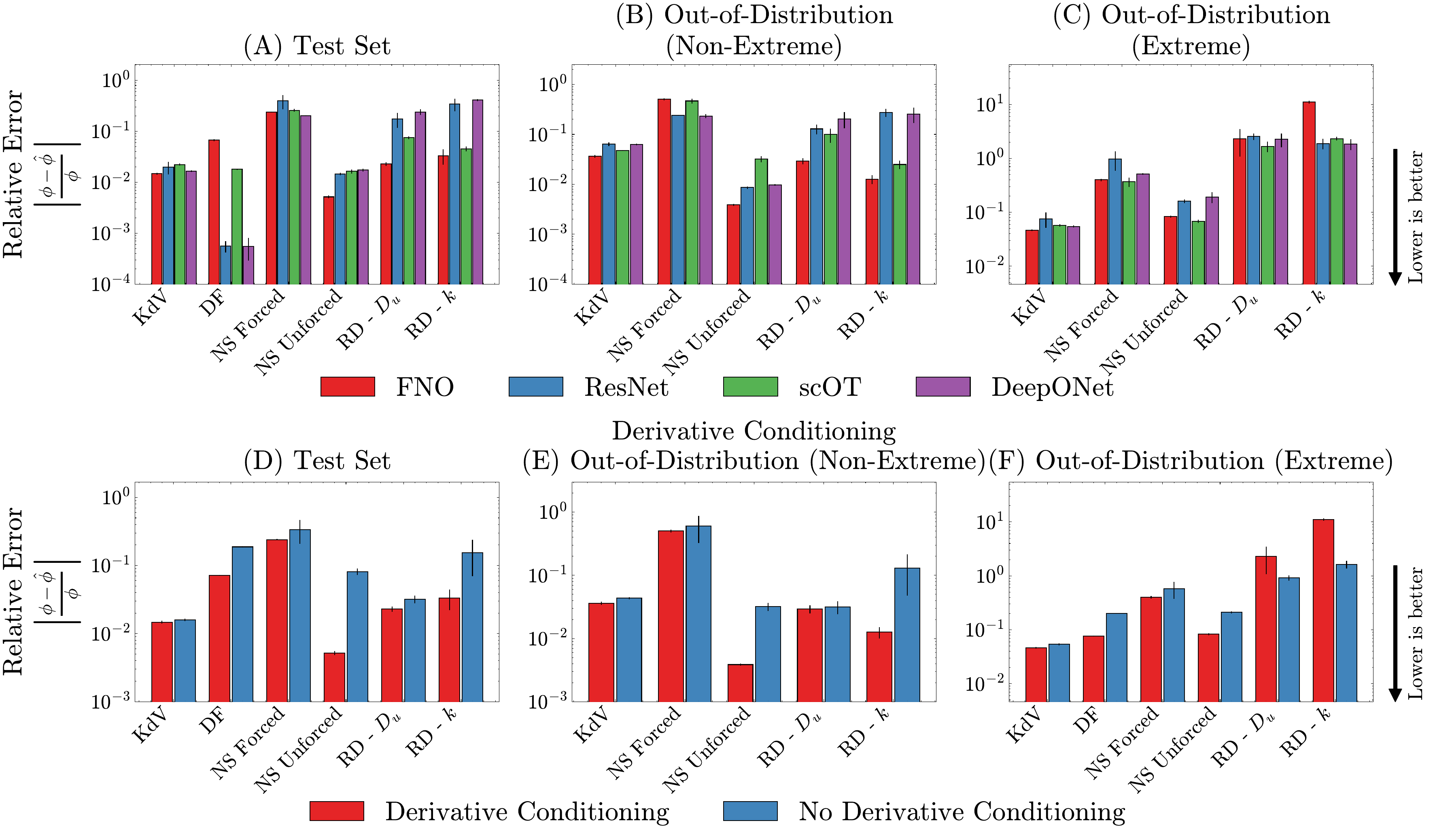}
    \caption{\textbf{Impact of problem representation on PDE inverse problem performance.} (\textbf{A}--\textbf{C}) Comparison of architectural inductive biases (FNO, ResNet, scOT, DeepONet) across evaluation splits. FNO generally outperforms ResNet, scOT and DeepONet on time-dependent PDEs, particularly in In-Distribution and Out-of-Distribution (Non-Extreme) regimes. Results are also shown in tabular form in Tables ~\ref{tab:arch_test}, ~\ref{tab:arch_ood_nonextreme}, and ~\ref{tab:arch_ood_extreme}. (\textbf{D}--\textbf{F}) Effect of partial derivative conditioning on the FNO architecture, \ie~concatenation versus omission of partial derivatives as features. Conditioning the FNO architecture on partial derivatives consistently outperforms the unconditioned variant.}
    \label{fig:architecture-comp-and-derivative-cond}
\end{figure}
We include model architecture and derivative conditioning results here, deferring temporal conditioning results to Appendix~\ref{sec:appendix-additional-results-repr}

\paragraph{Architectural inductive biases.} 
%\ak{Would be very careful about the wording and conclusions drawn here: it's hard to say that one thing (like spectral inductive bias) might be the reason a model is better, when there are many other architectural differences too. Same with in ``Key Insight 3'' in the takeaways}
%\nc{Added some text to reduce the strength of claims made (same for key insight 3)}
We compare four architectural inductive biases in a parameter-controlled manner: convolutional (ResNet), spectral (FNO), global attention (scOT) and branch-trunk decomposition (DeepONet). We provide the architectural comparison results in \Figref{fig:architecture-comp-and-derivative-cond}A-C and as tables in Appendix~\ref{sec:appendix-model-benchmark-results-tables}. We find that DeepONet performs similarly to ResNet in all settings, see (\Figref{fig:architecture-comp-and-derivative-cond}A-C). Since our inverse problems require
the model to make a scalar prediction over all co-location points, it is unsurprising that DeepONet and ResNet achieve similar performance, because they both use a ResNet architecture for the branch network.
For time-dependent PDEs, we find that FNO generally outperforms ResNet and scOT, particularly in OOD settings (\Figref{fig:architecture-comp-and-derivative-cond}A-C).
%While there are differences between the architectures beyond their inductive biases, our results suggest that FNO's performance might be attributable to its spectral inductive bias, which provides a natural representation for modeling continuous solution fields. 
While the architectures differ in factors beyond their inductive biases (e.g., normalization, skip connections, depth), FNO's performance is consistent with the idea that a spectral representation may be well-suited for modeling continuous solution fields.
For Darcy Flow, which is time-independent, ResNet and DeepOnet significantly outperform both FNO and scOT (\Figref{fig:architecture-comp-and-derivative-cond}A-C). 
The lack of time-dependency reduces the problem to a spatial modeling task, making it a good fit for convolutional methods.
% The performance gap between model architectures is most pronounced in challenging evaluation scenarios, indicating that selecting the appropriate inductive bias becomes increasingly important as the difficulty of the task increases.
% Future advancements in modeling techniques change these results, but nonetheless, as it currently stands, spectral methods outperform non-spectral representations on our time-dependent systems.

\paragraph{Appending partial derivatives.}
\Figref{fig:architecture-comp-and-derivative-cond} D--F visualizes the effect of including partial derivatives as additional input features.
Across all PDE systems and evaluation settings, models that receive derivative information generally outperform those that receive only the solution fields, demonstrating improved learning efficiency and generalization. 
While FNO should have the representational capacity to model the spatial derivative operator, we find it helpful to explicitly condition on partial derivatives. In the 2D RD OOD Extreme setting (\Figref{fig:architecture-comp-and-derivative-cond}F), both appending and omitting partial derivative information fail to achieve good performance with relative errors $\sim100\%$.

\begin{figure}[htbp!]
    \centering
    \includegraphics[width=\linewidth]{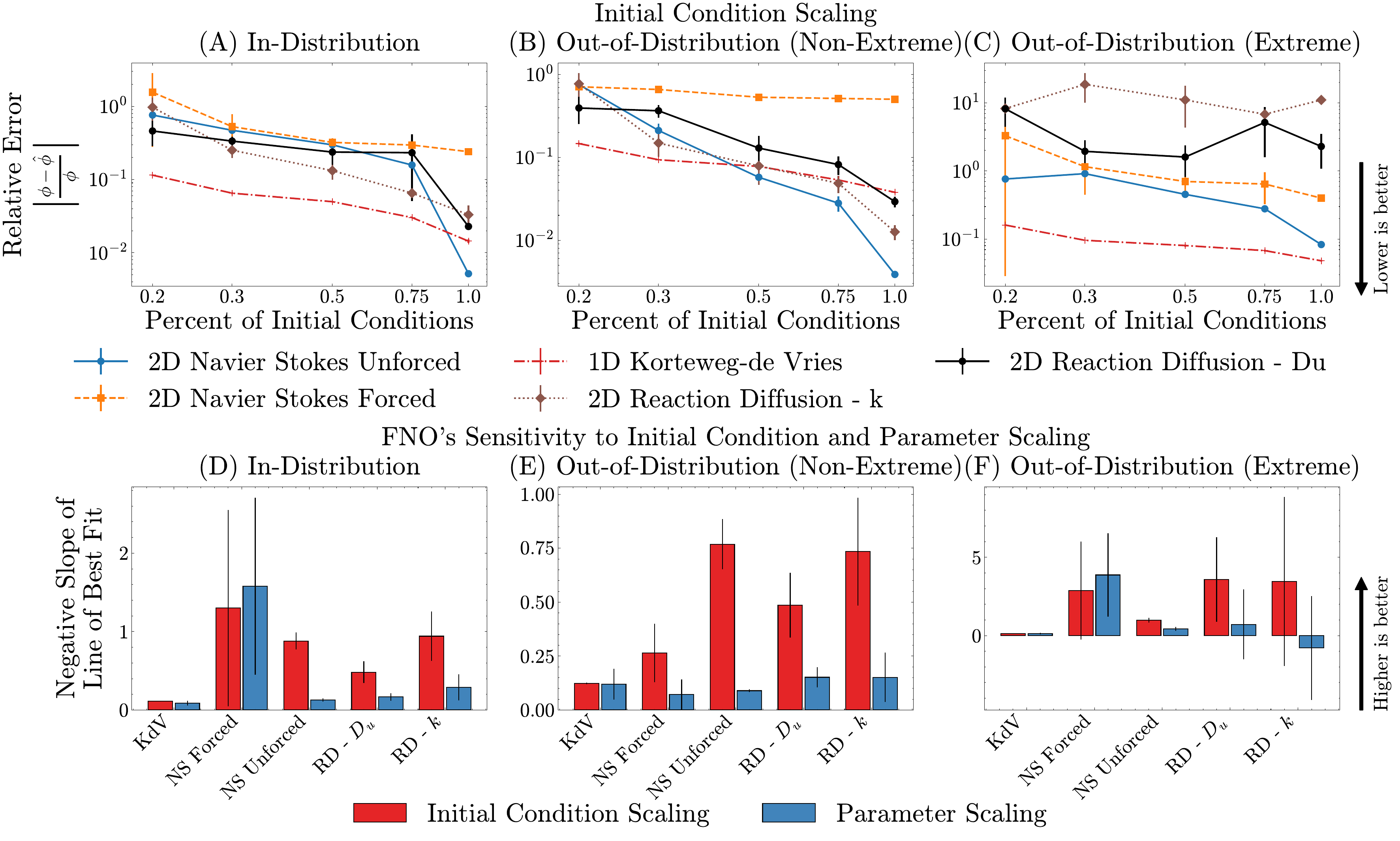}
    \caption{\textbf{Dataset scaling with the FNO architecture:} (\textbf{A}--\textbf{C}) Effect of initial condition scaling on FNO across evaluated systems. Increasing the percent of initial conditions used for training improves performance across all systems and evaluation settings. (\textbf{D}--\textbf{F}) FNO's sensitivity to initial condition scaling (while using 100\% of available PDE parameters) versus scaling the number of generated PDE parameters (while using 100\% of the available initial conditions) as measured by the negative slope of the line of best fit (NLS). The NLS is generated by fitting a line of best fit through each line in A--C, alongside parameter scaling, and taking the negative slope. Larger values indicate desirable faster scaling. FNO improves more quickly as the quantity of initial conditions is increased compared to when the quantity of PDE parameters in the training set is increased. }
    \label{fig:results-scaling}
\end{figure}

\paragraph{Partial observability.} We defer complete partial observability results to Appendix \ref{sec:noisy_inputs_results}
 and \ref{sec:non_uniform_grid_results}, with high-level conclusions stated here. On noisy solution fields, we find that removing partial derivative conditioning improves the robustness of inverse models to degradation, and we find higher sensitivity to Salt-and-Pepper corruption relative to frequency-based corruption. We also find that FNO and ScOT are more robust to non-uniform gridding than ResNet.

\subsection{Scaling}
\label{sec:scaling_results}
We include data scaling results here, deferring model scaling results to Appendix~\ref{sec:appendix-additional-results-scaling}.

\paragraph{Initial condition scaling.}
We find that scaling the number of initial conditions from 20\% to 100\% of all initial conditions leads to a clear reduction in error, with every system improving both ID and OOD Non-Extreme errors (\Figref{fig:results-scaling}A,B). 
The increase in initial conditions yields a modest reduction in relative error for the OOD Extreme case (\Figref{fig:results-scaling}C) and lacks the strong trend seen on other splits.

%\Figref{fig:results-scaling}(D-F) visualize the negative slope of the line of best fit (NLS) from~\Figref{fig:results-scaling}(A-C) with the NLS of scaling the number of generated PDE parameters.
%Spectral neural operators scale faster with an increased amount of initial conditions, indicated by the larger slope magnitudes.
%When compared to CNNs or transformers, spectral neural operators demonstrate higher initial condition efficiency across all systems.

\paragraph{Relative effect of data scaling.}
To discern which type of data scaling is the most impactful for model performance, we compare the relative effect of each type of data scaling on FNO in~\Figref{fig:results-scaling}(D-F).
Increasing the number of initial conditions generally yields larger gains in performance compared to increasing the number of PDE parameters. 
Seeing new physical parameters does not provide the same amount of information as seeing the evolution of new initial conditions with the same physical parameters.
In other words, in our setting, new initial conditions provide a better demonstration of the underlying mapping between solution fields and parameters.

%More parameters generally lead to better performance in all evaluation settings for neural operators for all most all systems (\ref{fig:data-scaling-and-model-scaling}). The improvements are not as high as increasing initial conditions for neural operators. 

% prev:  \subsubsection{Relative effect of data scaling axes of Neural Operators}

%\paragraph{Relative effect of data scaling.}  (\ref{fig:data-scaling-and-model-scaling}). This seems to hold for all evaluation settings.
\section{Conclusion}
\label{sec:conclusion}
We have introduced \inversebench, a comprehensive benchmark dataset for PDE inverse problems spanning diverse physical systems. 
We use \inversebench~to systematically explore and identify best practices for solving PDE inverse problems with neural networks. 
\subsection{Key Insights}
\label{sec:key_insights}
From our investigation of optimization procedures, problem representations, and scaling axes, we distill a set of practical, actionable guidelines for using neural networks to solve inverse problems.

\begin{tcolorbox}[insightbox]
\textbf{Key insight 1:} Practitioners should employ a two-stage training approach when learning neural networks for PDE inverse problems: 1) training with a supervised data loss, and 2) test-time training using the PDE residual to improve generalization to OOD PDE parameters and initial conditions.
\end{tcolorbox}
Self-supervised training on the PDE residual is consistently worse than purely data-driven supervision (Section \ref{sec:opt_results}) across systems and irrespective of the weight $\beta$ used for the residual loss term.
Additionally, given an initial pre-trained network, test-time training (TTT) using the PDE residual leads to mixed results, with larger improvements when the initial condition and PDE parameter are OOD with respect to the training distribution (Section \ref{sec:opt_results}). However, since TTT never appreciably hurts performance, we recommend that it be used as long as the application is not highly compute constrained. 

%Self-supervised training on the PDE residual alone is not as good as supervised training.
%Additionally, the optimal non-zero PDE residual weight for training neural operators with PINNS loss varies significantly by system, with none of them outperforming direct parameter supervision.

%Notably, incorporating unsupervised residual loss during initial training does not consistently improve generalization to out-of-distribution (extreme) scenarios.
%However, test-time fine-tuning with anchor loss does yield performance improvements, even in these challenging out-of-distribution settings.

%\sr{add explanation}
\begin{tcolorbox}[insightbox]
\textbf{Key insight 2:} For time-dependent PDEs, the FNO architecture, with its spectral inductive bias, generally outperforms ResNet and Transformer-style architectures. However, given the vast space of architectural factors, more studies should be conducted to better understand how model inductive biases affect predictions for PDE inverse problems.
\end{tcolorbox}
This, however, does not hold true for time-independent PDEs, like Darcy Flow, where FNO was significantly outperformed by both ResNet and scOT (Section \ref{sec:rep_results}).
Note that our findings are a result of our dataset size and the specific architectures we have chosen.
Beyond inductive biases, there are more differences in architectures including normalization, skip connections, depth etc. 
For the specific systems, dataset regime, and training objective, we found FNO to generally outperform ResNet and scOT.

\begin{tcolorbox}[insightbox]
\textbf{Key insight 3:} Conditioning on the PDE partial derivatives via concatenation consistently improves model performance for the Fourier Neural Operator.
\end{tcolorbox}

While expressive neural networks should be able to implicitly learn the differential operator, explicitly conditioning on PDE partial derivatives improves performance across all systems (Section \ref{sec:rep_results}).

%\sr{would maybe just remove this point and put in appendix}
%\nc{Moved for now - but i'm leaving this as a comment if we want to come back to it}
%The number of conditioning solution frames does not show any universal trend across PDE system and the best performing number of frames is highly system specific. 

\begin{tcolorbox}[insightbox]
\textbf{Key insight 4:} Generating trajectories with more initial conditions per PDE parameter is more beneficial than increasing the number of unique PDE parameters.
\end{tcolorbox}
This suggests which axis to prioritize when generating data on a strict compute budget. Nonetheless, we find that increasing the amount of data along \emph{any generation axis} universally improves performance (Section \ref{sec:scaling_results}).

%\nc{TODO: decide to include data efficiency blurb (see comment)}

%\nc{previously had a blurb about data efficiency trends (as NO being best) doesn't hold when scaling pde param count - I removed since it felt like it might detrack from the main message?}
%Furthermore, when compared to CNNs and transformers, Neural Operators generally benefit more from increased initial conditions (i.e are more data efficient). 
%\nc{I don't know what experiment or result this is referrign to. Is this a model scaling result?}
%However, this pattern becomes inconsistent when scaling training data across the total number of parameters, with improvements becoming highly system-specific.

%\begin{tcolorbox}[colback=cyan!15!white, boxrule=0.1mm, arc=1mm]
%\textbf{Key Insight 5:} 
%\end{tcolorbox}
%\sr{add explanation}

\subsection{Future Work}
Future work could extend our benchmarking study in several important directions. One avenue is to consider more PDEs beyond Darcy Flow with coefficients that vary freely in space, testing whether ML models can capture heterogeneous parameter fields rather than just scalar parameters ~\citep{leforestier_comparison_1991}. 
Another is to investigate inverse problems under sparse, irregularly sampled, or noisy measurements, which would better reflect realistic experimental conditions~\citep{stuart_inverse_2010}.
While we investigate Butterworth corruption as well as salt and pepper noise in Appendix~\ref{sec:noisy_inputs_results}, noise models are unique to application areas. Future work should explore application-specific noise models and their effects on performance.
In these settings, it would be interesting to predict a distribution of PDE parameters with rigorous uncertainty quantification, rather than single point estimates of PDE parameters as done in the present work. Beyond the 2D settings explored here, the inclusion of 3D benchmark datasets such as Navier–Stokes turbulence would further challenge models on higher-dimensional dynamics while also probing scalability under memory and compute constraints~\citep{du_eddyformer_2025}. Our dataset also currently only contains systems with a uniform (square or linear) geometry, while many realistic systems have more complex topologies (e.g., spherical). We plan to include such systems in future work. 
Additionally, our current benchmark evaluates grid-based architectures; extending the comparison to point-cloud-based methods designed for irregular meshes (e.g., Transolver~\citep{wu_transolver_2024}, UPT~\citep{alkin_universal_2024}) would require rethinking sections of our evaluation protocol, but is a valuable direction for future work. We simulate what this might look like for partially observed spatial fields in Appendix~\ref{sec:non_uniform_grid_results} but leave a more thorough treatment to future work.
Finally, incorporating solver-in-the-loop losses (e.g., running a PDE solver with the predicted parameters and enforcing consistency with a reference trajectory) alongside data and PDE residual losses could enforce stronger physics consistency and improve generalization across regimes and boundary conditions~\citep{um_solver_loop_2020}. 
We believe \inversebench~will act as a solid foundation for such further advances in ML approaches to PDE inverse problems.

% \paragraph{Generalized training recipe}

% Based on our findings, we propose a practical recipe for training neural operators in time-dependent PDE inverse problems with limited computational resources. This approach balances the trade-offs practitioners face when deploying these models in real-world settings. Our recommendations optimize the allocation of computational budget for training data generation, model architecture selection, and training methodology to achieve reliable performance across diverse system and evaluation settings. 

% % Dataset generation:

% % 1) For a fixed compute budget set up more time generating initial conditions over differenet parameter conditions for the PDE system of interest.

% % Pre-Training:

% % 1) Set up the dataloader such the partial derivative terms of the PDE are computed and then appended to the input solution fields.

% % 2) train the neural operator with parameter loss.

% % 3) as part of hyper-parameter tuning: vary the number of input solution frames over multiple training iterations and select the best performing number based off of the validation set performance 

% % Test-Time finetuning:

% % 1) Equally weight the anchor loss and perform self-supervised fine-tuning on the test-time solution fields for several steps.

%\subsubsection*{Author Contributions}

%\subsubsection*{Acknowledgments}

\bibliography{main}

@inproceedings{takamoto_pdebench_2022,
	title = {{PDEBENCH}: {An} {Extensive} {Benchmark} for {Scientific} {Machine} {Learning}},
	url = {https://openreview.net/forum?id=dh_MkX0QfrK},
	booktitle = {Proceedings of the 36th {Conference} on {Neural} {Information} {Processing} {Systems} ({NeurIPS})},
	author = {Takamoto, Makoto and Praditia, Timothy and Leiteritz, Raphael and MacKinlay, Dan and Alesiani, Francesco and Pflüger, Dirk and Niepert, Mathias},
	year = {2022},
}

@article{gupta_towards_2022,
	title = {Towards {Multi}-spatiotemporal-scale {Generalized} {PDE} {Modeling}},
	journal = {arXiv preprint arXiv:2209.15616},
	author = {Gupta, Jayesh K and Brandstetter, Johannes},
	year = {2022},
}

@inproceedings{ohana_well_2024,
	title = {The {Well}: {A} large-scale dataset collection for scientific machine learning},
	booktitle = {38th {Conference} on {Neural} {Information} {Processing} {Systems} ({NeurIPS} 2024) {Track} on {Datasets} and {Benchmarks}},
	author = {Ohana, Robin and {others}},
	year = {2024},
}

@article{hao_pinnacle_2023,
	title = {{PINNacle}: {A} {Comprehensive} {Benchmark} of {Physics}-{Informed} {Neural} {Networks} for {Solving} {PDEs}},
	journal = {arXiv preprint arXiv:2306.08827},
	author = {Hao, Zhongkai and Yao, Jiachen and Su, Chang and Su, Hang and Wang, Ziao and Lu, Fanzhi and Xia, Zeyu and Zhang, Yichi and Liu, Songming and Lu, Lu and {others}},
	year = {2023},
}

@misc{bhan_pde_2024,
	title = {{PDE} {Control} {Gym}: {A} {Benchmark} for {Data}-{Driven} {Boundary} {Control} of {Partial} {Differential} {Equations}},
	author = {Bhan, Luke and Bian, Yuexin and Krstic, Miroslav and Shi, Yuanyuan},
	year = {2024},
	note = {\_eprint: 2405.11401},
}

@article{kohl_benchmarking_2023,
	title = {Benchmarking {Autoregressive} {Conditional} {Diffusion} {Models} for {Turbulent} {Flow} {Simulation}},
	url = {https://doi.org/10.48550/arXiv.2309.01745},
	doi = {10.48550/arXiv.2309.01745},
	journal = {arXiv},
	author = {Kohl, Georg and Chen, Li-Wei and Thuerey, Nils},
	year = {2023},
	note = {Publisher: arXiv
\_eprint: 2309.01745},
}

@inproceedings{herde_poseidon_2024,
	title = {Poseidon: {Efficient} {Foundation} {Models} for {PDEs}},
	volume = {37},
	url = {https://proceedings.neurips.cc/paper_files/paper/2024/file/84e1b1ec17bb11c57234e96433022a9a-Paper-Conference.pdf},
	booktitle = {Advances in {Neural} {Information} {Processing} {Systems}},
	publisher = {Curran Associates, Inc.},
	author = {Herde, Maximilian and Raonić, Bogdan and Rohner, Tobias and Käppeli, Roger and Molinaro, Roberto and de Bézenac, Emmanuel and Mishra, Siddhartha},
	editor = {Globerson, A. and Mackey, L. and Belgrave, D. and Fan, A. and Paquet, U. and Tomczak, J. and Zhang, C.},
	year = {2024},
	pages = {72525--72624},
}

@inproceedings{hassan_bubbleml_2023,
	title = {{BubbleML}: {A} {Multi}-{Physics} {Dataset} and {Benchmarks} for {Machine} {Learning}},
	url = {https://openreview.net/forum?id=0Wmglu8zak},
	booktitle = {Advances in {Neural} {Information} {Processing} {Systems}},
	author = {Hassan, Sheikh Md Shakeel and Feeney, Arthur and Dhruv, Akash and Kim, Jihoon and Suh, Youngjoon and Ryu, Jaiyoung and Won, Yoonjin and Chandramowlishwaran, Aparna},
	year = {2023},
}

@article{toshev_lagrangebench_2024,
	title = {Lagrangebench: {A} lagrangian fluid mechanics benchmarking suite},
	volume = {36},
	journal = {Advances in Neural Information Processing Systems},
	author = {Toshev, Artur and Galletti, Gianluca and Fritz, Fabian and Adami, Stefan and Adams, Nikolaus},
	year = {2024},
}

@article{li_physics-informed_2024,
	title = {Physics-{Informed} {Neural} {Operator} for {Learning} {Partial} {Differential} {Equations}},
	volume = {1},
	url = {https://doi.org/10.1145/3648506},
	doi = {10.1145/3648506},
	number = {3},
	journal = {ACM / IMS J. Data Sci.},
	author = {Li, Zongyi and Zheng, Hongkai and Kovachki, Nikola and Jin, David and Chen, Haoxuan and Liu, Burigede and Azizzadenesheli, Kamyar and Anandkumar, Anima},
	month = may,
	year = {2024},
	note = {Place: New York, NY, USA
Publisher: Association for Computing Machinery},
}

@inproceedings{li_fourier_2021,
	title = {Fourier {Neural} {Operator} for {Parametric} {Partial} {Differential} {Equations}},
	url = {https://openreview.net/forum?id=c8P9NQVtmnO},
	booktitle = {International {Conference} on {Learning} {Representations}},
	author = {Li, Zongyi and Kovachki, Nikola Borislavov and Azizzadenesheli, Kamyar and liu, Burigede and Bhattacharya, Kaushik and Stuart, Andrew and Anandkumar, Anima},
	year = {2021},
}

@inproceedings{he_deep_2016,
	series = {{CVPR} '16},
	title = {Deep {Residual} {Learning} for {Image} {Recognition}},
	url = {http://ieeexplore.ieee.org/document/7780459},
	doi = {10.1109/CVPR.2016.90},
	booktitle = {Proceedings of 2016 {IEEE} {Conference} on {Computer} {Vision} and {Pattern} {Recognition}},
	publisher = {IEEE},
	author = {He, Kaiming and Zhang, Xiangyu and Ren, Shaoqing and Sun, Jian},
	month = jun,
	year = {2016},
	note = {ISSN: 1063-6919
Place: Las Vegas, NV, USA},
	pages = {770--778},
}

@inproceedings{molinaro_neural_2023,
	series = {Proceedings of {Machine} {Learning} {Research}},
	title = {Neural {Inverse} {Operators} for {Solving} {PDE} {Inverse} {Problems}},
	volume = {202},
	url = {https://proceedings.mlr.press/v202/molinaro23a.html},
	booktitle = {Proceedings of the 40th {International} {Conference} on {Machine} {Learning}},
	publisher = {PMLR},
	author = {Molinaro, Roberto and Yang, Yunan and Engquist, Björn and Mishra, Siddhartha},
	editor = {Krause, Andreas and Brunskill, Emma and Cho, Kyunghyun and Engelhardt, Barbara and Sabato, Sivan and Scarlett, Jonathan},
	month = jul,
	year = {2023},
	pages = {25105--25139},
}

@misc{jiao_solving_2024,
	title = {Solving forward and inverse {PDE} problems on unknown manifolds via physics-informed neural operators},
	url = {https://arxiv.org/abs/2407.05477},
	author = {Jiao, Anran and Yan, Qile and Harlim, Jhn and Lu, Lu},
	year = {2024},
	note = {\_eprint: 2407.05477},
}

@inproceedings{wang_latent_2024,
	title = {Latent {Neural} {Operator} for {Solving} {Forward} and {Inverse} {PDE} {Problems}},
	volume = {37},
	url = {https://proceedings.neurips.cc/paper_files/paper/2024/file/39f6d5c2e310a5a629dcfc4d517aa0d1-Paper-Conference.pdf},
	booktitle = {Advances in {Neural} {Information} {Processing} {Systems}},
	publisher = {Curran Associates, Inc.},
	author = {Wang, Tian and Wang, Chuang},
	editor = {Globerson, A. and Mackey, L. and Belgrave, D. and Fan, A. and Paquet, U. and Tomczak, J. and Zhang, C.},
	year = {2024},
	pages = {33085--33107},
}

@inproceedings{cho_physics-informed_2025,
	title = {Physics-{Informed} {Deep} {Inverse} {Operator} {Networks} for {Solving} {PDE} {Inverse} {Problems}},
	url = {https://openreview.net/forum?id=0FxnSZJPmh},
	booktitle = {The {Thirteenth} {International} {Conference} on {Learning} {Representations}},
	author = {Cho, Sung Woong and Son, Hwijae},
	year = {2025},
}

@article{lu_comprehensive_2022,
	title = {A comprehensive and fair comparison of two neural operators (with practical extensions) based on {FAIR} data},
	volume = {393},
	issn = {0045-7825},
	url = {https://www.sciencedirect.com/science/article/pii/S0045782522001207},
	doi = {https://doi.org/10.1016/j.cma.2022.114778},
	journal = {Computer Methods in Applied Mechanics and Engineering},
	author = {Lu, Lu and Meng, Xuhui and Cai, Shengze and Mao, Zhiping and Goswami, Somdatta and Zhang, Zhongqiang and Karniadakis, George Em},
	year = {2022},
	pages = {114778},
}

@inproceedings{mackinlay_model_2021,
	title = {Model {Inversion} for {Spatio}-temporal {Processes} using the {Fourier} {Neural} {Operator}},
	booktitle = {Neurips {Workshop} on {Machine} {Learning} for the {Physical} {Sciences}},
	author = {MacKinlay, Dan and Pagendam, Dan and Kuhnert, Petra M and Cui, Tao and Robertson, David and Janardhanan, Sreekanth},
	year = {2021},
	pages = {7},
}

@book{evans_partial_2022,
	edition = {2nd},
	series = {Graduate {Studies} in {Mathematics}},
	title = {Partial {Differential} {Equations}},
	volume = {19},
	isbn = {978-0-8218-4974-3},
	publisher = {American Mathematical Society},
	author = {Evans, Lawrence C.},
	year = {2022},
}

@article{krishnapriyan2021characterizing,
  title={Characterizing possible failure modes in physics-informed neural networks},
  author={Krishnapriyan, Aditi and Gholami, Amir and Zhe, Shandian and Kirby, Robert and Mahoney, Michael W},
  journal={Advances in neural information processing systems},
  volume={34},
  pages={26548--26560},
  year={2021}
}

@article{du_neural_2024,
	title = {Neural {Spectral} {Methods}: {Self}-supervised learning in the spectral domain},
	journal = {The Twelfth International Conference on Learning Representations},
	author = {Du, Yiheng and Chalapathi, Nithin and Krishnapriyan, Aditi},
	year = {2024},
}

@book{canuto_spectral_2007,
	series = {Scientific {Computation}},
	title = {Spectral {Methods}: {Evolution} to {Complex} {Geometries} and {Applications} to {Fluid} {Dynamics}},
	isbn = {978-3-540-30728-0},
	url = {https://books.google.com/books?id=iDckv0W52cQC},
	publisher = {Springer Berlin Heidelberg},
	author = {Canuto, C. and Hussaini, M.Y. and Quarteroni, A. and Zang, T.A.},
	year = {2007},
	lccn = {2007924823},
}

@article{dresdner_learning_2023,
	title = {Learning to correct spectral methods for simulating turbulent flows},
	issn = {2835-8856},
	url = {https://openreview.net/forum?id=wNBARGxoJn},
	journal = {Transactions on Machine Learning Research},
	author = {Dresdner, Gideon and Kochkov, Dmitrii and Norgaard, Peter Christian and Zepeda-Nunez, Leonardo and Smith, Jamie and Brenner, Michael and Hoyer, Stephan},
	year = {2023},
}

@article{dormand_family_1980,
	title = {A family of embedded {Runge}-{Kutta} formulae},
	volume = {6},
	issn = {0377-0427},
	url = {https://www.sciencedirect.com/science/article/pii/0771050X80900133},
	doi = {https://doi.org/10.1016/0771-050X(80)90013-3},
	number = {1},
	journal = {Journal of Computational and Applied Mathematics},
	author = {Dormand, J. R. and Prince, P. J.},
	year = {1980},
	pages = {19--26},
}

@article{hairer_stiff_1999,
	title = {Stiff differential equations solved by {Radau} methods},
	volume = {111},
	issn = {0377-0427},
	url = {https://www.sciencedirect.com/science/article/pii/S037704279900134X},
	doi = {https://doi.org/10.1016/S0377-0427(99)00134-X},
	number = {1},
	journal = {Journal of Computational and Applied Mathematics},
	author = {Hairer, Ernst and Wanner, Gerhard},
	year = {1999},
	pages = {93--111},
}

@inproceedings{brandstetter_lie_2022,
	series = {Proceedings of {Machine} {Learning} {Research}},
	title = {Lie {Point} {Symmetry} {Data} {Augmentation} for {Neural} {PDE} {Solvers}},
	volume = {162},
	url = {https://proceedings.mlr.press/v162/brandstetter22a.html},
	booktitle = {Proceedings of the 39th {International} {Conference} on {Machine} {Learning}},
	publisher = {PMLR},
	author = {Brandstetter, Johannes and Welling, Max and Worrall, Daniel E},
	editor = {Chaudhuri, Kamalika and Jegelka, Stefanie and Song, Le and Szepesvari, Csaba and Niu, Gang and Sabato, Sivan},
	month = jul,
	year = {2022},
	pages = {2241--2256},
}

@article{kovachki_neural_2023,
	title = {Neural operator: learning maps between function spaces with applications to {PDEs}},
	volume = {24},
	issn = {1532-4435},
	number = {1},
	journal = {Journal of Machine Learning Research},
	author = {Kovachki, Nikola and Li, Zongyi and Liu, Burigede and Azizzadenesheli, Kamyar and Bhattacharya, Kaushik and Stuart, Andrew and Anandkumar, Anima},
	month = jan,
	year = {2023},
	note = {Publisher: JMLR.org},
}

@inproceedings{liu_swin_2022,
	title = {Swin {Transformer} {V2}: {Scaling} {Up} {Capacity} and {Resolution}},
	doi = {10.1109/CVPR52688.2022.01170},
	booktitle = {2022 {IEEE}/{CVF} {Conference} on {Computer} {Vision} and {Pattern} {Recognition} ({CVPR})},
	author = {Liu, Ze and Hu, Han and Lin, Yutong and Yao, Zhuliang and Xie, Zhenda and Wei, Yixuan and Ning, Jia and Cao, Yue and Zhang, Zheng and Dong, Li and Wei, Furu and Guo, Baining},
	year = {2022},
	pages = {11999--12009},
}

@article{kochkov_machine_2021,
	title = {Machine learning–accelerated computational fluid dynamics},
	volume = {118},
	url = {https://www.pnas.org/doi/abs/10.1073/pnas.2101784118},
	doi = {10.1073/pnas.2101784118},
	number = {21},
	journal = {Proceedings of the National Academy of Sciences},
	author = {Kochkov, Dmitrii and Smith, Jamie A. and Alieva, Ayya and Wang, Qing and Brenner, Michael P. and Hoyer, Stephan},
	year = {2021},
	note = {\_eprint: https://www.pnas.org/doi/pdf/10.1073/pnas.2101784118},
	pages = {e2101784118},
}

@article{virtanen_scipy_2020,
	title = {{SciPy} 1.0: {Fundamental} {Algorithms} for {Scientific} {Computing} in {Python}},
	volume = {17},
	doi = {10.1038/s41592-019-0686-2},
	journal = {Nature Methods},
	author = {Virtanen, Pauli and Gommers, Ralf and Oliphant, Travis E. and Haberland, Matt and Reddy, Tyler and Cournapeau, David and Burovski, Evgeni and Peterson, Pearu and Weckesser, Warren and Bright, Jonathan and van der Walt, Stéfan J. and Brett, Matthew and Wilson, Joshua and Millman, K. Jarrod and Mayorov, Nikolay and Nelson, Andrew R. J. and Jones, Eric and Kern, Robert and Larson, Eric and Carey, C J and Polat, Ilhan and Feng, Yu and Moore, Eric W. and VanderPlas, Jake and Laxalde, Denis and Perktold, Josef and Cimrman, Robert and Henriksen, Ian and Quintero, E. A. and Harris, Charles R. and Archibald, Anne M. and Ribeiro, Antônio H. and Pedregosa, Fabian and van Mulbregt, Paul and {SciPy 1.0 Contributors}},
	year = {2020},
	pages = {261--272},
}

@inproceedings{tran_factorized_2023,
	title = {Factorized {Fourier} {Neural} {Operators}},
	url = {https://openreview.net/forum?id=tmIiMPl4IPa},
	booktitle = {The {Eleventh} {International} {Conference} on {Learning} {Representations}},
	author = {Tran, Alasdair and Mathews, Alexander and Xie, Lexing and Ong, Cheng Soon},
	year = {2023},
}

@inproceedings{hao_gnot_2023,
	series = {{ICML}'23},
	title = {{GNOT}: a general neural operator transformer for operator learning},
	booktitle = {Proceedings of the 40th {International} {Conference} on {Machine} {Learning}},
	publisher = {JMLR.org},
	author = {Hao, Zhongkai and Wang, Zhengyi and Su, Hang and Ying, Chengyang and Dong, Yinpeng and Liu, Songming and Cheng, Ze and Song, Jian and Zhu, Jun},
	year = {2023},
	note = {Place: Honolulu, Hawaii, USA},
}

@inproceedings{cao_choose_2021,
	title = {Choose a {Transformer}: {Fourier} or {Galerkin}},
	url = {https://openreview.net/forum?id=ssohLcmn4-r},
	booktitle = {Advances in {Neural} {Information} {Processing} {Systems}},
	author = {Cao, Shuhao},
	editor = {Beygelzimer, A. and Dauphin, Y. and Liang, P. and Vaughan, J. Wortman},
	year = {2021},
}

@inproceedings{alkin_universal_2024,
	title = {Universal {Physics} {Transformers}: {A} {Framework} {For} {Efficiently} {Scaling} {Neural} {Operators}},
	url = {https://openreview.net/forum?id=oUXiNX5KRm},
	booktitle = {The {Thirty}-eighth {Annual} {Conference} on {Neural} {Information} {Processing} {Systems}},
	author = {Alkin, Benedikt and Fürst, Andreas and Schmid, Simon Lucas and Gruber, Lukas and Holzleitner, Markus and Brandstetter, Johannes},
	year = {2024},
}

@misc{wu_transolver_2024,
	title = {Transolver: {A} {Fast} {Transformer} {Solver} for {PDEs} on {General} {Geometries}},
	shorttitle = {Transolver},
	url = {http://arxiv.org/abs/2402.02366},
	doi = {10.48550/arXiv.2402.02366},
	urldate = {2024-06-04},
	publisher = {arXiv},
	author = {Wu, Haixu and Luo, Huakun and Wang, Haowen and Wang, Jianmin and Long, Mingsheng},
	month = jun,
	year = {2024},
	note = {arXiv:2402.02366 [cs, math]},
}

@inproceedings{wolf_transformers_2020,
	address = {Online},
	title = {Transformers: {State}-of-the-{Art} {Natural} {Language} {Processing}},
	url = {https://aclanthology.org/2020.emnlp-demos.6/},
	doi = {10.18653/v1/2020.emnlp-demos.6},
	booktitle = {Proceedings of the 2020 {Conference} on {Empirical} {Methods} in {Natural} {Language} {Processing}: {System} {Demonstrations}},
	publisher = {Association for Computational Linguistics},
	author = {Wolf, Thomas and Debut, Lysandre and Sanh, Victor and Chaumond, Julien and Delangue, Clement and Moi, Anthony and Cistac, Pierric and Rault, Tim and Louf, Remi and Funtowicz, Morgan and Davison, Joe and Shleifer, Sam and von Platen, Patrick and Ma, Clara and Jernite, Yacine and Plu, Julien and Xu, Canwen and Le Scao, Teven and Gugger, Sylvain and Drame, Mariama and Lhoest, Quentin and Rush, Alexander},
	editor = {Liu, Qun and Schlangen, David},
	month = oct,
	year = {2020},
	pages = {38--45},
}

@incollection{borckmans_turing_1995,
	address = {Dordrecht},
	title = {Turing {Bifurcations} and {Pattern} {Selection}},
	isbn = {978-94-011-1156-0},
	url = {https://doi.org/10.1007/978-94-011-1156-0_10},
	booktitle = {Chemical {Waves} and {Patterns}},
	publisher = {Springer Netherlands},
	author = {Borckmans, P. and Dewel, G. and De Wit, A. and Walgraef, D.},
	editor = {Kapral, Raymond and Showalter, Kenneth},
	year = {1995},
	doi = {10.1007/978-94-011-1156-0_10},
	pages = {323--363},
}

@article{klaasen_stationary_1984,
	title = {Stationary {Wave} {Solutions} of a {System} of {Reaction}-{Diffusion} {Equations} {Derived} from the {FitzHugh}–{Nagumo} {Equations}},
	volume = {44},
	url = {https://doi.org/10.1137/0144008},
	doi = {10.1137/0144008},
	number = {1},
	journal = {SIAM Journal on Applied Mathematics},
	author = {Klaasen, Gene A. and Troy, William C.},
	year = {1984},
	note = {\_eprint: https://doi.org/10.1137/0144008},
	pages = {96--110},
}

@book{isakov_inverse_2017,
	edition = {3},
	series = {Applied {Mathematical} {Science}},
	title = {Inverse {Problems} for {Partial} {Differential} {Equations}},
	volume = {127},
	isbn = {978-3-319-51658-5},
	url = {https://doi.org/10.1007/978-3-319-51658-5},
	publisher = {Springer Cham},
	author = {Isakov, Victor},
	year = {2017},
}

@book{tarantola_inverse_2005,
	title = {Inverse {Problem} {Theory} and {Methods} for {Model} {Parameter} {Estimation}},
	url = {https://epubs.siam.org/doi/abs/10.1137/1.9780898717921},
	publisher = {Society for Industrial and Applied Mathematics},
	author = {Tarantola, Albert},
	year = {2005},
	doi = {10.1137/1.9780898717921},
	note = {\_eprint: https://epubs.siam.org/doi/pdf/10.1137/1.9780898717921},
}

@article{molesky_inverse_2018,
	title = {Inverse design in nanophotonics},
	volume = {12},
	issn = {1749-4893},
	url = {https://doi.org/10.1038/s41566-018-0246-9},
	doi = {10.1038/s41566-018-0246-9},
	number = {11},
	journal = {Nature Photonics},
	author = {Molesky, Sean and Lin, Zin and Piggott, Alexander Y. and Jin, Weiliang and Vucković, Jelena and Rodriguez, Alejandro W.},
	month = nov,
	year = {2018},
	pages = {659--670},
}

@book{vlaardingerbroek_magnetic_2013,
	edition = {3},
	title = {Magnetic {Resonance} {Imaging}: {Theory} and {Practice}},
	isbn = {978-3-662-05252-5},
	url = {https://doi.org/10.1007/978-3-662-05252-5},
	publisher = {Springer Berlin, Heidelberg},
	author = {Vlaardingerbroek, Marinus T. and Boer, Jacques A.},
	year = {2013},
}

@article{karnakov_solving_2024,
	title = {Solving inverse problems in physics by optimizing a discrete loss: {Fast} and accurate learning without neural networks},
	volume = {3},
	issn = {2752-6542},
	url = {https://doi.org/10.1093/pnasnexus/pgae005},
	doi = {10.1093/pnasnexus/pgae005},
	number = {1},
	journal = {PNAS Nexus},
	author = {Karnakov, Petr and Litvinov, Sergey and Koumoutsakos, Petros},
	month = jan,
	year = {2024},
	note = {\_eprint: https://academic.oup.com/pnasnexus/article-pdf/3/1/pgae005/56347037/pgae005.pdf},
	pages = {pgae005},
}

@article{rahman_u-no_2023,
	title = {U-{NO}: {U}-shaped {Neural} {Operators}},
	issn = {2835-8856},
	url = {https://openreview.net/forum?id=j3oQF9coJd},
	journal = {Transactions on Machine Learning Research},
	author = {Rahman, Md Ashiqur and Ross, Zachary E. and Azizzadenesheli, Kamyar},
	year = {2023},
}

@article{zabusky_interaction_1965,
	title = {Interaction of "{Solitons}" in a {Collisionless} {Plasma} and the {Recurrence} of {Initial} {States}},
	volume = {15},
	url = {https://link.aps.org/doi/10.1103/PhysRevLett.15.240},
	doi = {10.1103/PhysRevLett.15.240},
	number = {6},
	journal = {Phys. Rev. Lett.},
	author = {Zabusky, N. J. and Kruskal, M. D.},
	month = aug,
	year = {1965},
	note = {Publisher: American Physical Society},
	pages = {240--243},
}

@book{polyanin_handbook_2012,
	edition = {2nd},
	title = {Handbook of {Nonlinear} {Partial} {Differential} {Equations}},
	isbn = {1-58488-355-3},
	publisher = {Chapman \& Hall/CRC Press, Boca Raton-London-New York},
	author = {Polyanin, Andrei D. and Zaitsev, Valentin F.},
	year = {2012},
}

@software{jax2018github,
  author = {James Bradbury and Roy Frostig and Peter Hawkins and Matthew James Johnson and Chris Leary and Dougal Maclaurin and George Necula and Adam Paszke and Jake Vander{P}las and Skye Wanderman-{M}ilne and Qiao Zhang},
  title = {{JAX}: composable transformations of {P}ython+{N}um{P}y programs},
  url = {http://github.com/jax-ml/jax},
  version = {0.3.13},
  year = {2018},
}

@article{bar-sinai_learning_2019,
	title = {Learning data-driven discretizations for partial differential equations},
	volume = {116},
	url = {https://www.pnas.org/doi/abs/10.1073/pnas.1814058116},
	doi = {10.1073/pnas.1814058116},
	number = {31},
	journal = {Proceedings of the National Academy of Sciences},
	author = {Bar-Sinai, Yohai and Hoyer, Stephan and Hickey, Jason and Brenner, Michael P.},
	year = {2019},
	note = {\_eprint: https://www.pnas.org/doi/pdf/10.1073/pnas.1814058116},
	pages = {15344--15349},
}

@inproceedings{
blanke2024interpretable,
title={Interpretable Meta-Learning of Physical Systems},
author={Matthieu Blanke and Marc Lelarge},
booktitle={The Twelfth International Conference on Learning Representations},
year={2024},
url={https://openreview.net/forum?id=nnicaG5xiH}
}

@InProceedings{pmlr-v162-kirchmeyer22a,
  title = 	 {Generalizing to New Physical Systems via Context-Informed Dynamics Model},
  author =       {Kirchmeyer, Matthieu and Yin, Yuan and Dona, Jeremie and Baskiotis, Nicolas and Rakotomamonjy, Alain and Gallinari, Patrick},
  booktitle = 	 {Proceedings of the 39th International Conference on Machine Learning},
  pages = 	 {11283--11301},
  year = 	 {2022},
  editor = 	 {Chaudhuri, Kamalika and Jegelka, Stefanie and Song, Le and Szepesvari, Csaba and Niu, Gang and Sabato, Sivan},
  volume = 	 {162},
  series = 	 {Proceedings of Machine Learning Research},
  month = 	 {17--23 Jul},
  publisher =    {PMLR},
  url = 	 {https://proceedings.mlr.press/v162/kirchmeyer22a.html}
}

@inproceedings{
koupa2024boosting,
title={Boosting Generalization in Parametric {PDE} Neural Solvers through Adaptive Conditioning},
author={Armand Kassa{\"\i} Koupa{\"\i} and Jorge Mifsut Benet and Yuan Yin and Jean-No{\"e}l Vittaut and Patrick Gallinari},
booktitle={The Thirty-eighth Annual Conference on Neural Information Processing Systems},
year={2024},
url={https://openreview.net/forum?id=GuY0zB2xVU}
}

@incollection{ljung1998system,
  title={System identification},
  author={Ljung, Lennart},
  booktitle={Signal analysis and prediction},
  pages={163--173},
  year={1998},
  publisher={Springer}
}

@article{aastrom1971system,
  title={System identification—a survey},
  author={{\AA}str{\"o}m, Karl Johan and Eykhoff, Peter},
  journal={Automatica},
  volume={7},
  number={2},
  pages={123--162},
  year={1971},
  publisher={Elsevier}
}

@article{budanur2022scale,
  title = {Scale-dependent error growth in Navier-Stokes simulations},
  author = {Budanur, Nazmi Burak and Kantz, Holger},
  journal = {Phys. Rev. E},
  volume = {106},
  issue = {4},
  pages = {045102},
  numpages = {7},
  year = {2022},
  month = {Oct},
  publisher = {American Physical Society},
  doi = {10.1103/PhysRevE.106.045102},
  url = {https://link.aps.org/doi/10.1103/PhysRevE.106.045102}
}

@article{lee2020convergence,
author = {Lee, Minhyung and Park, Gwanyong and Park, Changyoung and Kim, Changmin},
title = {Improvement of Grid Independence Test for Computational Fluid Dynamics Model of Building Based on Grid Resolution},
journal = {Advances in Civil Engineering},
volume = {2020},
number = {1},
pages = {8827936},
doi = {https://doi.org/10.1155/2020/8827936},
url = {https://onlinelibrary.wiley.com/doi/abs/10.1155/2020/8827936},
eprint = {https://onlinelibrary.wiley.com/doi/pdf/10.1155/2020/8827936},
year = {2020}
}

@Article{schober2018correlation,
author={Schober, Patrick
and Boer, Christa
and Schwarte, Lothar A.},
title={Correlation Coefficients: Appropriate Use and Interpretation},
journal={Anesthesia {\&} Analgesia},
year={2018},
volume={126},
number={5},
issn={0003-2999},
url={https://journals.lww.com/anesthesia-analgesia/fulltext/2018/05000/correlation_coefficients__appropriate_use_and.50.aspx}
}

@article{leforestier_comparison_1991,
	title = {A comparison of different propagation schemes for the time dependent {Schrödinger} equation},
	volume = {94},
	issn = {0021-9991},
	url = {https://www.sciencedirect.com/science/article/pii/002199919190137A},
	doi = {https://doi.org/10.1016/0021-9991(91)90137-A},
	number = {1},
	journal = {Journal of Computational Physics},
	author = {Leforestier, C. and Bisseling, R. H. and Cerjan, C. and Feit, M. D. and Friesner, R. and Guldberg, A. and Hammerich, A. and Jolicard, G. and Karrlein, W. and Meyer, H.-D. and Lipkin, N. and Roncero, O. and Kosloff, R.},
	year = {1991},
	pages = {59--80},
}

@article{stuart_inverse_2010,
	title = {Inverse problems: {A} {Bayesian} perspective},
	volume = {19},
	doi = {10.1017/S0962492910000061},
	journal = {Acta Numerica},
	author = {Stuart, A. M.},
	year = {2010},
	pages = {451--559},
}

@inproceedings{du_eddyformer_2025,
 author = {Du, Yiheng and Krishnapriyan, Aditi},
 booktitle = {Advances in Neural Information Processing Systems},
 editor = {D. Belgrave and C. Zhang and H. Lin and R. Pascanu and P. Koniusz and M. Ghassemi and N. Chen},
 pages = {103372--103403},
 publisher = {Curran Associates, Inc.},
 title = {EddyFormer: Accelerated Neural Simulations of Three-Dimensional Turbulence at Scale},
 url = {https://proceedings.neurips.cc/paper_files/paper/2025/file/9555307c4c4675a3222024f6e2055586-Paper-Conference.pdf},
 volume = {38},
 year = {2025}
}

@article{um_solver_loop_2020,
	title = {Solver-in-the-{Loop}: {Learning} from {Differentiable} {Physics} to {Interact} with {Iterative} {PDE}-{Solvers}},
	journal = {Advances in Neural Information Processing Systems},
	author = {Um, Kiwon and Brand, Robert and Fei, Yun and Holl, Philipp and Thuerey, Nils},
	year = {2020},
}

@book{heinz_regularization_1996,
title = "Regularization of Inverse Problems",
author = "Heinz Engl and Martin Hanke and Andreas Neubauer",
year = "1996",
language = "English",
isbn = "0-7923-4157-0",
publisher = "Kluwer",
}

@incollection{kitanidis_bayesian_2010,
	title = {Bayesian and {Geostatistical} {Approaches} to {Inverse} {Problems}},
	isbn = {978-0-470-68585-3},
	url = {https://onlinelibrary.wiley.com/doi/abs/10.1002/9780470685853.ch4},
	booktitle = {Large‐{Scale} {Inverse} {Problems} and {Quantification} of {Uncertainty}},
	publisher = {John Wiley \& Sons, Ltd},
	author = {Kitanidis, P. K.},
	year = {2010},
	doi = {https://doi.org/10.1002/9780470685853.ch4},
	note = {Section: 4
\_eprint: https://onlinelibrary.wiley.com/doi/pdf/10.1002/9780470685853.ch4},
	pages = {71--85},
}

@inproceedings{
koshizuka_expressivity_2024,
title={Understanding the Expressivity and Trainability of Fourier Neural Operator: A Mean-Field Perspective},
author={Takeshi Koshizuka and Masahiro Fujisawa and Yusuke Tanaka and Issei Sato},
booktitle={The Thirty-eighth Annual Conference on Neural Information Processing Systems},
year={2024},
url={https://openreview.net/forum?id=QJr02BTM7J}
}

@misc{qin_specboost_2024,
      title={Toward a Better Understanding of Fourier Neural Operators from a Spectral Perspective}, 
      author={Shaoxiang Qin and Fuyuan Lyu and Wenhui Peng and Dingyang Geng and Ju Wang and Xing Tang and Sylvie Leroyer and Naiping Gao and Xue Liu and Liangzhu Leon Wang},
      year={2024},
      eprint={2404.07200},
      archivePrefix={arXiv},
      primaryClass={cs.LG},
      url={https://arxiv.org/abs/2404.07200}, 
}

@article{hestness2017deep,
  title={Deep learning scaling is predictable, empirically},
  author={Hestness, Joel and Narang, Sharan and Ardalani, Newsha and Diamos, Gregory and Jun, Heewoo and Kianinejad, Hassan and Patwary, Md Mostofa Ali and Yang, Yang and Zhou, Yanqi},
  journal={arXiv preprint arXiv:1712.00409},
  year={2017}
}

@article{bahri2024explaining,
  title={Explaining neural scaling laws},
  author={Bahri, Yasaman and Dyer, Ethan and Kaplan, Jared and Lee, Jaehoon and Sharma, Utkarsh},
  journal={Proceedings of the National Academy of Sciences},
  volume={121},
  number={27},
  pages={e2311878121},
  year={2024},
  publisher={National Academy of Sciences}
}

@article{zheng2025inversebench,
  title={Inversebench: Benchmarking plug-and-play diffusion priors for inverse problems in physical sciences},
  author={Zheng, Hongkai and Chu, Wenda and Zhang, Bingliang and Wu, Zihui and Wang, Austin and Feng, Berthy T and Zou, Caifeng and Sun, Yu and Kovachki, Nikola and Ross, Zachary E and others},
  journal={arXiv preprint arXiv:2503.11043},
  year={2025}
}

@article{lu2021deeponet,
  title   = {Learning nonlinear operators via DeepONet based on the universal approximation theorem of operators},
  author  = {Lu, Lu and Jin, Pengzhan and Pang, Guofei and Zhang, Zhongqiang and Karniadakis, George Em},
  journal = {Nature Machine Intelligence},
  volume  = {3},
  number  = {3},
  pages   = {218--229},
  year    = {2021},
  doi     = {10.1038/s42256-021-00302-5},
  url     = {https://www.nature.com/articles/s42256-021-00302-5}
}
\bibliographystyle{abbrvnat}

\newpage

\appendix

% Preliminaries and dataset information
\section{Neural operator framework}
\label{sec:appendix-neural-operators-framework}

Here we recount the mathematical definition of Neural Operators (NOs) as presented by~\citet{li_fourier_2021} and~\citet{kovachki_neural_2023}.
NOs are a class of deep learning models which learn mappings between infinite-dimensional function spaces, offering a powerful framework for solving partial differential equations (PDEs). 
Unlike traditional neural networks, which operate on finite-dimensional Euclidean spaces, neural operators model the non-linear relationship $\mathcal{G}: \mathcal{A} \rightarrow \mathcal{U}$, where $\mathcal{A}$ and $\mathcal{U}$ are Banach spaces of functions defined on bounded sets $D_{\mathcal{A}} \subset \mathbb{R}^{d_\mathcal{A}}$ and $D_\mathcal{U} \subset \mathbb{R}^{d_\mathcal{U}}$, respectively. 
This is particularly relevant for PDEs, where inputs (e.g., coefficients, initial conditions, and solution fields) and outputs (e.g., PDE parameters, and solution fields) are functions.

% Defining the operator learning objective
The objective is to approximate a nonlinear operator $\mathcal{G}$ using a parameterized NO $\mathcal{G}_\theta$, minimizing the expected error over probability measure $\mu$ on $\mathcal{A}$:

\begin{align}
\min_\theta ~\mathbb{E}_{a \sim \mu} \left[ C \left( \mathcal{G}_\theta(a), \mathcal{G}(a) \right) \right],
\end{align}

where $C: \mathcal{U} \times \mathcal{U} \rightarrow \mathbb{R}$ is a cost function.

% Describing the architecture
Each layer of a NO consists of iterative kernel integration layers, where the $i+1$-th layer update consists of
\begin{align}
    v_{i+1} := \sigma(\mathcal{W} (v_i) + \mathcal{K} (v_i)),
\end{align}
$\sigma$ is a non-linear activation function (e.g., ReLU), $\mathcal{W}$ is a local linear operator, and $\mathcal{K}$ is a nonlocal integral kernel operator.
The entire architecture consists of a point-wise lifting operation $\mathcal{P}$, a series of iterative kernel integration layers, and a projection layer $\mathcal{Q}$ (Equation~\ref{eq:neural-operator}).
\begin{align}
\label{eq:neural-operator}
\mathcal{G}_\theta := \mathcal{Q} \circ \sigma \left( \mathcal{W}^{(L)} + \mathcal{K}^{(L)} \right) \circ \ldots \circ \sigma \left( \mathcal{W}^{(1)} + \mathcal{K}^{(1)} \right) \circ \mathcal{P}.
\end{align}

% Highlighting key properties
Neural operators are identifiable by two key properties: \textit{discretization invariance}, allowing the same model parameters to be used across different discretizations, and \textit{universal approximation}, enabling them to approximate any continuous operator on compact sets.    
% A 
% \input{_s7_appendix_related_works}      
\section{Datasets and evaluation metrics}
We provide additional details on the datasets that are part of \inversebench, as well as the evaluation metrics used in our design axis investigation.
\subsection{Datasets.} 
\label{sec:appendix-dataset-eqn}

We study five systems: reaction diffusion, unforced Navier-Stokes, forced Navier-Stokes (turbulent flow), Korteweg-De Vries, and Darcy flow. 
We also include the break-down of evaluation splits by parameter value in Table~\ref{tab:parameter_splits}.
For each parameter, a fixed set of initial conditions are evolved according to the dynamics.

\paragraph{2D Reaction Diffusion (RD).} We use the following form for the activator $u$ and the inhibitor $v$ coupled system:

\begin{align*}
    \partial_t u = D_u \partial_{xx} u + D_u \partial_{yy} u + R_u \\
    \partial_t v = D_v \partial_{xx} v + D_v \partial_{yy} v + R_v\\
\end{align*}

where $R_u$ and $R_v$ are defined by the Fitzhugh-Nagumo equations:

\begin{align*}
    R_u(u, v) = u - u^3 - k - v \\
    R_v (u, v) = u - v
\end{align*}

The parameters of interest are $D_u$ (activator diffusion coefficient), $D_v$ (inhibitor diffusion coefficient), and $k$ (threshold for excitement).
We only run experiments on $D_u$ and $k$ due to computational limitations.

\paragraph{Unforced 2D Navier-Stokes (NS).} 
We consider the vorticity form of the unforced Navier-Stokes equations:

\begin{align}
    \frac{\partial w(t, x, y)}{\partial t} + u(t, x, y) \cdot \nabla w(t, x, y)  = \nu \Delta w(t, x, y),  &&t \in [0, T], \quad (x, y) \in (0, 1)^2\label{eqn:ns}\\
    w = \nabla \times u, \quad \nabla \cdot u = 0, \nonumber\\  w(0, x, y) = w_0(x, y), &&\text{(Boundary Conditions)}\nonumber
\end{align}

$\nu$ is the physical parameter of interest, representing viscosity.

\paragraph{Forced 2D Navier-Stokes/Turbulent Flow (TF).}  
The forced Navier-Stoked equations with the Kolmogorov forcing function are similar to the unforced case with an additional forcing term:

\begin{align*}
    \partial_t w + u \cdot \nabla w = \nu \Delta w + f(k, y) - \alpha w \\
    f(k, y) = -k \cos (ky) 
\end{align*}

Similar to unforced NS, we similarly use the vorticity ($w$) form.
$\alpha$, the drag coefficient, and $k=2$, the forced wavenumber, are fixed quantities. 
$\alpha$ primarily serves to keep the total energy of the system constant, acting as drag.
Similar to~\citet{kochkov_machine_2021}, we use $\alpha=0.1$.
The task is to predict $\nu$.

\paragraph{Korteweg–De Vries (KdV).} KdV is a 1D PDE representing waves on a shallow-water surface. 
The governing equation follows the form: 

\begin{align*}
    0 = \partial_t u + u \cdot \partial_x u + \delta^2 \partial_{xxx} u
\end{align*}

$\delta$, the physical parameter, represents the strength of the dispersive effect on the system. 
In shallow water wave theory, $\delta$ is a unit-less quantity roughly indicating the relative depth of the water~\citep{polyanin_handbook_2012}.

\paragraph{Darcy Flow.} 
We use the same formulation of Darcy flow as~\citet{li_fourier_2021}.
The Darcy flow equations model fluid flow through porous media such as groundwater movement through soil or oil through reservoir rock. 
The 2-D steady-state Darcy flow equation on the unit box $\Omega = (0,1)^2$ is a second-order linear elliptic PDE with Dirichlet boundary conditions:
\begin{align}
    -\nabla \cdot (a(x) \nabla u(x)) &= f(x), && x \in \Omega, \\
    u(x) &= 0, && x \in \partial\Omega,
\end{align}

where $a \in L^{\infty}((0,1)^2; \mathbb{R}^+)$ is a piecewise constant diffusion coefficient, $u(x)$ is the pressure field, and $f(x) = 1$ is a fixed forcing function.

\label{sec:appendix-pde-parameter-ranges}
\begin{table}[htbp!]
\centering
\begin{tabular}{cccc}
\toprule
\textbf{PDE} & \textbf{Test} & \textbf{OOD (Non-Extreme)} & \textbf{OOD (Extreme)} \\
\midrule
2D RD & $k \in [0.01, 0.04] \cup [0.08, 0.09]$ & $k \in [0.04, 0.08]$ & $k \in [0.001, 0.01] \cup [0.09, 0.1]$ \\
 & $D_u \in [\text{0.08}, \text{0.2}] \cup [\text{0.4}, \text{0.49}]$ & $D_u \in [\text{0.2}, \text{0.4}]$ & $D_u \in [\text{0.02}, \text{0.08}] \cup [\text{0.49}, \text{0.5}]$ \\
 & $D_v \in [\text{0.08}, \text{0.2}] \cup [\text{0.4}, \text{0.49}]$ & $D_u \in [\text{0.2}, \text{0.4}]$ & $D_u \in [\text{0.02}, \text{0.08}] \cup [\text{0.49}, \text{0.5}]$ \\

\midrule
2D NS & $\nu \in [10^{-3.8}, 10^{-3.2}] \cup [10^{-2.8}, 10^{-2.2}]$ & $\nu \in [10^{-3.2}, 10^{-2.8}]$ & $\nu \in [10^{-4}, 10^{-3.8}] \cup [10^{-2.2}, 10^{-2}]$ \\
\midrule
2D TF & $\nu \in [10^{-4.7}, 10^{-3.8}] \cup [10^{-3.2}, 10^{-2.3}]$ & $\nu \in [10^{-3.8}, 10^{-3.2}]$ & $\nu \in [10^{-5}, 10^{-4.7}] \cup [10^{-2.3}, 10^{-2}]$ \\
\midrule
KdV & $\delta \in [1.22, 2.48] \cup [3.32, 4.58]$ & $\delta \in [2.48, 3.32]$ & $\delta \in [0.8, 1.22] \cup [4.58, 5]$ \\
\midrule
2D DF & Central mass of max-value fraction distribution & - & Tails beyond $\pm 1.5 \sigma$ \\
\bottomrule
\end{tabular}
\caption{Parameter splits for different PDEs}
\label{tab:parameter_splits}
\end{table}

\subsection{Dataset generation parameters}
\label{sec:appendix-pdes-solver}

\paragraph{Computational requirements.}
Some systems are solved using a CPU solver and while others are GPU-accelerated. 
For the GPU-accelerated systems, we use 40GB A100s, parallelizing over 4 GPUs.
We ran the CPU solvers on two AMD EPYC 9354 32-core processors.
In the ensuing paragraphs, for each system, we note which type of solver it uses and the approximate wall clock time on a single CPU / GPU machine.

\paragraph{2D Reaction Diffusion (RD) [parabolic].} 
Solutions are generated using an explicit Runge-Kutta method of order 5(4) (RK45) for temporal integration, as implemented in the PDEBench framework~\citep{dormand_family_1980,takamoto_pdebench_2022}, with a relative error tolerance of $10^{-6}$. 
The spatial discretization employs a Finite Volume Method (FVM) with a uniform grid of $128 \times 128$ cells over the domain $[-1, 1] \times [-1, 1]$, yielding a cell size of \(\Delta x = \Delta y = 0.015625\). The simulations have a burn in period of 1 simulation second. The subsequent dataset simulations span a time interval of \([0, 5]\) seconds, discretized into 101 time steps, resulting in a nominal time step of \(\Delta t \approx 0.05\) seconds, adaptively adjusted by the RK45 solver to meet the error tolerance. 
We use SciPy's \textsf{solve\_ivp} on CPU, taking $\approx$ 1 week to generate.

\paragraph{Unforced 2D Navier-Stokes (NS) [parabolic].}
We use a pseudo-spectral solver~\citep{du_neural_2024} with a Crank-Nicolson time-stepping scheme~\citep{canuto_spectral_2007, li_fourier_2021}. 
The solver is written in Jax~\citep{jax2018github} and accelerated using GPUs. 
Generation takes $\approx 3.5$ GPU days (batch size=32). The solver has a burn in period of 15 simulation seconds, with the next 3 simulation seconds being saved as the dataset.
Initial conditions are sampled according to a Gaussian random field (length scale=0.8).
The solution is recorded every 1 simulation second and uses a simulation $\partial t=1e^{-4}$. 
Solutions are simulated and recorded at a resolution of $256\times256$.

\paragraph{Forced 2D Navier-Stokes (TF).}
All solutions in this dataset exhibit turbulence and are generated using a pseudo-spectral solver with a Crank-Nicolson time-stepping scheme~\citep{dresdner_learning_2023}. 
Similar to 2D NS, the solver is written in Jax (specifically leveraging Jax-CFD), and takes $\approx 4$ GPU days (A100).
The solver has a burn in period of 40 simulation seconds, with the next 15 simulation seconds being saved as the dataset.
The simulator runs at a resolution of $256\times256$ and downsamples to $64\times64$ before saving at a temporal resolution of $\partial t = 0.25$ simulation seconds.

\paragraph{Korteweg-De Vries (KdV) [hyperbolic].}
The KdV equation is solved on a periodic domain $[0, L]$ using a pseudospectral method with a Fourier basis for spatial discretization ($N_x = 256$ grid points) and an implicit Runge-Kutta method (Radau IIA, order 5) for time integration, implemented via SciPy's \texttt{solve\_ivp} ~\citep{hairer_stiff_1999,virtanen_scipy_2020, brandstetter_lie_2022}.
Simulations are ran on CPU and takes $\approx 12$ hours. The solver has a burn in period of 40 simulation seconds, with the next 102 simulation seconds being saved as the dataset.
The time step is adaptive with absolute and relative tolerances of $10^{-9}$.
Like~\citet{brandstetter_lie_2022} and~\citet{bar-sinai_learning_2019}, initial conditions are sampled from a distribution over a truncated Fourier Series with coefficients $\{A_k, l_k, \phi_k\}_k$.
%\ak{Is this the exact same PDE and dataset as in these other papers?} \nc{Same PDE equation but different dataset. We simulate KdV ourselves to generate the dataset. The only dataset we do not simulate is Darcy Flow.}

\begin{align*}
    u_0(x) = \sum_{k=1}^{K} A_k \sin\left(2\pi l_k \cdot \frac{x}{L} + \phi_k\right),
\end{align*}

where $A_k, \phi_k  \sim U (0, 1)$ and $l_k \sim U(1, 3)$.

\paragraph{Darcy Flow (DF) [elliptic].}
To complement our generated datasets, we also include Darcy flow from~\citet{li_fourier_2021}.
Though we do not generate the Darcy flow data itself, we include relevant parameters as documented by~\citet{li_fourier_2021} for completeness.
The original solver is a second-order finite difference method, with a resolution of $421\times421$. 
Originally written in Matlab, the solver runs on CPU.
The lower resolution dataset is generated by downsampling the $421\times421$ dataset. 
The coefficient field $a(x)$ is sampled from $\mu  = \Gamma(\mathcal{N} (0, - \Delta + 9I)^{-2})$.
$\Gamma$ is a element-wise map $\Gamma: a_i \sim \mathcal{N}(0, - \Delta + 9I)^{-2} \mapsto \{3, 12\}$.
$a_i \mapsto 12$ when $a_i \geq 0$ and $a_i \mapsto 3$ when $a_i < 0$.
Over the coefficient field, zero Neumann boundary conditions on the Laplacian are enforced.

\subsection{Numerical Convergence}
\label{sec:numerical_convergence_1}

\paragraph{Grid independence test.} 
To validate the convergence and correctness of the data we generate, we use the grid-independence test~\citep{lee2020convergence}.
The grid-independence test ensures numerical simulation results are not dependent on the size or resolution of the underlying mesh.
To perform the grid-independence test, we compute the Pearson correlation~\citep{schober2018correlation} and relative $L_2$ loss between solutions generated at the training resolution and a down-sampled high-resolution solution.
Similar to the parameter relative loss, the $L_2$ relative loss defined by $RelErr(u_{ref}, u) = \frac{||u - u_{ref}||_2} {||u_{ref}||_2}$.
To downsample the high resolution solutions, we use nearest neighbor interpolation. 
We report results on both OOD-Extreme splits separately, represented by ``small" (lower parameter values) and ``large" (higher parameter values) in table~\ref{tab:appendix-grid-independence-test}.
The OOD-Extreme splits capture the smallest and largest parameter values, probing whether solutions are converged at extreme ends and provide a principled way to test for convergence.
High resolution solutions for 2D reaction diffusion are generated at $512^2$ and at $256^2$ for unforced 2D Navier-Stokes.
Both 2D reaction diffusion and unforced 2D Navier-Stokes exhibit low relative error and high correlation in both parameter regimes, indicating convergence.
The chaotic nature of turbulence means the grid independence test is not a good measure of convergence for forced 2D Navier-Stokes, and we instead defer to the energy spectrum analysis in~\ref{sec:appendix-tf-convergence}.

\begin{table}[htbp!]
\centering
\begin{tabular}{|l|ll|ll|}
\hline
\multicolumn{1}{|c|}{\textbf{System}} &
  \multicolumn{1}{c|}{\textbf{$L_2$ Rel. Err. (\%)}} &
  \multicolumn{1}{c|}{\textbf{Pearson Corr.}} &
  \multicolumn{1}{c|}{\textbf{$L_2$ Rel. Err. (\%)}} &
  \multicolumn{1}{c|}{\textbf{Pearson Corr.}} \\
\multicolumn{1}{|c|}{\textit{Split}} &
  \multicolumn{2}{c|}{\textit{OOD-Extreme Small}} &
  \multicolumn{2}{c|}{\textit{OOD-Extreme Large}} \\ \hline
2D RD - $k$ &
  \multicolumn{1}{l|}{3.2611\% $\pm$ 0.8585 \%} &
  0.999874 $\pm 1.26e^{-4}$ &
  \multicolumn{1}{l|}{2.1440\% $\pm$ 0.1016 \%} &
  0.999857 $\pm1.42e^{-4}$\textit{}
   \\
2D RD - $D_u$ &
  \multicolumn{1}{l|}{2.6687\% $\pm$ 0.8578 \%} &
  0.999836 $\pm 1.18e^{-4}$ &
  \multicolumn{1}{l|}{2.7365\% $\pm$ 0.7957 \%} &
  0.999896 $\pm 1.43e^{-4}$
   \\
Unforced 2D NS &
  \multicolumn{1}{l|}{2.1656\% $\pm$ 0.0002 \%} &
  1.008181 $\pm 2.45e^{-4}$ &
  \multicolumn{1}{l|}{2.1303\% $\pm$ 0.0000\%} &
  1.008066 $\pm 3.e^{-6}$ \\
% Forced 2D NS &
  % \multicolumn{1}{l|}{135.0635\% $\pm$ 2.2865 \%} &
  % 0.075284 $\pm$ 0.008251 &
  % \multicolumn{1}{l|}{133.6116\% $\pm$ 4.4670 \%} &
  % 0.109574 $\pm$ 0.031887
   % \\ 
  \hline
\end{tabular}
\caption{Grid independence test for 2D reaction diffusion and unforced 2D Navier-Stokes partitioned by parameter regime.}
\label{tab:appendix-grid-independence-test}
\end{table}

\subsection{Numerical Convergence of Forced 2D Navier-Stokes}
\label{sec:appendix-tf-convergence}

While~\cite{dresdner_learning_2023} discuss and verify the convergence of forced 2D Navier-Stokes, they do not cover the broad range of parameters we generate data for.
Our dataset includes Reynolds numbers outside the range of parameters they study.
We perform a similar analysis by comparing the energy spectra relative to a high resolution reference solution. 
The reference solution is generated with a $2048^2$ spatial grid over 1 simulation second. 
While we generate training and evaluation data for a longer length of time, the chaotic nature of the problem means errors will diverge~\citep{budanur2022scale} for sufficiently long simulations.

\begin{figure}[htbp!]
    \centering
    \includegraphics[width=\linewidth]{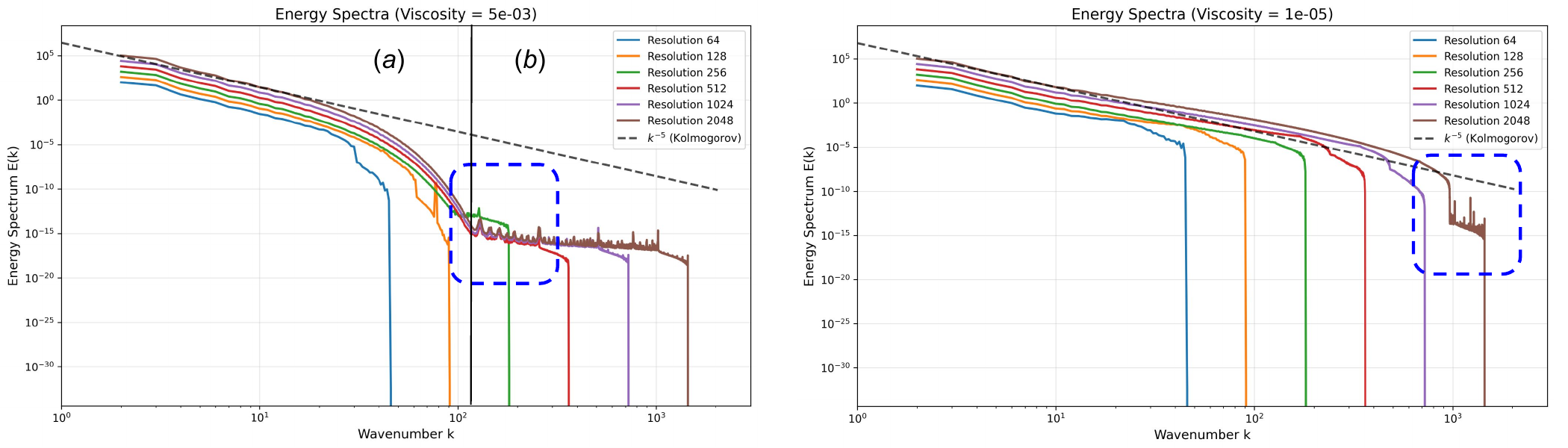}
    \caption{\textbf{Energy spectra convergence of 2D Navier-Stokes Forced.} 
    (Left) Energy spectra for $\nu=5e^{-3}$. 
    Region \emph{a}: inertial range, with rough adherence to the $k^{-5}$ Kolmogorov power law.
    Region \emph{b}: dissipative region where energy leaves the system. 
    (Right) Energy spectra for $\nu=1e^{-5}$. 
    (Both) The blue bounding box highlights convergence behavior, characterized by a steep drop-off in energy spectra into minor oscillations.
    For $\nu=5e^{-3}$ multiple resolutions exhibit convergence including $256^2$.
    In the high turbulence regime ($\nu=1e^{-5}$), $2048^2$ captures the full energy spectra.
    }
    \label{fig:appendix-spectra-2d-tf}
\end{figure}

\paragraph{Energy spectrum.} 
An important method for understanding the convergence of numerical simulations, particularly pseudo-spectral methods for turbulence, is to examine the resulting energy spectra~\citep{canuto_spectral_2007, evans_partial_2022}. 
The energy spectra is an important tool for understanding the dynamics of turbulence and describes the energy distribution across different spatial length scales.
In Figure~\ref{fig:appendix-spectra-2d-tf} we plot the energy spectra of two viscosities, representing simpler turbulence ($\nu = 5e^{-3}$) and the most turbulence captured by our parameter range ($\nu=1e^{-5})$
For both parameter values, we see convergence into the dissipative range characterized by a steep drop-off in energy followed by mild oscillations, highlighted via the blue bounding box.

At the most turbulence regimes (e.g., $\nu=1e^{-5}$), generating and storing solutions at $2048^2$ is necessary to capture all fine features.
We include a small subset of high resolution data as a separate evaluation for downstream users of \emph{PDEInvBench}.
However, we keep our main text evaluations on a $64^2$ grid as it is a harder problem than predicting on a finely discretized grid.
With a sufficiently fine discretization that resolves the dissipative cutoff, the viscosity is strongly constrained by the smallest active length scales.
$\nu$ directly controls the high-wavenumber energy drop-off and the onset of the dissipative range in the energy spectrum.
In contrast, evaluations on a coarse grid should be viewed as a partial observation of the flow after a low-pass filter or downsampling projection.
For the inverse problem, we infer $\nu$ from the observation where the downsampling removes the small-scale content that is most informative about viscous dissipation.
When the dissipative range is only partially observed, different viscosities can induce very similar large-scale trajectories and spectra over the resolved band, making parameter identification ill-posed and therefore a strictly harder than when the full field is available.
At $64^2$, our evaluation should be interpreted as a partially observed inverse problem.
The model is given only the large-scale components of a resolved simulation, and is required to infer parameters from incomplete scale information, providing a test of robustness.

% \paragraph{Remark on training data resolution.}
% We store the 2D Navier-Stokes Forced training data at a $64^2$ resolution, which inherently limits the fine turbulence dynamics capable of being captured when stored in a downsampled solution.
% Directly training on the high resolution solutions 

% \begin{figure}[htbp!]
%     % \centering
%     \includegraphics[width=.7\linewidth]{Figures/downsampled-spectra-2dtf.pdf}
%     \caption{Energy spectra for $\nu=10^{-5}$ demonstrating the loss of high frequency information. The spectrum of all resolutions finer than $256^2$ are limited not by numerical convergence, but storage resolution.}
%     \label{fig:placeholder}
% \end{figure}

% However, as mentioned in the main text, we would like high resolution converged numerical simulations for our evaluation splits.
% We still include our $2048^2$ dataset and additionally include a separately generated high resolution $4096^2$ dataset which is a subset of our evaluation parameters. 
% We include both the energy spectra and relative error for a sample trajectory in Figure~\nc{todo, include this figure}.
% \nc{WIP, need to expand this argument / de duplicate from the main text? Talk with Sanjeev + Divyam on where to put this. Also discuss how to frame the high resolution 2048 dataset.}

\subsection{Additional information on evaluation splits}
\label{sec:appendix-eval-splits}
\begin{figure}[htbp!]
    \centering
    \includegraphics[width=\linewidth]{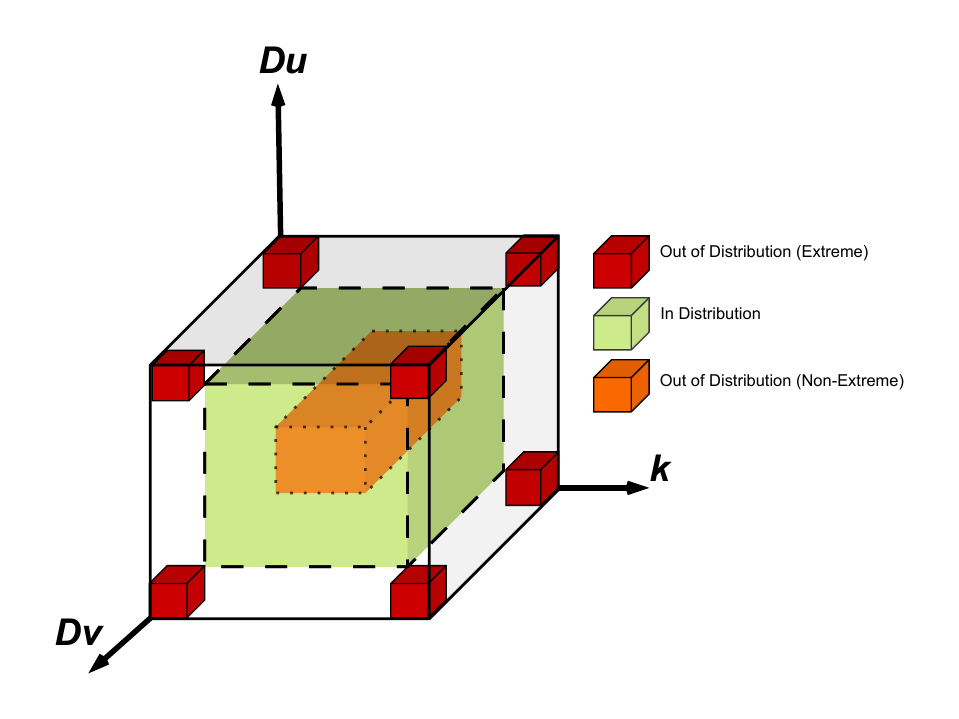}
    \caption{\textbf{Parameter partitioning for 2D RD.} The parameter space for 2D RD forms a cube with each dimension corresponding to possible values along a parameter. The inner cube (orange) with each edge covering the middle 16\% of parameter values corresponds to the OOD (Non-Extreme) split. The middle cube (green) with each edge covering 32 \% of parameters corresponds to the ID setting. The corner cubes (red) with edge length covering 10\% of parameter values correspond to the OOD (Extreme) setting.}
    \label{fig:eval_split}
\end{figure}

We take the best-performing models from our validation set and evaluate on the ID Test, OOD (Extreme) and OOD (Inclusive) at test time. 
This partitioning strategy allowed us to assess model performance both within the training distribution and at different levels of extrapolation.
In the case of 2D reaction diffusion with multiple physical parameters, we generalize the notion of parameter ranges on a line to a hyper-cube (Figure~\ref{fig:eval_split}). 

For Darcy flow, we compute the fraction of grid points in each coefficient field that take the maximum value (12). 
This statistic is approximately normally distributed across the dataset. 
We partition coefficient fields according to this distribution using $\pm 1.5$ standard deviations, with the central mass defining the in-distribution (ID) test set and the tails defining the OOD-Extreme setting.

We also include the specific parameter ranges of each split for each dataset in Table~\ref{tab:parameter_splits}.

\subsection{Evaluation metrics}
\label{sec:appendix-metrics}

\paragraph{Error bars for relative error}
For each of the three seeded runs, we select the model with the best performance on the validation set. We compute the relative error for each model on the evaluation splits. 
The mean and standard deviation of these three relative errors are calculated and reported as the error bars in our plots (1 $\sigma$ error bars).

\paragraph{Error bars for negative slope of the line of best fit}
Using the best-performing models from the validation sets of the three seeded runs, we compute the negative slope of the line of best fit (NLS) for each seed and data scaling procedure. 
For each scaling procedure, we calculate the mean and standard deviation of the three NLS values and report these as the error bars in our plots (1 $\sigma$ error bars).

For our evaluation splits, we deterministically sample every 10th window of PDE frames along the generated trajectory.           
% B

% Additional investigations and results
\section{Methodology for Investigating Design Axes}
\label{sec:appendix-additional-investigations}
We now provide additional details on our methodology for investigating the three key design axes considered in our work: 1) optimization procedure, 2) problem representation and inductive bias, and 3) scaling properties. In this section, we focus on the methodology for experiments included in the main text. We include methodology and results for additional experiments in Section \ref{sec:appendix-additional-results}.

\subsection{General details}
\label{sec:appendix-general-details}
We first provide details shared across all design axes.

\subsubsection{Base model architecture}
\label{sec:appendix-baseline-model}

The base model for solving the inverse problem is a neural network designed to predict PDE parameters from input data, consisting of three primary components: a FNO-based encoder, a convolutional downsampler to reduce the dimensionality of the activation maps, and a MLP regression head to predict the PDE parameter. The encoder processes 2 temporal PDE frames of input data, incorporating partial derivatives specific to the PDE. The encoder consists of four FNO layers, 16 Fourier modes per spatial dimension, and hidden channel dimension of 64. The encoder output is passed to a four-layer convolutional downsampler. Each downsampler layer applies a two-dimensional convolution with 64 input and output channels, a kernel size of 3, a stride of 1, and padding of 2, followed by ReLU activation and two-dimensional max-pooling with a kernel size of 2. Thus, the downsampler reduces the spatial resolution by a factor of 16. The downsampler output is flattened and fed into an MLP head with one hidden layer of 64 units using ReLU activation. A single value is returned, corresponding to the PDE parameter. 

\paragraph{DeepONet.} Our DeepONet implementation follows the standard branch-truck decomposition with slight changes adapting it for the inverse problem setting. The branch network, which processes the PDE solution at collocation points, takes in the full spatial solution and is implemented as a ResNet, primarily for computational efficiency when handling the full grid. The trunk network for encoding sensor locations is implemented as a pointwise MLP with residual connections and GeLU activations. Since most of our inverse problems predict a scalar PDE parameter rather than a spatio-temporal field, we apply mean pooling over the pointwise branch-trunk inner produce to produce the final scalar output.

\subsubsection{Training details}
We provide training details and hyperparameters common to all systems and experiments. We use a batch size of 32.  We use the Adam optimizer with a cosine decay learning rate scheduler. The initial learning rate is set to 1e-4, and is decayed to 0. We train our models for 200 epochs.

\subsubsection{Compute usage}
Here we provide the total training time, and memory usage of all datasets and model performed on NVIDIA RTX A100 in Table \ref{tab:compute_usage}.

\begin{table}[h]
\centering
\begin{tabular}{|c|c|c|c|c|}
\hline
Dataset & Model  & Total Training Time & Per Batch Inference Speed & GPU Memory Usage \\
\hline
\multirow{3}{*}{2DNS} & FNO    & 1 hour  & 33.64 it/s & 4.7 GB \\
                      & SCOT   & 2.5 hours & 11.15 it/s  & 4.2 GB \\
                      & ResNet & 1.5 hours  & 22.07 it/s  & 3.8 GB   \\
\hline
\multirow{3}{*}{2DRD} & FNO    & 6.5 hours  & 1.95 it/s & 8.5 GB \\
                      & SCOT   & 6.45 hours & 1.73 it/s  & 4.5 GB \\
                      & ResNet & 6.4 hours  & 1.71 it/s  & 8 GB   \\
\hline
\multirow{3}{*}{2DTF} & FNO    & 1 hours  & 20.14 it/s & 3.3 GB \\
                      & SCOT   & 1.2 hours & 12.6 it/s  & 2.3 GB \\
                      & ResNet & 1.7 hours  & 7.86 it/s  & 6.3 GB   \\
\hline
\multirow{3}{*}{1DKDV} & FNO    & 33 minutes  & 32.35 it/s & 4.7 GB \\
                       & SCOT   & 1.5 hours & 23.6 it/s  & 5.8 GB \\
                       & ResNet & 46.68 minutes  & 23.6 it/s  & 4.2 GB   \\
\hline
\multirow{3}{*}{2DDF} & FNO    & 18 minutes  & 9.5 it/s & 7.5 GB \\
                      & SCOT   & 29 minutes & 3.45 it/s  & 4.5 GB \\
                      & ResNet & 60 minutes  & 1.1 it/s  & 8 GB   \\
\hline
\end{tabular}
\label{tab:compute_usage}
\caption{Training time, inference speed, and memory usage of FNO, SCOT, and ResNet across different datasets.}
\end{table}

\subsection{Optimization procedure}
\label{sec:appendix-opt-procedure}
We provide additional details on the optimization procedure design axis.

\subsubsection{Test-Time Training details}

For test-time training (TTT), we consider two configurations: \textit{per-batch} tailoring, in which we adapt the model for each batch, and \textit{per-element} tailoring, where we finetune individually for each test sample. In both cases, we perform tailoring for 50 gradient steps with a learning rate of 1e-5 and the adam optimizer. We show the results of \textit{per-element} tailoring in \ref{sec:Results}. Here, we include comparisons of \textit{per-batch} tailoring and \textit{per-element} tailoring. We also include ablations on varying the anchor loss weight $\alpha \in \{0, 0.01, 1\}$.

\subsection{Problem representation and inductive biases}
\label{sec:appendix-investigations-problem-rep}

We provide additional details about the experiments relating to the problem representation and inductive biases design axis.

\subsubsection{Model hyperparameters and details}
\label{sec:appendix-model-hparams}
We provide hyperparameters for all model variants considered. For all model variants, only the encoder portion of the model is changed. The convolutional downsampler and MLP head remain the same as what was described in Section \ref{sec:appendix-baseline-model}.
% \paragraph{FNO}
% \paragraph{ResNet}
% \paragraph{scOT}

% Defining the plan: Writing subsections for each architecture (FNO, ScOT, ResNet)
\paragraph{Fourier Neural Operator (FNO)}
The FNO architecture is configured with 4 Fourier layers, 16 modes, and 64 hidden channels.

\paragraph{Scalable operator Transformer (scOT)}
The scOT architecture employs 4 Swin Transformer layers with an embedding dimension of 36, patch size of 4, varying attention heads (3, 6, 12, 24), and a hidden size of 32.

\paragraph{ResNet}
The ResNet architecture features 6 residual layers with 128 hidden channels.

When predicting a single scalar parameter, each model is followed by a convolutional downsampler and single-layer MLP. When predicting a spatial parameter field as in Darcy-Flow, the convolution downsampler is replaced with an identity map. The MLP head is replaced with a 3 layer convolution head that uses point-wise convolutions to generate a segmentation map. The final activations are converted to a binary segmentation map using the sigmoid function.

\subsection{Data scaling}
\label{sec:appendix_scaling}

We provide additional results on the data scaling experiments, deferring a description of model scaling to Section \ref{sec:appendix-model-scaling}. 

We investigate how the amount of training data affects performance by systematically varying three aspects of the dataset:

\begin{enumerate}
\item \textbf{Initial condition scaling}: We vary the number of initial conditions per parameter value at \{20\%, 35\%, 50\%, 75\%, and 100\%\} of the full dataset. 

\item \textbf{Parameter scaling}: We vary the density of parameter sampling at \{20\%, 35\%, 50\%, 75\%, and 100\%\} of the full range.

\item \textbf{Temporal scaling}: We vary the total temporal horizon on which the model is trained by varying the sampled frames from the first \{10\%, 20\%, 50\%, 75\%\} of the total generated temporal range of our dataset. The evaluation set is the final 25\% of the generated temporal range for the in-distribution test setting and the entire temporal trajectory for the OOD evaluations settings.

\end{enumerate}

We also investigate how scaling the amount of training data along initial conditions and parameters impacts performance across the different architectures. 
When the percentage-based data scaling results in non-integer number of total frames, we round up to the nearest integer.

\subsection{Darcy Flow experimental constraints}
\label{sec:darcy_flow_experiments}

Since Darcy flow is time-independent, initial-condition scaling, temporal-horizon scaling, and temporal conditioning experiments are not applicable to the system.
     
% C
\section{Additional results}
\label{sec:appendix-additional-results}
We provide the results of additional experiments not included in the main text. Again, we split these into the same three principle design axes.

\subsection{Optimization}
\label{sec:appendix-additional-results-optimization}
\subsubsection{Benchmarking classical methods against Neural Operators}

We compare FNO to the following classical baselines Newton-CG, L-BFGS-B, SLSQP. For solvers requiring an initial estimate, we uniformly sample 10 initial parameters (or log-spaced uniformly for log-spaced parameters) from within the parameter range. Across all
systems, the classical methods we try perform worse than the FNO baseline, suggesting the viability of Neural
Operators for PDE inverse problems Table \ref{tab:optimizer_comparison_10_pts}.

\begin{table}[h]
\centering
\begin{tabular}{|l|c|c|c|c|}
\hline
\textbf{System} & \textbf{L-BFGS-B} & \textbf{Newton-CG} & \textbf{SLSQP} & \textbf{FNO} \\
\hline
2D RD -- $k$   & $1.6 \pm 6.0 \times 10^{-4}$ & $1.6 \pm 3.7 \times 10^{-9}$ & $1.6 \pm 2.0 \times 10^{-4}$ & $\underline{0.35 \pm 0.021}$ \\
2D RD -- $D_u$ & $0.98 \pm 9.8 \times 10^{-9}$ & $0.98 \pm 2.1 \times 10^{-15}$ & $0.98 \pm 2.0 \times 10^{-10}$ & $\underline{0.15 \pm 0.067}$ \\
2D NS Unforced & $5.1 \pm 7.2$ & $5.1 \pm 7.2$ & $5.2 \pm 7.1$ & $\underline{0.021 \pm 0.0011}$ \\
2D NS Forced   & $31 \pm 50$ & $9.1 \pm 7.7 \times 10^{-13}$ & $32 \pm 49$ & $\underline{0.37 \pm 0.011}$ \\
1D KdV         & $0.092 \pm 0.0049$ & $0.090 \pm 0.0012$ & $0.21 \pm 0.025$ & $\underline{0.03 \pm 0.0016}$ \\
\hline
\end{tabular}
\caption{Comparison of classical optimization methods and FNO across PDE systems (mean $\pm$ std). Comparing FNO with L-BFGS-B, Newton-CG, and SLSQP. The classical optimization methods are initialized with 10 uniformly sampled seeds from the parameter range. FNO consistently achieves the lowest error across all systems.}
\label{tab:optimizer_comparison_10_pts}
\end{table}

\subsubsection{Physics-Informed Loss Function}

\begin{figure}[htbp!]
    \centering
    \includegraphics[width=\linewidth]{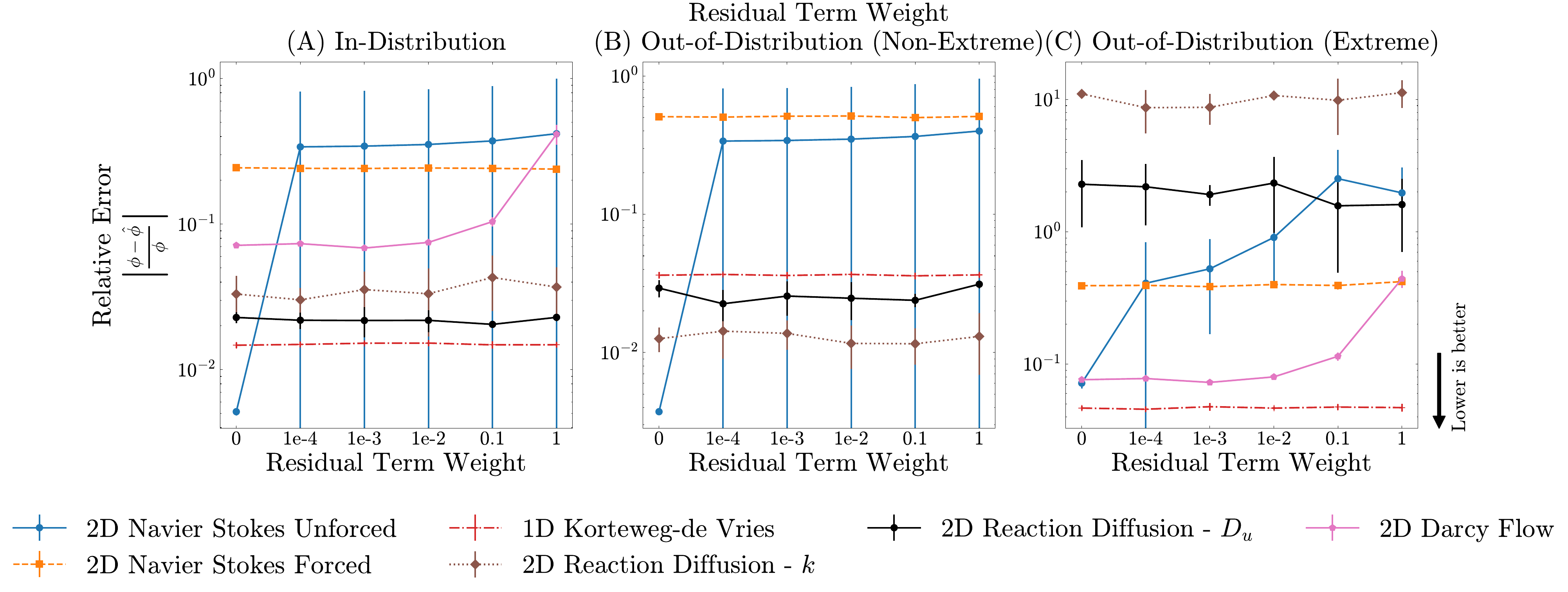}
    \caption{\textbf{Effect of PDE residual loss weights in the PINNs loss.} Relative error from the best performing models on the validation set across three scenarios: ID (left), OOD Non-Extreme (middle), and OOD Extreme (right). The x-axis represents PDE residual term weight ranging from 0 to 1 on a logarithmic scale. Joint training with the PDE residual offers no significant improvements over direct parameter supervision, as all nonzero weights yield worse performance than with a weight of zero.}

    \label{fig:residual_weights}
\end{figure}

In addition to the loss formulations considered in the main text, we consider a generalized, ``physics-informed" loss combining data-driven supervision with self-supervision from the PDE residual:
\begin{equation}
    \label{eq:pinn_loss}
   \mathcal{L}_{\text{physics-informed}} = \alpha \mathcal{L}_{\text{data}} + \beta \mathcal{L}_{\text{res}}.
\end{equation}
Here, $\alpha, \beta$ are weighting coefficients. Note that this is a physics-informed loss for the inverse problem, meaning $\mathcal{L}_\text{data}$ is taken with respect to the ground truth PDE parameter $\phi$ and predicted parameter $\hat{\phi}$ (e.g., $\mathcal{L}_\text{data} = ||\hat{\phi}^i - \phi^i||_2 / ||\phi^i||_2$).
Section \ref{sec:opt_results} presented results on two extremes: a supervised, purely data-driven setting ( $\alpha = 1, \beta = 0$), and a purely self-supervised, residual-driven setting ($\alpha = 0, \beta = 1$). Building on this, we vary the residual weight $\beta$ used in the physics-informed loss (Equation \ref{eq:pinn_loss}) from 0 to 1, showing results in~\Figref{fig:residual_weights}.

Across all evaluation settings and PDE systems, we observe that no nonzero value of the residual weight significantly outperforms basic parameter supervision alone ($\beta = 0$). Most systems show relatively flat performance curves as $\beta$ increases from 0 to 0.1, followed by performance degradation with higher weights. The residual term becomes actively detrimental at higher weights, likely because it creates a more complex loss landscape that complicates optimization \cite{krishnapriyan2021characterizing}. These findings suggest that if paired data is available, practitioners may safely omit the residual term during initial training without sacrificing performance.

\subsubsection{Varying anchor weight in test-time training}
\begin{figure}[htbp!]
    \centering
    \includegraphics[width=\linewidth]{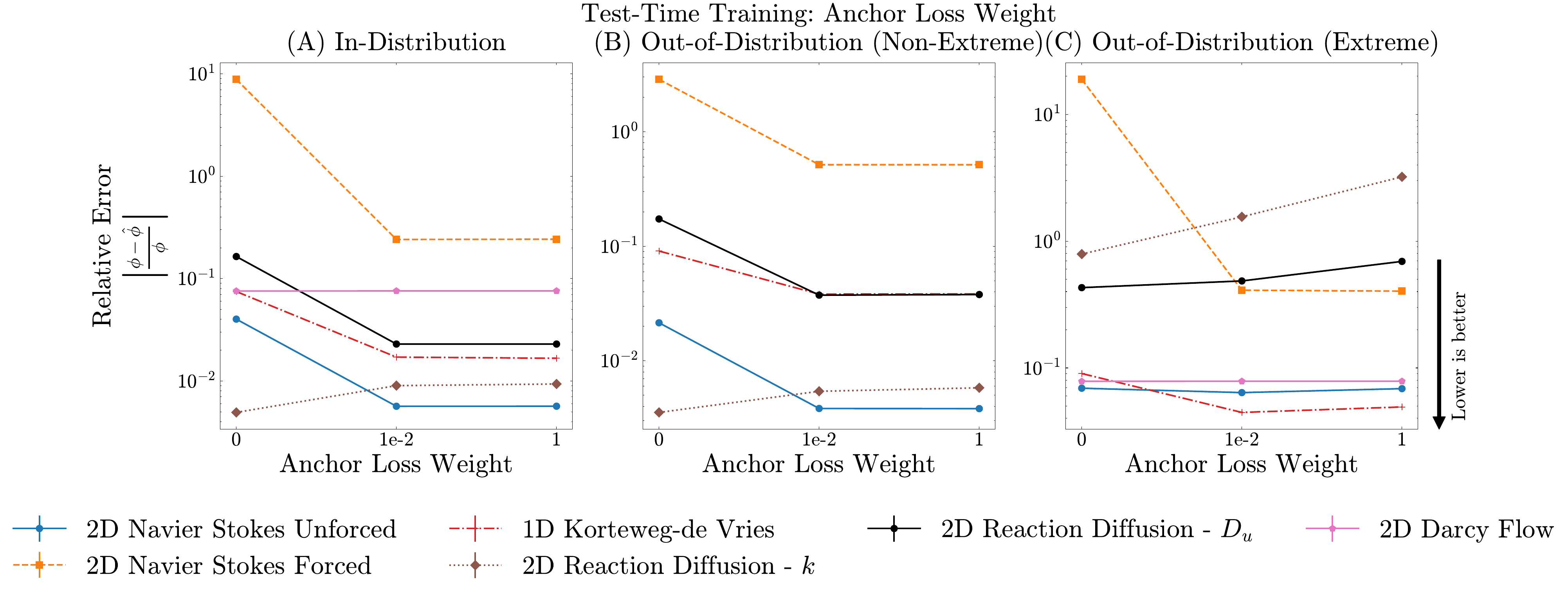}
    \caption{Effect of test time tailoring with different anchor loss weights per batch. Using an anchor loss weight equal to the PDE residual loss weight (1) generally gives the best performance}
    \label{fig:tailoring_anchor_loss_weights}
\end{figure}

% \begin{figure}[htbp!]
%     \centering
%     \includegraphics[width=\linewidth ]{Figures/test-time-tuning.pdf}
%     \caption{Effect of test time tailoring per batch-element}
%     % \label{fig:tailoring_final_results}
% \end{figure}

\Figref{fig:tailoring_anchor_loss_weights} demonstrates how varying the anchor loss weight in the test-time training objective ($\alpha$ in Equation \ref{eq:ttt_loss}) influences performance. We find that the optimal setting uses equal weighting between residual and anchor terms. When the anchor loss is weighted too lightly, the relative error tends to increase with training steps, indicating optimization instability. 

\subsubsection{Per-element vs per-batch tailoring}
\begin{figure}[htbp!]
    \centering
    \includegraphics[width=\linewidth]{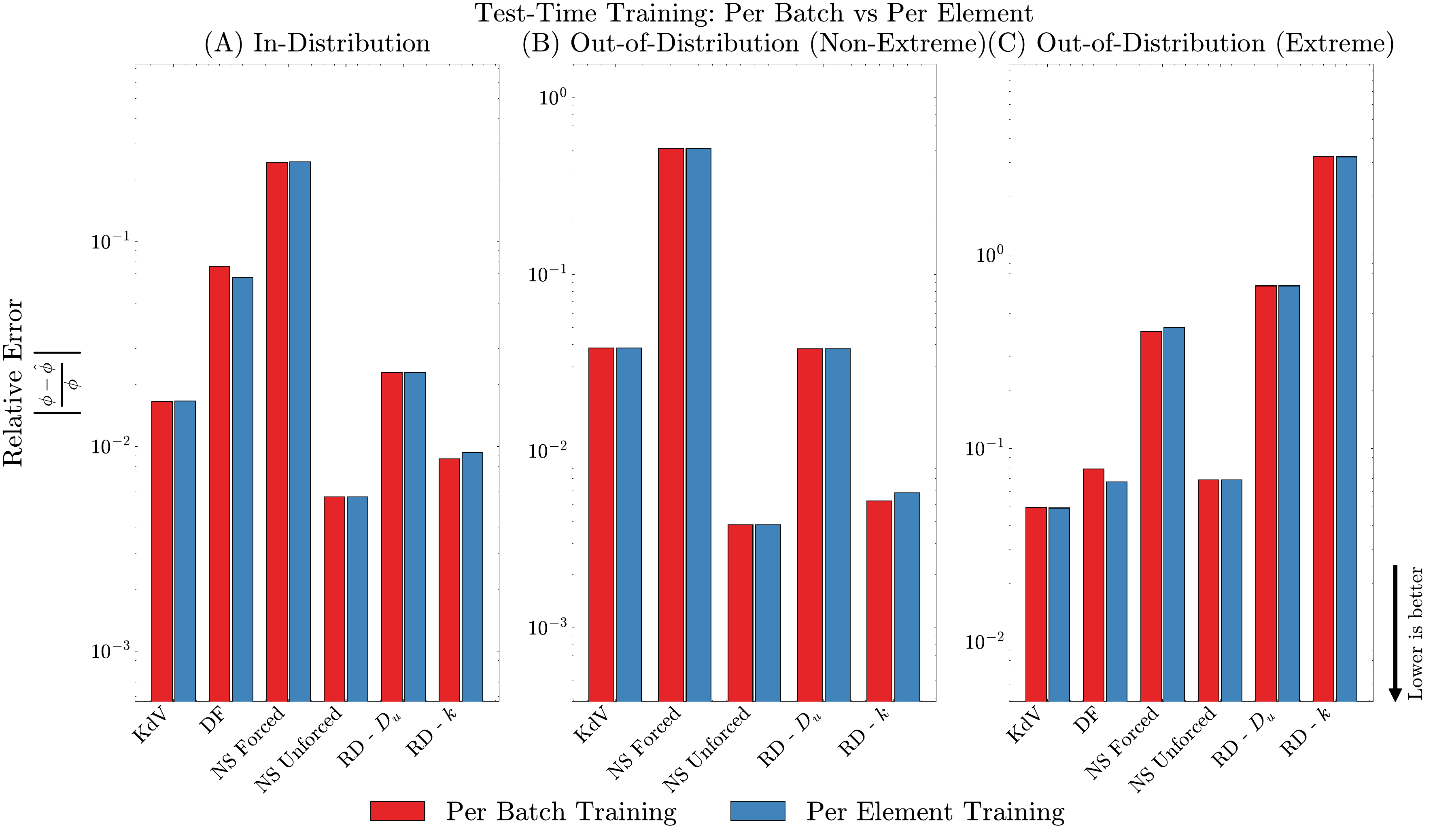}
    \caption{Comparing \textit{per-batch} vs \textit{per-batch} test time training. Both settings generally perform the same across systems and evaluation settings.}
    \label{fig:ttt-per-batch-vs-per-element}
\end{figure}

We compare performing TTT on a \textit{per-element} basis (batch size of 1) versus a \textit{per-batch} basis (batch size of 32) with anchor loss weights of 1 and show results in~\Figref{fig:ttt-per-batch-vs-per-element}.

TTT \textit{per-batch} generally perform the same \textit{per-sample} in all evaluation settings. 

\subsubsection{Test Time Tailoring Comparison with varying levels of ICs}
\begin{figure}[htbp!]
    \centering
    \includegraphics[width=\linewidth]{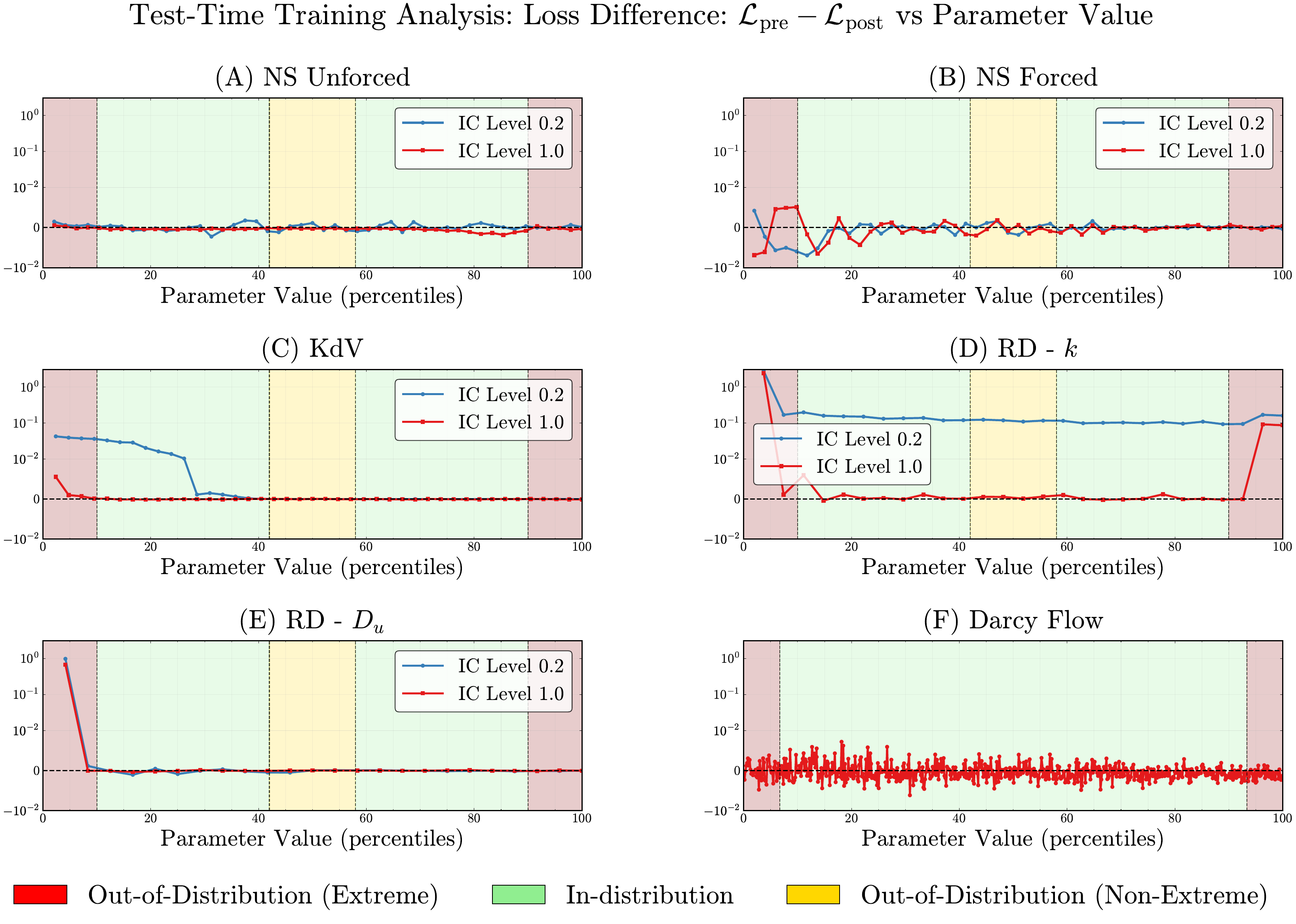}
    \caption{Evaluating test-time training with varying quantities of initial conditions during training time and across parameter regimes. For the KdV and 2D RD systems, TTT yields noticeable performance improvements in the extreme out-of-distribution parameter regime, and the improvement is more pronounced when performing TTT on unseen initial conditions during test time.}
    \label{fig:ttt_ics}
\end{figure}

We compare the performance of test-time training on models trained on 20\% of total available initial conditions and 100\% of initial conditions by system in \Figref{fig:ttt_ics} (this is a per-system view of Figure \ref{fig:datavresidual-and-ttt}). We see that test-time-tailoring generally helps in the case of limited number of iniital conditions and out of distribution parameter regimes parameter regimes. It never appreciably reduces performance. Therefore, we recommend performing TTT for inverse parameter estimation, especially over batches of input frames.

\subsection{Problem representation}
\label{sec:appendix-additional-results-repr}

\subsubsection{Varying number of temporal conditioning frames}

\begin{figure}[htbp!] 
    \centering
    \includegraphics[width=\linewidth]{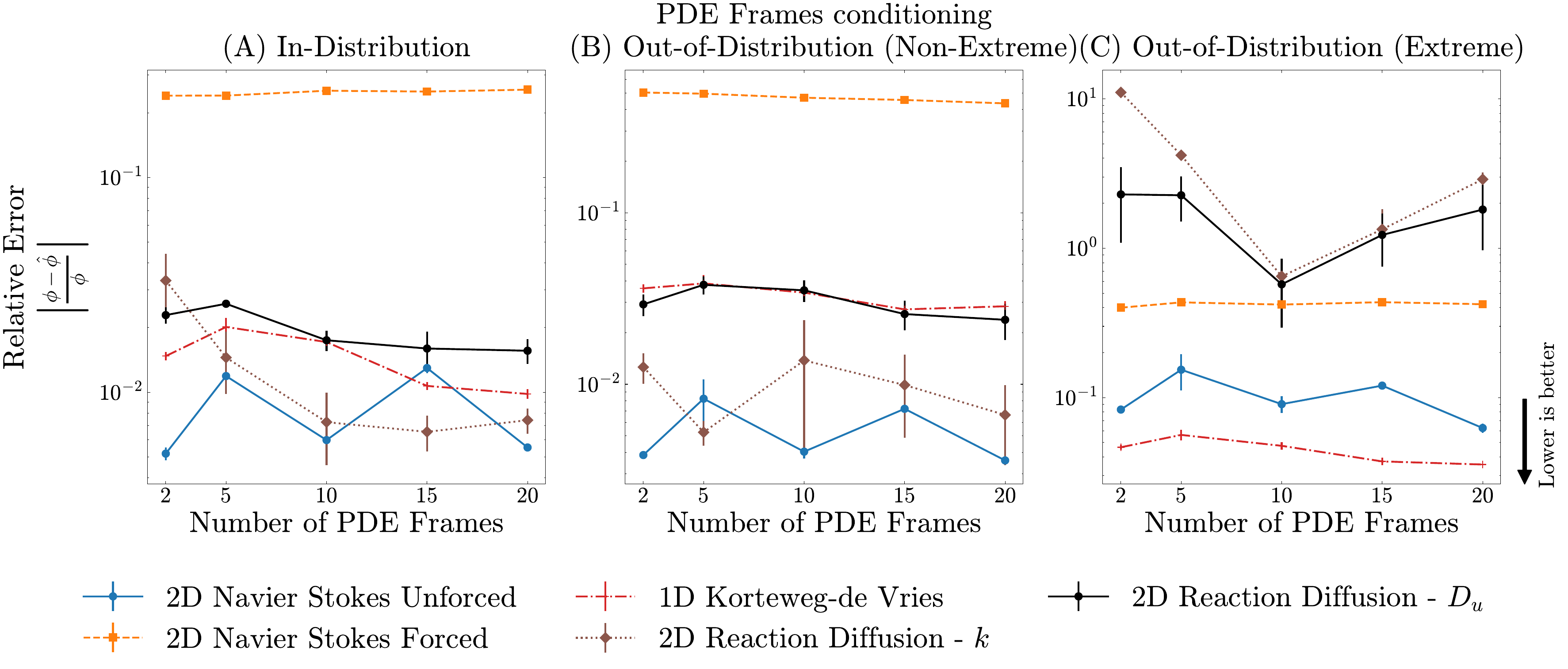}
    \caption{\textbf{Varying temporal conditioning frames.} Comparison of inverse problem performance with increasing number of temporal frames across evaluation settings for all time dependent settings. There is no consistent trend in scaling the conditioning PDE frames and inverse problem performance for FNO.}
    \label{fig:n_past_comparison}
\end{figure}

We vary the number of conditioning frames from the ground truth solution trajectories that are passed as input to the inverse model. We vary the number of input conditioning frames across $\{2, 5, 10, 15, 20\}$ for each system. For each configuration, we train separate models while maintaining consistent architecture and optimization settings. This experiment is not applicable to Darcy Flow since it is a time-independent PDE (Appendix~\ref{sec:darcy_flow_experiments}).
Results are shown in~\Figref{fig:n_past_comparison}.

 Unlike appending partial derivative information, which shows consistent benefits as demonstrated in Section \ref{sec:rep_results}, the optimal number of conditioning frames varies significantly across PDE systems. The Navier-Stokes unforced system generally shows better performance with more frames (10-20), while KdV attains optimal performance around 10 frames. The 2D reaction-diffusion parameters (k, Du, Dv) show less sensitivity to frame count in most evaluation settings. This system-specific variability highlights the importance of empirical tuning of this hyperparameter for each PDE family rather than assuming a universal "more is better" approach. The lack of consistent improvement with more frames may reflect redundancy in temporal information beyond a certain point or increased input dimensionality complicating the learning process.

% \subsubsection{Architectural Inductive Biases}

% \begin{figure}[htbp!] 
%     \centering
%     \includegraphics[width=\linewidth]{Figures/model_inductive_bias_comparison.pdf}
%     \caption{Comparison of model architectures across PDE systems. The chart shows relative parameter error for FNO, ResNet, and Transformer models under different evaluation scenarios. Lower values indicate better performance.}
%     \label{fig:model_comparison}
% \end{figure}

%   For time-dependent PDEs, Neural Operators generally outperform alternative architectures, particularly in out-of-distribution settings. This advantage likely stems from FNO's spectral bias, which naturally aligns with the smoothness properties of physical systems and enables better function space mapping. However, for the time-independent Darcy Flow system, ResNet significantly outperforms both FNO and ScOT. This suggests that the convolutional inductive bias may be better suited for static spatial patterns without temporal dynamics. The performance gap between architectures is most pronounced in challenging evaluation scenarios, indicating that appropriate inductive bias becomes increasingly important as the difficulty of the inverse problem increases.

\subsection{Scaling}
\label{sec:appendix-additional-results-scaling}

\subsubsection{Model scaling}
\label{sec:appendix-model-scaling}

\begin{figure}[htbp!]
   \centering
   \includegraphics[width=\linewidth]{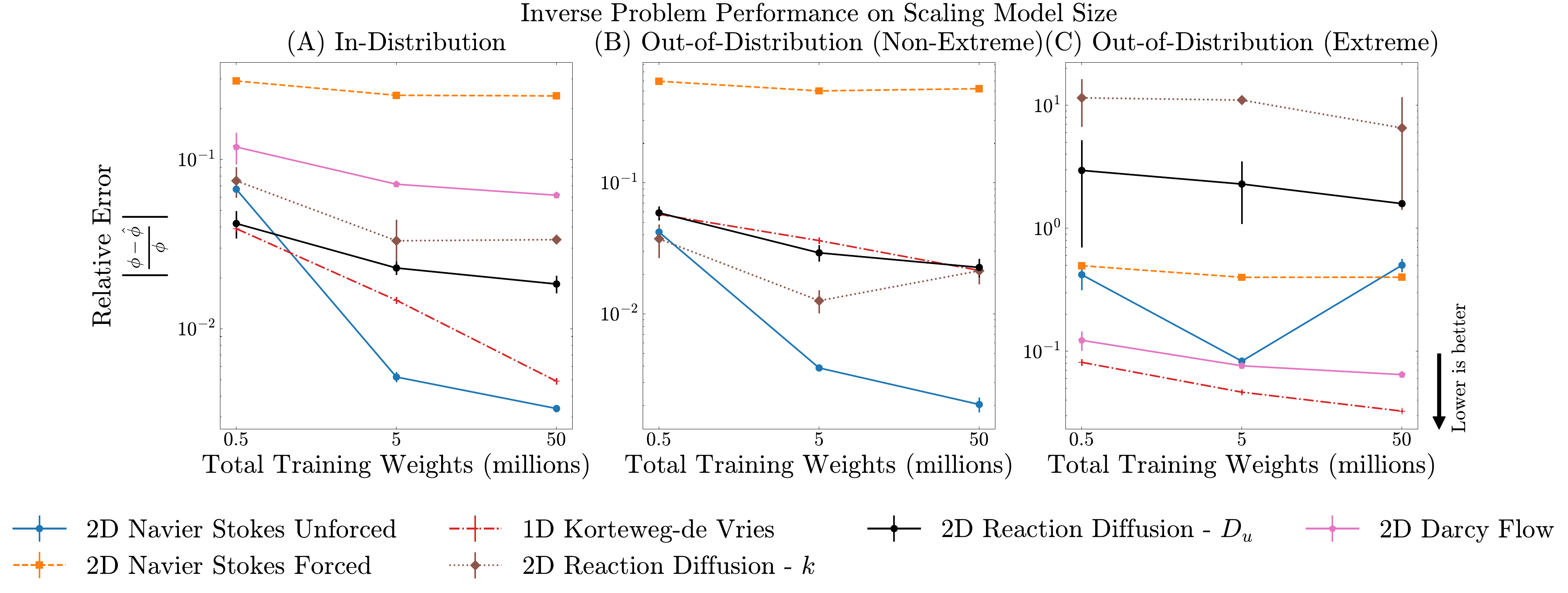}
   \caption{Inverse Problem Performance as a function of model size. Performance curves for different PDE systems across test scenarios as the number of trainable parameters increases from 0.5 million to 50 million. Model performance generally improves in all evaluation regimes when increasing the total number of model parameters.}
   \label{fig:model_scaling}
\end{figure}

To investigate the scaling behavior of neural operators for inverse PDE problems, we conducted a systematic model scaling experiment. We scaled the model size logarithmically across three configurations: 0.5M, 5M, and 50M parameters, focusing primarily on increasing width rather than depth.
Our scaling strategy deliberately avoided increasing model depth beyond 4-6 layers due to the known training difficulties associated with deeper neural operators. Instead, we significantly increased the hidden channel dimensions from 16 to 64 to 200 as we moved from 0.5M to 5M to 50M parameters. All models share the same core architecture: an FNO encoder with 16 modes, a convolutional downsampler with identical kernel configurations (3×3 kernels, stride 1, padding 2), and a consistent MLP design with ReLU activation. Importantly, we maintained a relatively constant ratio between the parameter count of the main FNO encoder and the downsampler networks across all scale configurations.
We present results in~\Figref{fig:model_scaling}.

As a general trend, we observe that ID and OOD non-extreme performance generally improves with increase model size (\Figref{fig:model_scaling}A and B). However, this trend does not hold with the OOD non-extreme split. This suggests that bigger models show better interpolative performance but do not guarantee better extrapolation.

\subsubsection{Data scaling}
In addition to scaling the number of initial conditions in the dataset, we also scale the number of unique PDE parameters and the length of the ground truth trajectories to which the model has access.

\begin{figure}[htbp!]
    \centering
    \includegraphics[width=\linewidth]{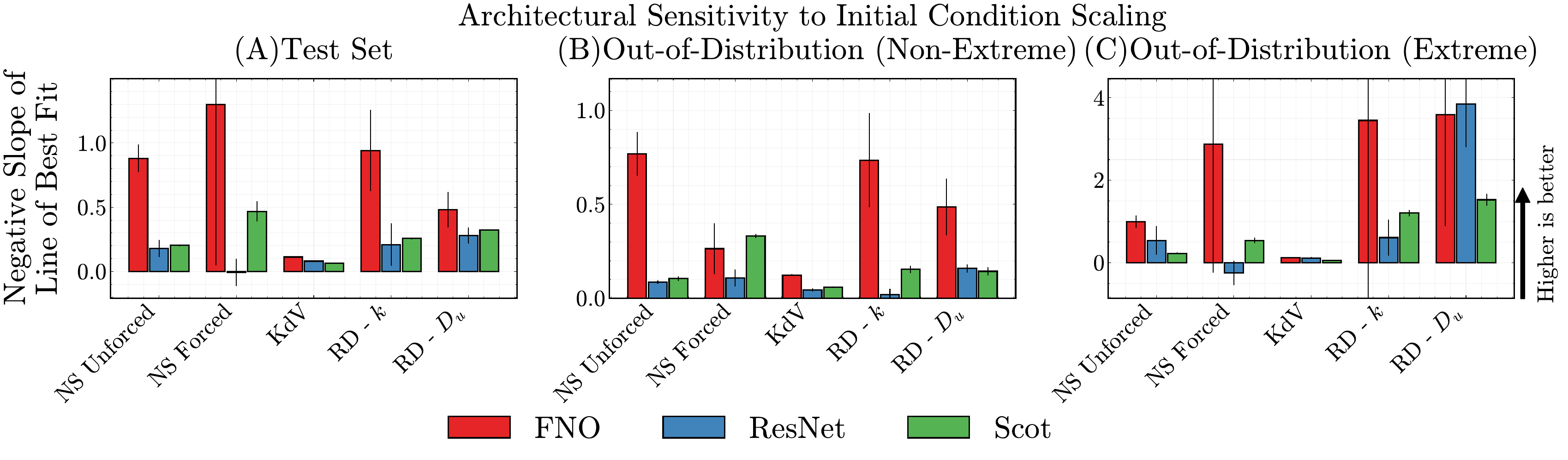}
    \caption{\textbf{Effect of initial condition scaling across architectures} Evaluating data efficiency of different architectures to scaling the number of initial conditions. Increasing the total number of initial conditions during training improves performance for all architectures, with FNO being the most data efficient.}
    \label{fig:inductive_bias_scaling_ic_comparison}
\end{figure}

\paragraph{Effect of scaling initial conditions across architectures}
More initial conditions generally lead to better performance for all architectures in most evaluation settings and systems, see~\Figref{fig:inductive_bias_scaling_ic_comparison}. FNO generally exhibits greater data efficiency when scaling up the number of initial conditions compared to ResNet and scOT.

\paragraph{Scaling the number of PDE parameters.}
We investigate the effect of reducing the fraction of unique PDE parameters included in the training dataset. We remove the selected fraction of parameters from the training set and uniformly sample frames from the remaining parameters.
\begin{figure}[htbp!]
    \centering
    \includegraphics[width=\linewidth]{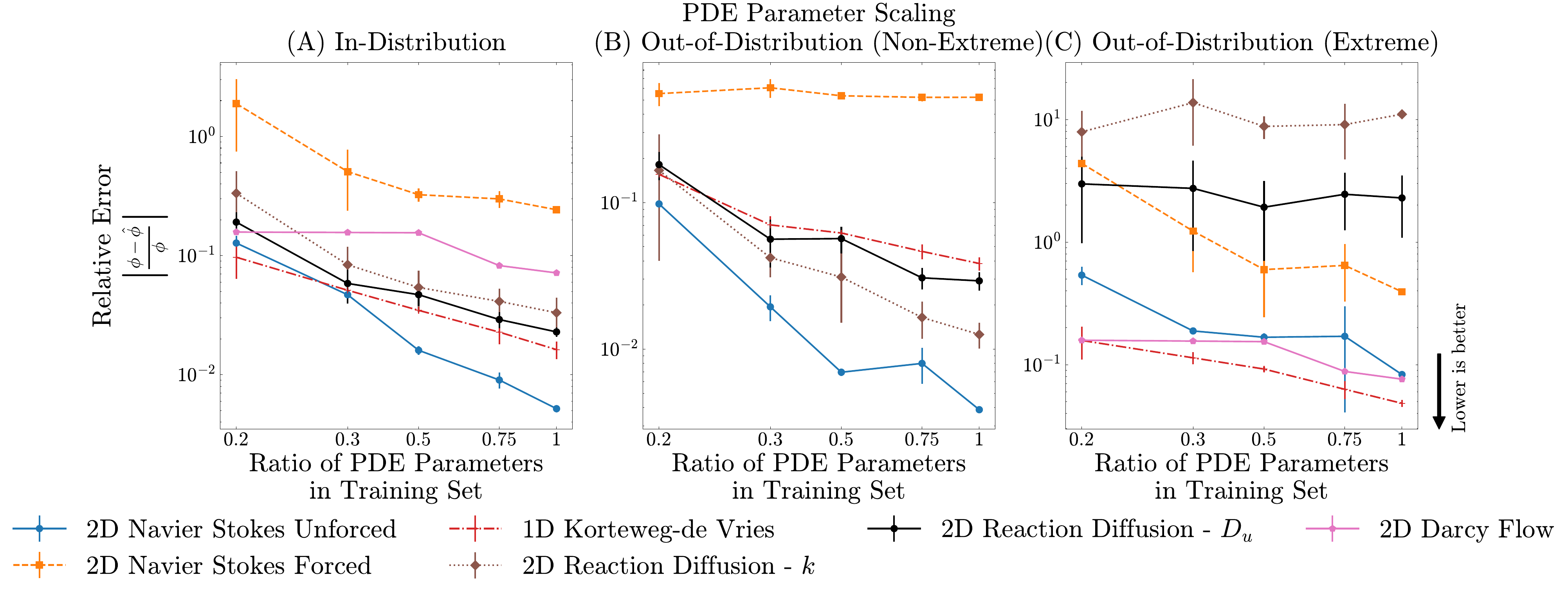}
    \caption{\textbf{Effect of scaling the total number of generated PDE parameters.} Evaluating inverse problem performance on different quantities of available data by scaling the total number of generated parameter settings of the PDEs. Increasing the number of training trajectories along generated PDE parameters improves test time performance on unseen parameters.}
    % \label{fig:scaling_params}
\end{figure}

Increasing the number of parameters generally leads to better performance in all evaluation settings for most systems. However, as demonstrated in~\Figref{fig:results-scaling}, the improvements are not as large as those resulting from increasing the number of initial conditions per PDE parameter.

\begin{figure}[htbp!]
    \centering
    \includegraphics[width=\linewidth]{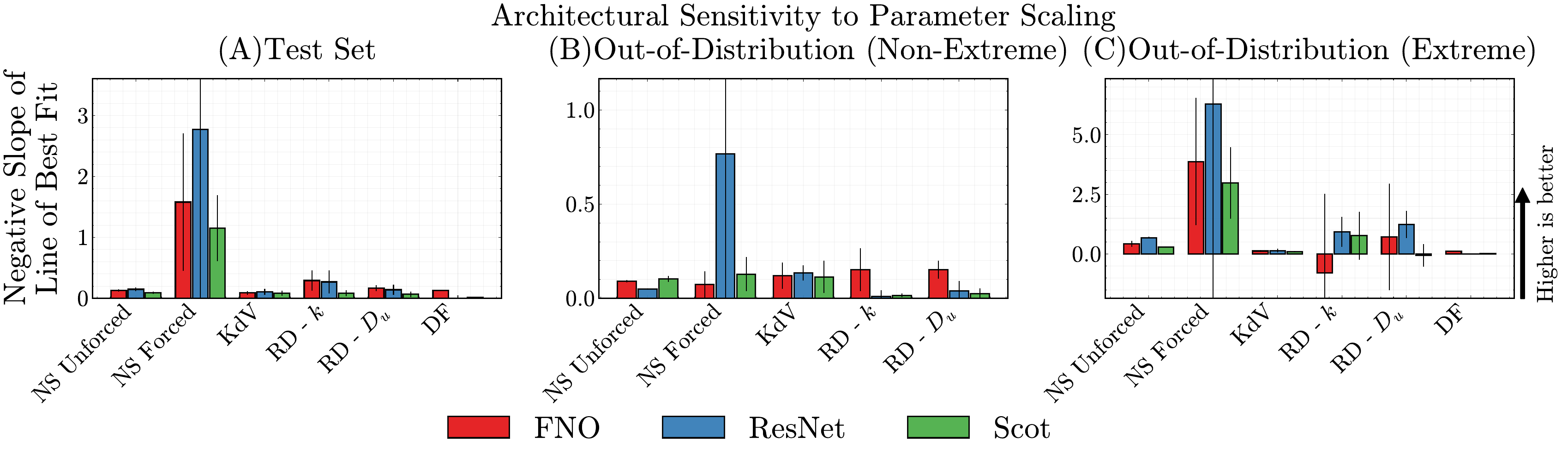}
    \caption{\textbf{Effect of PDE parameter scaling across architectures} Evaluating data efficiency of different architectures to scaling the number of PDE parameters. Increasing the total number of initial conditions during training improves performance for all architectures with FNO being the most data efficient.}
    \label{fig:inductive_bias_scaling_params_comparison}
\end{figure}

\paragraph{Effect of scaling PDE parameters across architectures}
More PDE parameters generally lead to better performance for all architectures in most evaluation settings and systems ~\Figref{fig:inductive_bias_scaling_params_comparison}. 

\paragraph{Scaling the length of generated trajectories}

\begin{figure}[htbp!]
    \centering
    \includegraphics[width=\linewidth]{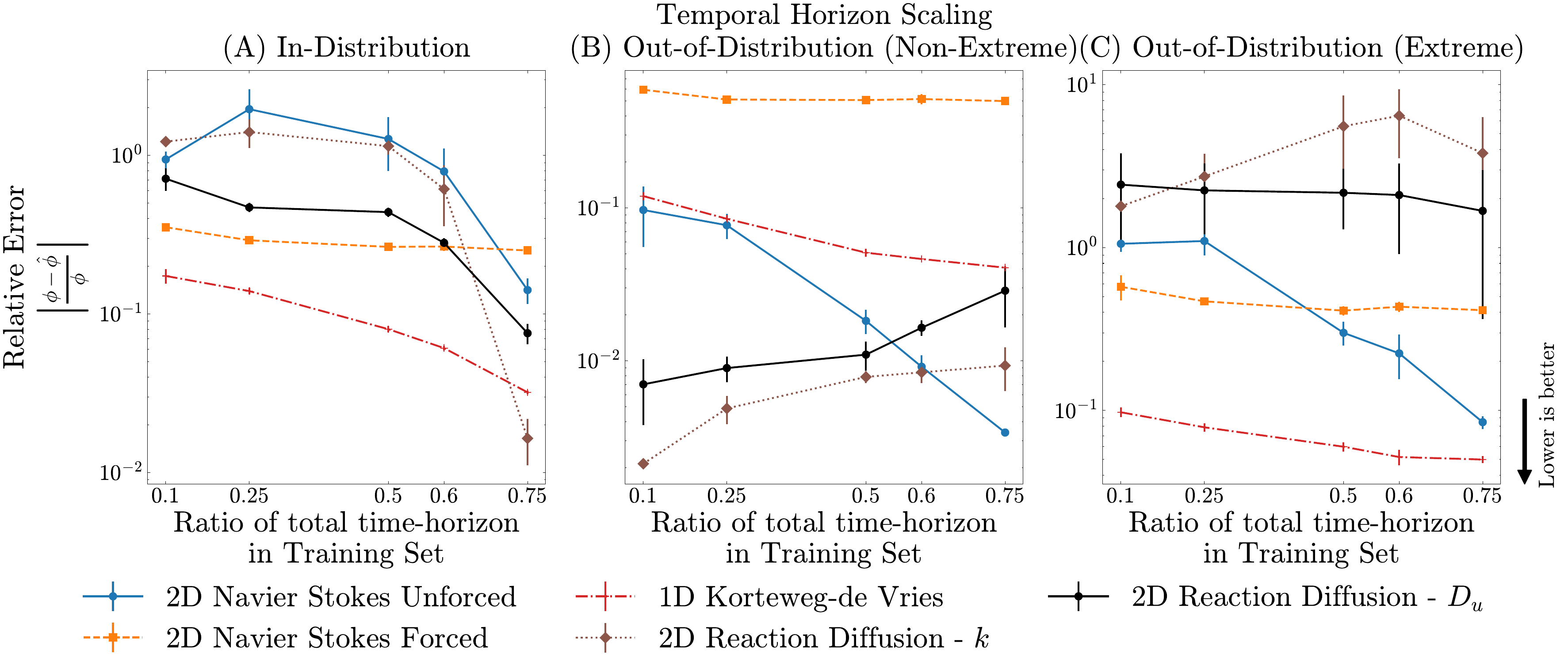}
    \caption{\textbf{Effect of increasing the ground truth time horizon.}
    Evaluating Inverse problem performance on different quantities of available data by scaling the total time horizon of the training solution fields. Increasing the total time horizon of training trajectories improves performance on held out future time frames.}
    \label{fig:scaling_time}
\end{figure}

As another way to investigate data scaling properties, we vary the total time horizon of the ground truth trajectories used for training, leaving out the final 25\% of the in-distribution trajectories. We show results in~\Figref{fig:scaling_time}. This experiment is not applicable to Darcy Flow since it is a time-independent PDE (Appendix~\ref{sec:darcy_flow_experiments}).

Longer training time horizons generally lead to better performance for neural operators in the in-distribution setting, as shown in~\Figref{fig:scaling_time}(A) where relative error decreases as the percentage of PDE parameters increases. However, this general pattern does not hold in any of the out-of-distribution settings~\Figref{fig:scaling_time} (B, C). Therefore, training on longer temporal trajectories does not lead to representations that perform better on unseen parameters in either non-extreme or extreme out-of-distribution test cases. This suggests that extending the temporal horizon during training does not inherently improve generalization capability when estimating PDE parameters from solution fields outside the training distribution.

\subsection{Physicality of Predicted Parameters}
\label{sec:self-consistency-metric}

While relative error is a convenient way to compare predicted and true parameters, it does not directly test whether the predictions are physically meaningful. A complementary evaluation is to evolve a numerical solution using the predicted parameter and compare physically meaningful diagnostics of the resulting trajectory to the reference trajectory, thus capturing a notion of ``self-consistency" of the predictions.

\begin{figure}
    \centering
    \includegraphics[width=0.9\linewidth]{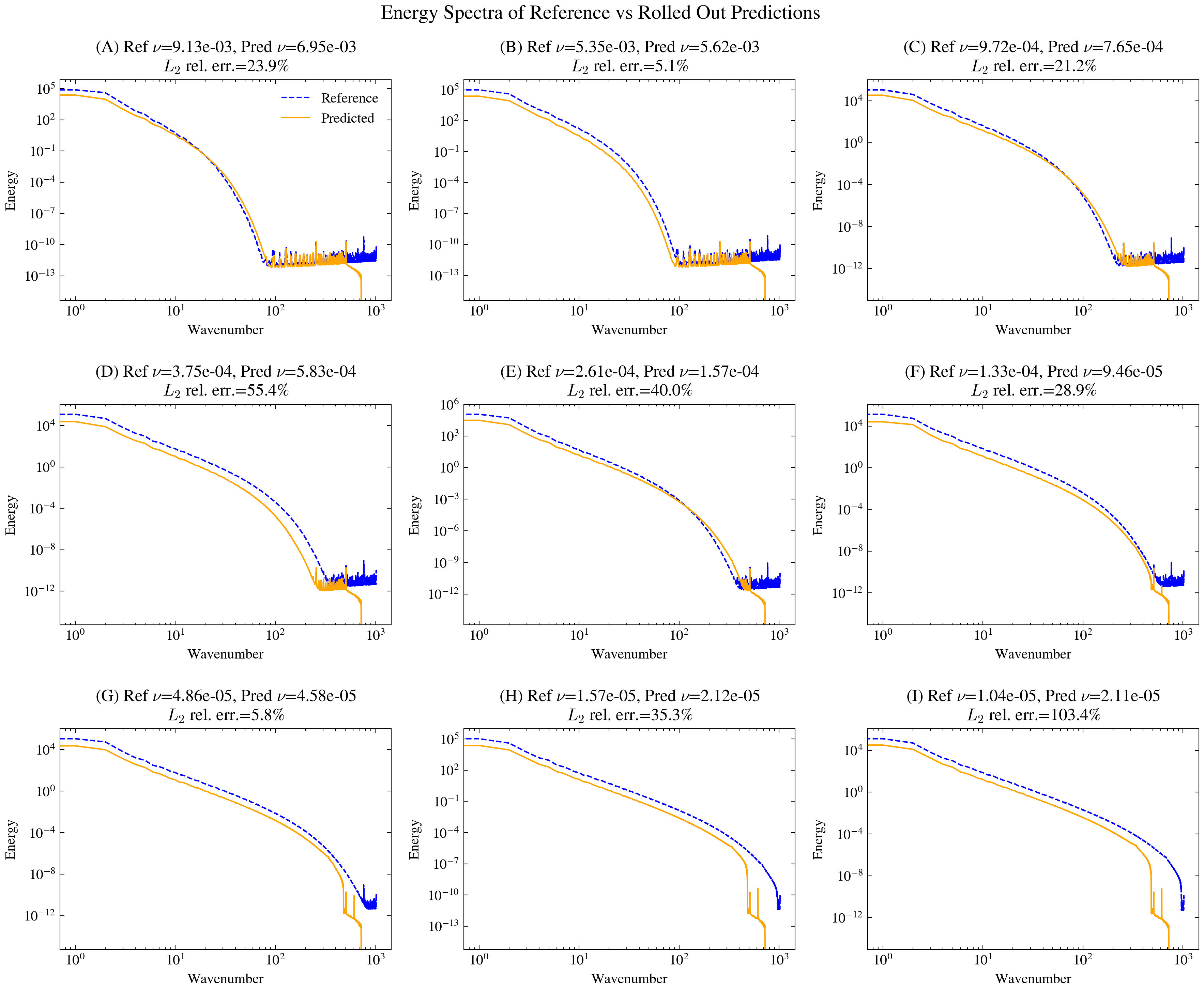}
    \caption{\textbf{Energy spectra of predicted Forced 2D Navier Stokes vs. reference solutions.} Energy spectra corresponding to predictions are in solid lines, while references are dashed.}
    \label{fig:2dtf-predicted-vs-reference-energy}
\end{figure}

We apply this idea to forced 2D Navier–Stokes, where the viscosity $\nu$ strongly controls the system’s dissipation and thus the energy distribution across length scales. In Figure~\ref{fig:2dtf-predicted-vs-reference-energy} we plot the energy spectra from (i) a reference simulation run with the true PDE parameter, and (ii) a simulation run with the parameter predicted by an FNO-based inverse model when conditioned on frames from the reference simulation. Predictions are shown with orange, solid lines and references with blue, dashed lines. We repeat this for 9 pairs of predicted and true parameters spanning the parameter range. 

In the reference simulations, more viscous flows (higher $\nu$ / lower Reynolds number) exhibit an earlier drop off in the energy spectrum, whereas lower-viscosity (more turbulent) regimes maintain appreciable energy up to higher wavenumbers, before entering the dissipative range.

Overall, simulations run with the predicted parameters largely preserve the energy spectra. Rolled-out solutions reproduce the qualitative location and shape of the spectral drop-off associated with dissipation.
The low-viscosity regime, Figure~\ref{fig:2dtf-predicted-vs-reference-energy}(G-I), is noticeably more sensitive, with small parameter discrepancies leading to visibly larger shifts in the high-wavenumbers, consistent with the turbulent and chaotic nature at low viscosity.
For high viscosity parameters, even predictions with higher relative error in Figure~\ref{fig:2dtf-predicted-vs-reference-energy}(A-F) produce physically consistent solutions.

This analysis underscores that evaluation is ultimately task-dependent.
Scalar relative errors are useful summary statistics, but physics-based diagnostics such as spectra more directly test whether a predicted parameter leads to the correct qualitative behavior when used in a forward simulator. 
However, performing this kind of rollout-based, physics-level evaluation exhaustively across all parameters and systems is computationally prohibitive. Even for a single trained model, it would require regenerating a set of simulations comparable in size to the full dataset, multiplied by the number of random seeds (three in our setup), which quickly becomes intractable. We thus recommend that it be used sparingly as a diagnostic tool.

\subsection{ Noisy Inputs}
\label{sec:noisy_inputs_results}
We induce partial observability via two degradation operators: salt-and-pepper (S\&P) noise and Butterworth (BW) filtering. Salt-and-pepper noise replaces a specified proportion of pixels with either white (salt) or black (pepper) values, modeling random sensor failure or dropout. In contrast, the Butterworth filter removes a fraction of high-frequency modes from the input solution fields, simulating instruments with limited bandwidth or spatial resolution. These two corruption models probe complementary failure modes: S\&P evaluates robustness to spatially sparse, unstructured corruption, while BW filtering tests the ability to recover parameters when fine-scale (high-frequency) information is systematically removed.

\begin{figure}[h]
    \centering
    \includegraphics[width=0.9\linewidth]{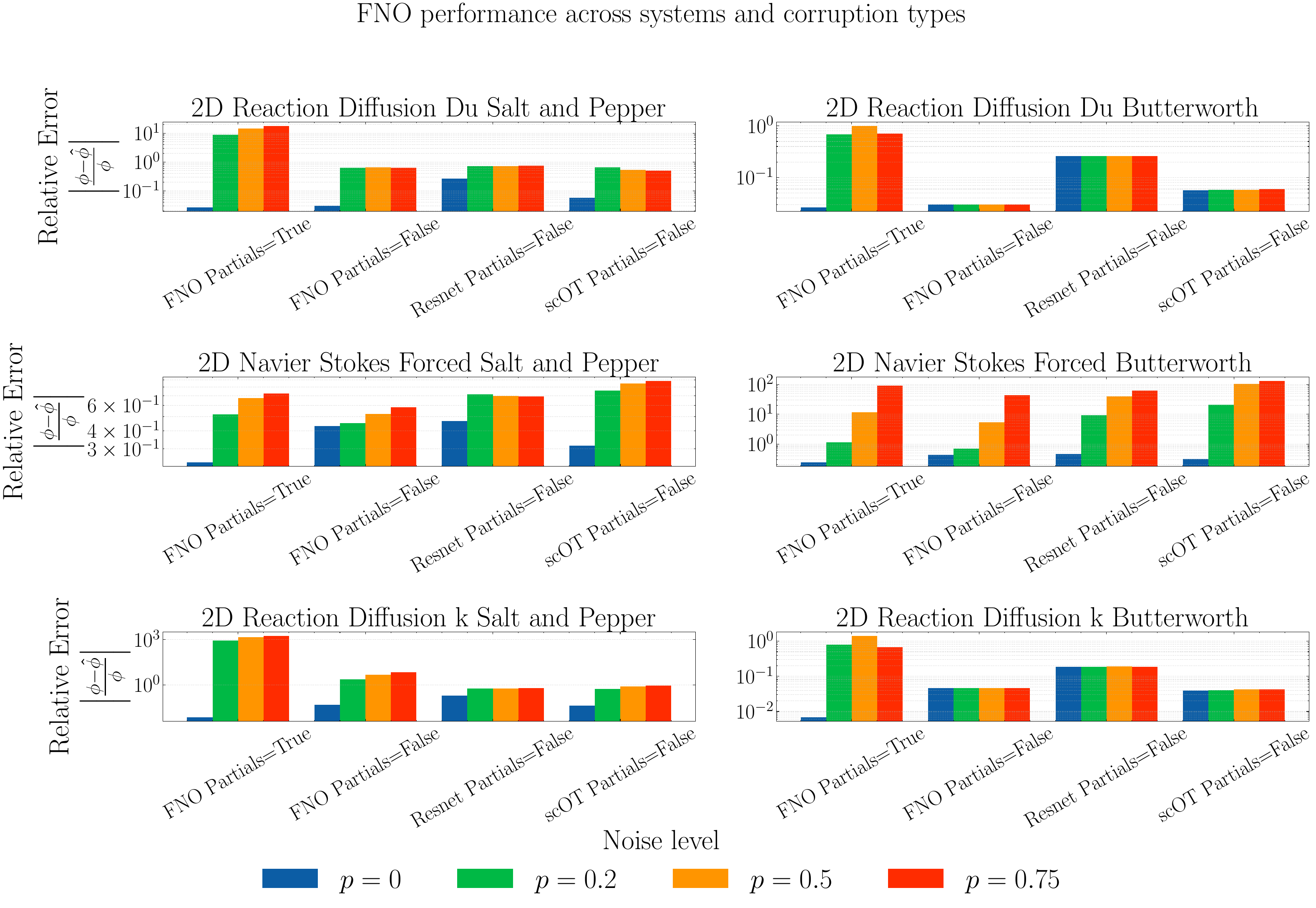}
    \caption{\textbf{Impact of noisy inputs on models trained on non-noisy solution fields.} Evaluating models trained on clean inputs across all design axes under varying levels of salt-and-pepper noise and Butterworth filtering. Models trained without partial-derivative inputs exhibit improved robustness to these degradations. }
    \label{fig:noisy-solution-fields}
\end{figure}

We construct evaluation splits with varying degradation levels. For S\&P noise, we apply corruption levels with probabilities $p \in \{0.2, 0.5, 0.75\}$. For BW filtering, we use a filter of order $6$ and remove high-frequency modes with ratios $p \in \{0.2, 0.5, 0.75\}$. We first evaluate models trained on clean data ($p = 0$) to assess their robustness under increasing levels of test-time degradation (Figure \ref{fig:noisy-solution-fields}).

\begin{figure}[h]
    \centering
    \includegraphics[width=0.9\linewidth]{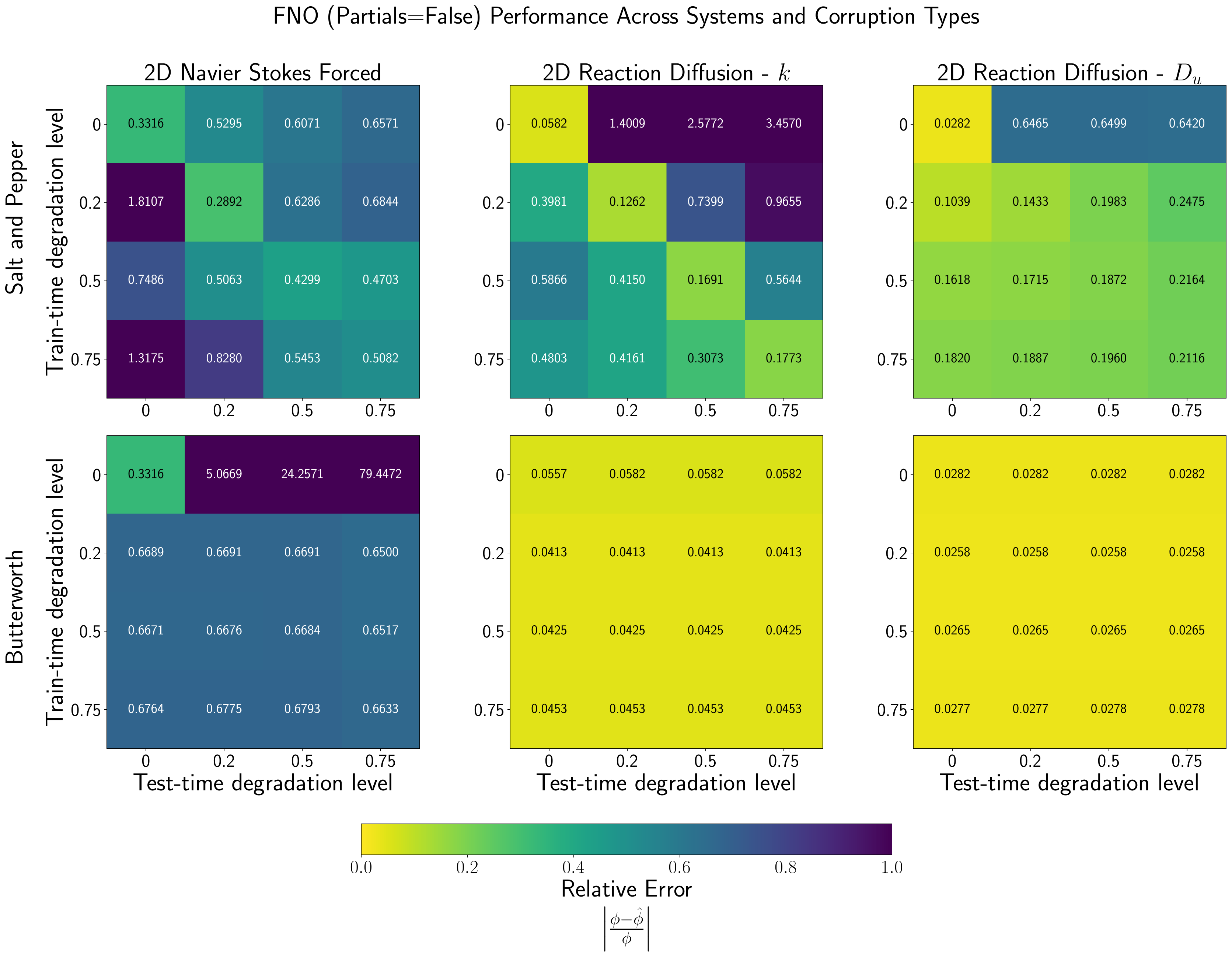}
    \caption{\textbf{Evaluation of FNO robustness after training under observational degradation}. Heatmaps show relative error as a function of train-time (rows) and test-time (columns) degradation levels for salt-and-pepper (top) and Butterworth (bottom) corruption across three PDE systems. Salt-and-pepper noise induces a strong diagonal structure, indicating sensitivity to mismatch between training and inference corruption levels, whereas Butterworth filtering yields nearly uniform performance across test-time degradation once seen during training. Training on corrupted inputs improves robustness but introduces a trade-off in performance on clean data.}
    \label{fig:noisy-fno-performance}
\end{figure}

Armed with the insight that removing partial derivative conditioning improves the robustness of the models to degradation (Figure \ref{fig:noisy-solution-fields}), we fix this setting and now train models on varying levels of degradation without conditioning on partial derivatives. In~\Figref{fig:noisy-fno-performance}, we visualize a heat map of model performance as a function of the level of degradation applied at both train- and test-time.

Across Salt-and-Pepper (SNP) corruption, a clear diagonal preference emerges: models achieve the lowest error when the training corruption level matches the inference corruption level, and performance degrades as one moves away from this match in either direction. In particular, models are generally robust when the inference noise is less than or approximately equal to the training noise, but performance drops more sharply when the inference corruption exceeds what was seen during training. This suggests that the learned inverse operator is effectively calibrated to the corruption distribution encountered during training and does not extrapolate well beyond it. Training with non-zero corruption significantly improves robustness relative to clean-only training, which exhibits the largest degradation under noisy inference. Among the tested settings, training at p=0.5p=0.5p=0.5 provides the most balanced behavior, offering reasonable performance across a range of noise levels without over-specializing. However, this robustness comes at a cost: models trained on corrupted fields consistently underperform on clean inputs compared to models trained purely on clean data.

In contrast, the Butterworth (BW) corruption experiments exhibit qualitatively different behavior. Once models are trained with even moderate filtering (e.g., $p \approx 0.2$), performance becomes largely invariant across the inference corruption axis, with near-constant error across test noise levels. More precisely, training at a given level of degradation yields models that are robust to equal or lower levels of filtering at inference. This suggests that the learned inverse mapping is insensitive to the exact magnitude of spectral corruption, provided the corruption class is observed during training. This behavior can be attributed to the smooth, structured nature of Butterworth filtering: the roll-off suppresses high-frequency components while preserving low-frequency structure, effectively constraining the inverse problem to a stable subspace that generalizes across filter strengths.

\subsection{Non-Uniform Grids}
\label{sec:non_uniform_grid_results}
\begin{figure}[H]
    \centering
    \includegraphics[width=0.9\linewidth]{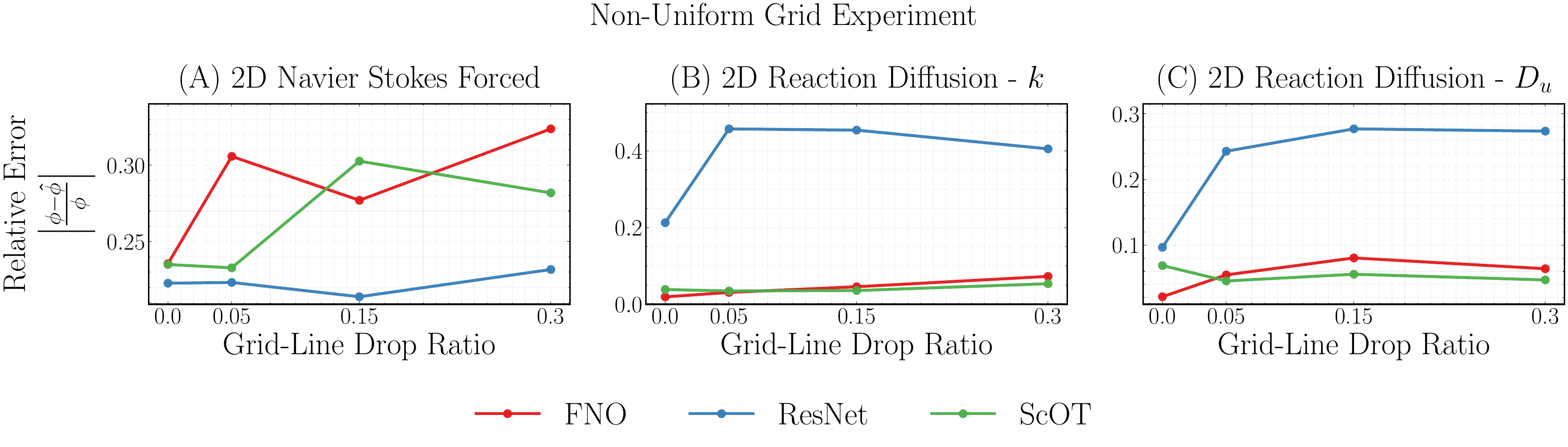}
    \caption{\textbf{Impact of Non-Uniform Grids on PDE inverse problem performance.} Comparison of architectural inductive biases (FNO, ResNet, scOT) to varying degrees of non-uniformity in solution fields}
    \label{fig:non-uniform-grid-exp}
\end{figure}

We study the impact of non-uniform spatial discretization by randomly removing spatial grid lines with probability $p$, yielding solution fields defined on irregular grids. We evaluate FNO, ResNet, and scOT across increasing levels of grid sparsification, with $p \in \{0.05, 0.15, 0.3\}$~\Figref{fig:non-uniform-grid-exp}. As expected, higher drop probabilities lead to increased reconstruction error; however, the magnitude and trend of degradation vary across models and underlying PDE systems. We hypothesize that ResNet, with it’s convolutional bias, performs the worst due to the limited receptive field in early layers, while FNO and scOT can more easily learn global features via the Fourier layers or attention.

\subsection{Benchmark Results}
\label{sec:appendix-model-benchmark-results-tables}

We provide the relative errors of the various models across all systems and settings in a table format.

\begin{table}[H]
\centering
\begin{tabular}{lcccc}
\toprule
System & FNO & ResNet & scOT & DeepONet \\
\midrule
2D NS Unforced & $\underline{0.0052 \pm 0.0004}$ & $0.0146 \pm 0.0008$ & $0.0164 \pm 0.0015$ & $0.0174 \pm 0.0010$ \\
2D NS Forced   & $0.2402 \pm 0.0045$ & $0.3964 \pm 0.1228$ & $0.2589 \pm 0.0173$ & $\underline{0.2032 \pm 0.0035}$ \\
1D KdV         & $\underline{0.0147 \pm 0.0007}$ & $0.0197 \pm 0.0055$ & $0.0223 \pm 0.0011$ & $0.0166 \pm 0.0006$ \\
2D RD -- $k$   & $\underline{0.0331 \pm 0.0110}$ & $0.3426 \pm 0.0968$ & $0.0453 \pm 0.0049$ & $0.4129 \pm 0.0194$ \\
2D RD -- $D_u$ & $\underline{0.0229 \pm 0.0020}$ & $0.1740 \pm 0.0559$ & $0.0749 \pm 0.0049$ & $0.2406 \pm 0.0306$ \\
2D DF          & $0.0671 \pm 0.0032$ & $\underline{0.0006 \pm 0.0001}$ & $0.0181 \pm 0.0005$ & $\underline{0.0006 \pm 0.0003}$ \\
\bottomrule
\end{tabular}
\caption{Model relative error on the test set. The best value in each row is underlined.}
\label{tab:arch_test}

\end{table}

\begin{table}[H]
\centering
\begin{tabular}{lcccc}
\toprule
System & FNO & ResNet & scOT & DeepONet \\
\midrule
2D NS Unforced & $\underline{0.0039 \pm 0.0002}$ & $0.0086 \pm 0.0005$ & $0.0320 \pm 0.0046$ & $0.0097 \pm 0.0003$ \\
2D NS Forced   & $0.5026 \pm 0.0242$ & $0.2390 \pm 0.0038$ & $0.4657 \pm 0.0550$ & $\underline{0.2331 \pm 0.0215}$ \\
1D KdV         & $\underline{0.0362 \pm 0.0021}$ & $0.0635 \pm 0.0054$ & $0.0472 \pm 0.0008$ & $0.0624 \pm 0.0022$ \\
2D RD -- $k$   & $\underline{0.0126 \pm 0.0025}$ & $0.2737 \pm 0.0508$ & $0.0249 \pm 0.0048$ & $0.2551 \pm 0.0863$ \\
2D RD -- $D_u$ & $\underline{0.0292 \pm 0.0042}$ & $0.1285 \pm 0.0280$ & $0.0991 \pm 0.0310$ & $0.2041 \pm 0.0727$ \\
\bottomrule
\end{tabular}
\caption{Model relative error on OOD (Non-Extreme) set. The best value in each row is underlined.}
\label{tab:arch_ood_nonextreme}

\end{table}

\begin{table}[H]
\centering
\begin{tabular}{lcccc}
\toprule
System & FNO & ResNet & scOT & DeepONet \\
\midrule
2D NS Unforced & $0.0831 \pm 0.0039$ & $0.1608 \pm 0.0128$ & $\underline{0.0679 \pm 0.0046}$ & $0.1919 \pm 0.0433$ \\
2D NS Forced   & $0.3998 \pm 0.0208$ & $0.9751 \pm 0.3837$ & $\underline{0.3669 \pm 0.0715}$ & $0.5144 \pm 0.0149$ \\
1D KdV         & $\underline{0.0465 \pm 0.0018}$ & $0.0758 \pm 0.0245$ & $0.0569 \pm 0.0028$ & $0.0546 \pm 0.0024$ \\
2D RD -- $k$   & $11.0376 \pm 0.5903$ & $1.8791 \pm 0.4135$ & $2.3177 \pm 0.1775$ & $\underline{1.8405 \pm 0.4136}$ \\
2D RD -- $D_u$ & $2.2911 \pm 1.2060$ & $2.5445 \pm 0.3617$ & $\underline{1.6681 \pm 0.3646}$ & $2.2492 \pm 0.6320$ \\
\bottomrule
\end{tabular}
\caption{Model relative error on the OOD (Extreme) set. The best value in each row is underlined.}
\label{tab:arch_ood_extreme}
\end{table}
            
% D

\end{document}